\documentclass[acmtog]{acmart}
\usepackage{overpic}
\usepackage{xcolor}
\usepackage{multirow}
\usepackage{placeins}

\newcommand{\under}[1]{\underline{\textbf{#1}}}

\usepackage[ruled]{algorithm2e} % For algorithms

\SetAlFnt{\small}
\SetAlCapFnt{\small}
\SetAlCapNameFnt{\small}
\SetAlCapHSkip{0pt}

\begin{document}
% Title portion
\title{Neural-Singular-Hessian:~Implicit Neural Representation of Unoriented Point Clouds by Enforcing Singular Hessian
%over Signed Distance Function
%Implicit Neural Representation from Unoriented Point Clouds by Eliminating Critical Points Near the Surface
%Neural-CPS:~Implicit Neural Representation from Unoriented Point Cloud by Critical Point Suppressing
%Eliminating Critical Points Near the Surface
}
%Saddle points-reject shape implicit neural representation for }

% DO NOT ENTER AUTHOR INFORMATION FOR ANONYMOUS TECHNICAL PAPER SUBMISSIONS TO SIGGRAPH 2019!
\author{Zixiong Wang}
\orcid{0000-0002-6170-7339}
\affiliation{%
 \institution{Shandong University}
  \country{China}
 }
\email{zixiong_wang@outlook.com}

\author{Yunxiao Zhang}
\affiliation{%
\institution{Shandong University}
 \country{China}
}
\email{zhangyunxiaox@gmail.com}

\author{Rui Xu}
\affiliation{%
\institution{Shandong University}
 \country{China}
}
\email{xrvitd@163.com}

\author{Fan Zhang}
\affiliation{%
 \institution{Shandong Technology and Business University}
  \country{China}
 }
\email{zhangfan@sdtbu.edu.cn}

\author{Peng-Shuai Wang}
\affiliation{
  \department{Wangxuan Institute of Computer Technology}
  \institution{Peking University}
  \country{China}
}
\email{wangps@hotmail.com}

\author{Shuangmin Chen}
\affiliation{%
 \institution{Qingdao University of Science and Technology}
 \country{China}
}
\email{csmqq@163.com}

\author{Shiqing Xin*}
\affiliation{%
\institution{Shandong University}
 \country{China}
}
\email{xinshiqing@sdu.edu.cn}

\author{Wenping Wang}
\affiliation{%
\institution{Texas A\&M University}
 \country{USA}
}
\email{wenping@tamu.edu}

\author{Changhe Tu}
\affiliation{%
\institution{Shandong University}
 \country{China}
}
\email{chtu@sdu.edu.cn}

\begin{abstract}
Neural implicit representation is a promising approach for reconstructing surfaces from point clouds. Existing methods combine various regularization terms, such as the Eikonal and Laplacian energy terms, to enforce the learned neural function to possess the properties of a Signed Distance Function (SDF). However, inferring the actual topology and geometry of the underlying surface from poor-quality unoriented point clouds remains challenging. In accordance with Differential Geometry, the Hessian of the SDF is singular for points within the differential thin-shell space surrounding the surface. Our approach enforces the Hessian of the neural implicit function to have a zero determinant for points near the surface. This technique aligns the gradients for a near-surface point and its on-surface projection point, producing a rough but faithful shape within just a few iterations. By annealing the weight of the singular-Hessian term, our approach ultimately produces a high-fidelity reconstruction result. Extensive experimental results demonstrate that our approach effectively suppresses ghost geometry and recovers details from unoriented point clouds with better expressiveness than existing fitting-based methods.

\end{abstract}


%
% End-generated code
%


\begin{teaserfigure}
    \centering
    \includegraphics[width=\textwidth]{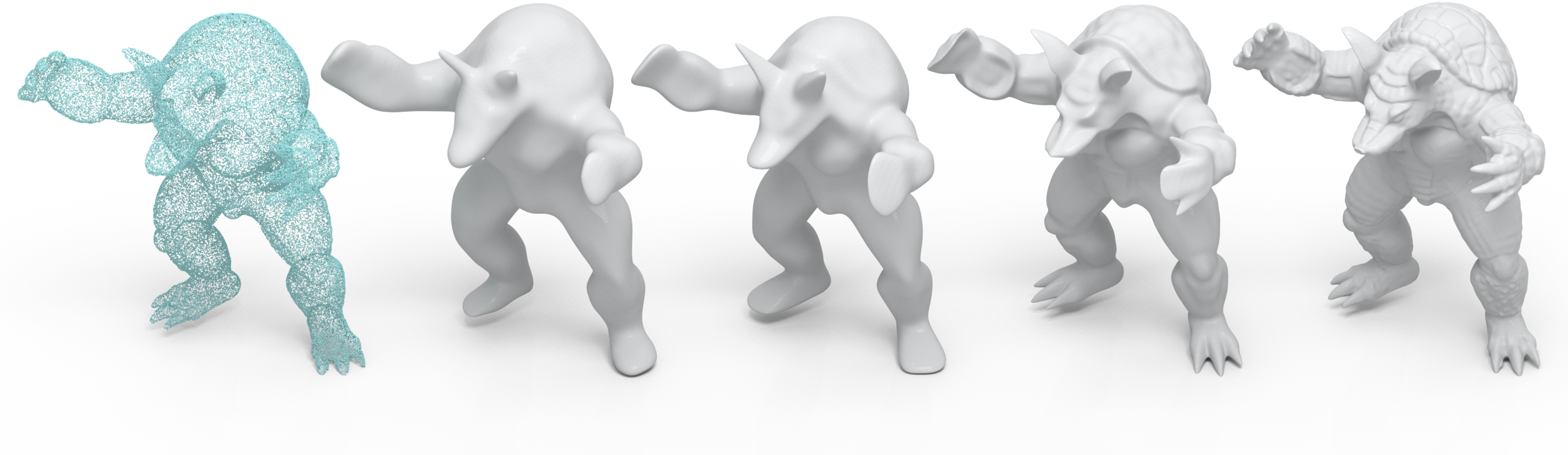}
    \vspace{-16mm}
    \caption{Starting with an unoriented point cloud (left), our approach proposes an implicit neural representation 
    by enforcing the Hessian of the implicit function to be singular for points in close proximity to the surface. By gradually reducing the weight of the singular-Hessian term, it works in a coarse-to-fine fashion and can ultimately produce a high-fidelity reconstruction result (right).}
    \label{fig:teaser}
\end{teaserfigure}

\maketitle

\section{Introduction}
% In recent years, many learning-based approaches have 
% been proposed to recover the implicit representation of the underlying surface from a point cloud,
% which is a fundamental task in computer graphics and computer vision.
% %considerable attention in recent years due to their excellent expressiveness.
% Despite the great progress~\cite{IGR, SIREN, IDF} on surface reconstruction from a high-quality point cloud with reliable normals,
% it remains a challenging yet fascinating research problem to predict a faithful surface from an unoriented point cloud due to the lack of sufficient geometric priors. 

In recent years, numerous learning-based approaches have been developed to recover the implicit representation of underlying surfaces from point clouds, a fundamental task in computer graphics and computer vision. Despite significant progress~\cite{IGR, SIREN, IDF, Rui2022RFEPS} in surface reconstruction from high-quality point clouds with reliable normals, predicting a faithful surface from an unoriented point cloud remains a challenging and intriguing research problem due to the lack of sufficient geometric priors.

% \begin{figure}[!tp]
%     \centering
%     \includegraphics[width=\linewidth]{figures/teaser.png}
%     % \vspace{-5mm}
%     \caption{Our neural network aims at eliminating critical points of the implicit representation~$f$. 
%     It can learn high-fidelity surfaces from unoriented raw point clouds. 
%     Besides, it can freely switch between SDFs and UDFs to enable open and multi-layers surface reconstruction.
%     % The two shapes of Happy Buddha and Dragon have about 30K points and are from Stanford 3DScanRepo~\cite{standford3d}.
%     % The close-up windows show the normal maps of the reconstructed shapes.
%     % \SQ{remember to cite}
%     %\ZX{To revise...}
%     % \ZX{May add a figure to show an open surface?}
%     }
%     \label{fig:teaser}
%     \vspace{-3mm}
% \end{figure}
\begin{figure}[!t]
\centering
\begin{overpic}[width=\linewidth]{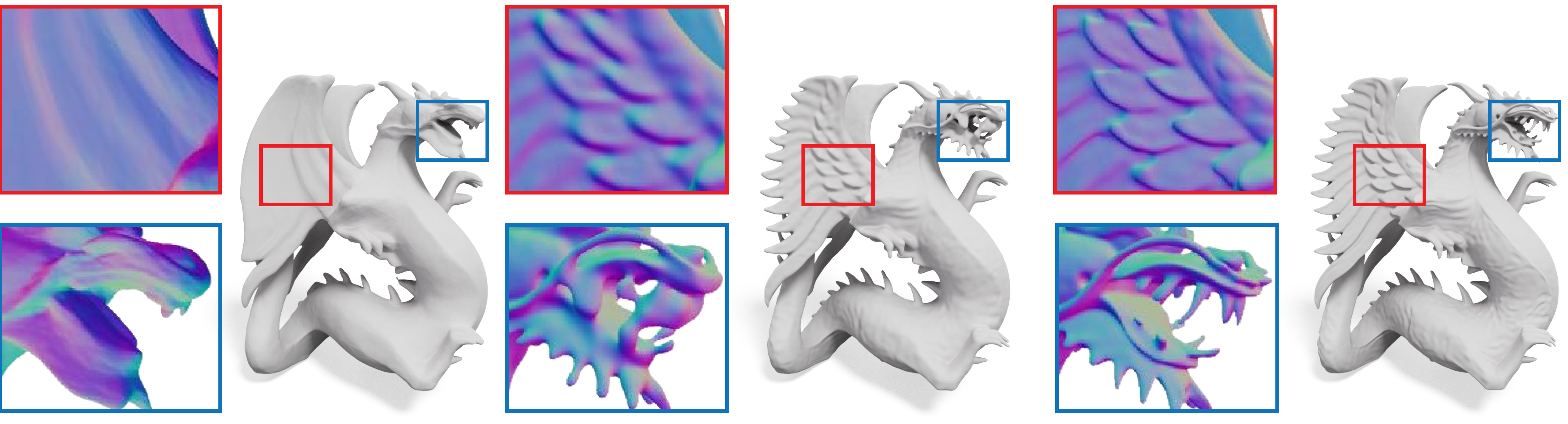}
    
    \put(10.5, -3.0){PCP~\shortcite{PCP}}

    \put(43, -3.0){DiGS~\shortcite{DiGS}}

    \put(79, -3.0){\textbf{Ours}}
\end{overpic}

 \caption{
    Our approach can reconstruct rich geometric details from unoriented point clouds while PCP~\cite{PCP} and DiGS~\cite{DiGS} cannot.
    }
    \label{fig:top_fig}

    \vspace{-5mm}
\end{figure}

Relevant research on reconstructing unoriented point clouds includes both traditional approaches~\cite{iPSR, PGR} and learning-based approaches~\cite{DeepSDF, ONet, Peng20ConvONet, Points2surf, POCO, SAL, NeuralPull, OnSurface, PCP, DiGS}. Traditional approaches typically involve repeatedly adjusting normals to be orthogonal to the zero isosurfaces of an implicit representation, such as a signed distance function~(SDF) or an occupancy field. 
% Relevant research works about reconstruction from an unoriented point cloud include not only traditional approaches~\cite{iPSR, PGR}
% but also learning-based approaches.
% Generally speaking, traditional approaches work by repeatedly 
% tuning the normals so that they are orthogonal to the zero isosurfaces of the implicit representation,
% e.g., the signed distance function (SDF) or the occupancy field.
Learning-based approaches can be further categorized into supervised and fitting-based methods.
Supervised learning approaches involve fitting data samples using ground-truth implicit representations as a guide~\cite{DeepSDF, ONet, Peng20ConvONet, Points2surf, POCO}.
However, these approaches may not generalize well to shapes or point distributions not present in the training set~\cite{sulzer2023dsr}.
Fitting-based approaches~\cite{SAL, NeuralPull, OnSurface, PCP, DiGS}, instead, employ different combinations of regularization terms to ensure specific properties of a signed distance function (SDF) in order to solve an optimization problem for each input point cloud, such as the Eikonal term~\cite{IGR} and the Laplacian energy term~\cite{DiGS}. While these approaches have strong generalization capabilities, their performance may be compromised when normals are unavailable.
Despite the use of the Eikonal term to suppress vanishing gradients, controlling the overall shape to adapt to the geometry and topology complexity of an input point cloud remains challenging. On the one hand, it is essential to accurately capture geometric details. On the other hand, unnecessary shape variations and ghost geometry must be eliminated. 
% Another type of learning-based methods, so-called overfitting methods~\cite{SAL, NeuralPull, OnSurface, PCP, DiGS}, utilize various combinations of regularization terms to enforce the neural function to own some desired properties from raw scans directly for the generalization ability, e.g., the Eikonal term~\cite{NeuralPull} and the Laplacian energy term~\cite{DiGS}. 
% Although the commonly used regularization terms work well for a high-quality point cloud (dense, equipped with normals), the performance is compromised in the pervasive situation where the normals are unavailable.

% \begin{figure*}[!t]
%     \centering
%     \includegraphics[width=\linewidth]{figures/iter.pdf}
%     \caption{ 
%     Starting with an unoriented point cloud (left), we iteratively optimize the network leading to a high-fidelity reconstruction.
%     Our coarse-to-fine method generates the coarse surface first and then captures the real topology gradually.
%     % The Hausdorff distance-based reconstruction errors are visualized by color mapping. 
%     }
%     \label{fig:iter}
% \end{figure*}

% Although the Eikonal term helps suppress vanishing gradients, controlling the overall shape to adapt to the input point cloud’s geometry/topology complexity remains challenging. On one hand, it is essential to capture geometric details accurately. On the other hand, unnecessary shape variations and ghost geometry must be eliminated. 
% In this paper, we address this problem using the shape operator from differential geometry\cite{?}.

In this paper, we address this problem based on the shape operator~\cite{shape_operator} from differential geometry.
Given a smooth surface, there must exist a narrow thin-shell space surrounding it where the signed distance function (SDF) is differentiable everywhere. For a point~$x$ in the thin-shell space whose projection onto the surface is~$x'$, 
%the gradient of the SDF exists and has a unit norm. 
the Hessian of the SDF at~$x$ has three eigenvectors, two of which align with principal curvature directions at~$x'$, and the other aligns with normal vector at~$x'$ with a corresponding eigenvalue of zero, making the Hessian singular.
In other words, by enforcing the Hessian to own a zero determinant for points near the surface, it helps align the gradient at~$x$ with the normal vector at~$x'$.
Based on this observation, we regularize the direction of the gradient of the SDF by enforcing the Hessian to own a zero determinant for points near the surface.%, rather than depend on a smoothness term. 

Making the Hessian singular differs from enforcing smoothness energy, such as Hessian energy~\cite{SSD, zhang2022critical} or Laplacian energy~\cite{DiGS}. The main difference is that the former can align the gradients of a near-surface point and its corresponding on-surface point, effectively suppressing surplus shape variations and adapting the implicit function to the inherent complexity of the input point cloud. Smoothness energy, on the other hand, tends to reduce the volatility of the implicit function so that the resulting surface is not overly complicated. Additionally, while the theoretical minimum value of smoothness energy cannot be zero~(otherwise, the implicit function degenerates to a globally constant or linear field), the determinant of the Hessian of the implicit function can be reduced as far as possible. Finally, the singular-Hessian constraint can effectively eliminate critical points of the target implicit function near the surface and avoid unnecessary variations~(e.g., ghost geometry). This can be explained by Morse theory~\cite{audin2014morse}, which reveals the deep link between the geometry and topology complexity of the surface and the number of critical points of the implicit function.

In implementation, we use sinusoidal activation~\cite{SIREN} to enable 
the computation of the first-order and the second-order derivatives of the implicit function.
% the punishment of non-degenerate critical points by second-order optimization.
Our approach first generates a rough but faithful shape by emphasizing the singular-Hessian constraint and then anneals the constraint to gradually capture fine details and real topology in a coarse-to-fine fashion.
% In the implementation, we use sinusoidal activation~\cite{SIREN} to enable the punishment of non-degenerate critical points by second-order optimization.
% The neural network naturally adapts itself to the real topological/geometric complexity encoded by the given point cloud (See Fig.~\ref{fig:teaser}). 
Extensive experimental results demonstrate that our approach can eliminate ghost geometry while remaining expressive enough to recover geometric details and sharp features. Compared to state-of-the-art methods, our approach has superior performance in both fitting a single unoriented point cloud and learning a shape space from a group of point data.
It can be seen from Fig.~\ref{fig:top_fig} that the reconstructed result by our approach has richer geometric details and higher fidelity than recently proposed methods such as PCP~\cite{PCP} and DiGS~\cite{DiGS}. 

\section{Related Work}
In recent decades, numerous surface reconstruction algorithms have been proposed~\cite{Jin20203DRU, Huangsurvey, sulzer2023dsr}. While most existing research assumes the presence of normals, there has been growing interest in surface reconstruction from unoriented point clouds. This section provides an overview of implicit surface reconstruction methods, including both traditional and learning-based approaches. 

% Numerous surface reconstruction algorithms~\cite{Jin20203DRU, Huangsurvey, sulzer2023dsr} have been proposed in the last decades. 
% Most of the existing research works assume the existence of normals,
% but surface reconstruction from unoriented point clouds has attracted increasing attention in recent years.
% In this section, we briefly review the implicit surface reconstruction methods, including traditional and learning-based methods.
%Also, we briefly review the applications of Morse theory for computer graphics.

\subsection{Traditional Implicit Methods}
The earliest implicit method
computes the signed distance to the tangent plane of the closest point~\cite{Hoppe1992SurfaceRF}.
After that, radial basis function~(RBF) based methods~\cite{carr2001reconstruction, li2016sparse, VIPSS} represent the underlying SDF as a weighted combination of radial basis kernels, resulting in higher smoothness.
Besides, the implicit moving least-square methods (IMLS)~\cite{shen2004interpolating, kolluri2008imls, oztireli2009RIMLS, schroers2014HessianIMLS}
approximates the underlying SDF by linearly blending local smooth planes.
The MPU method~\cite{MPU} models shape by blending piecewise quadratic functions that fit the local shape.
Poisson reconstruction~\cite{Kazhdan2006PoissonSR} and its variants~\cite{SPR, Kazhdan2020PoissonSR, Vizzo2021PoissonSR, sto_spr} formulate the occupancy field as the solution to Poisson's equation. 
The SSD method~\cite{SSD} computes the smooth signed distance by minimizing least-square style energy with a multi-grid solver.
Recently, iPSR~\cite{iPSR} iteratively runs the Poisson reconstruction solver to produce a reconstructed surface from an unoriented point cloud, using the normals of the reconstructed surface from the previous iteration as input for the next iteration. 
PGR~\cite{PGR} infers the occupancy field based on Gauss’s formula in potential theory by considering surface element areas and normals as unknown parameters.

\subsection{Learning-based Methods}
Learning-based approaches can be further categorized into supervised and fitting-based methods,
which are briefly introduced as follows.
\paragraph{Supervised Learning-based Reconstruction}
%Supervised learning-based reconstruction involves learning a signed distance function (SDF) or occupancy field using a dataset containing precomputed ground-truth field values. The neural model adjusts its weights as input data is fed into it until the model is appropriately fitted. Thanks to the data prior provided by the ground-truth data, these methods generally produce impressive reconstruction results. Early works encoded a global shape into a fixed-length latent code and recovered the underlying surface through a decoding operation [1-2]. While these methods can encode the overall representation for a group of similar shapes, they struggle to generalize from training examples to unseen shapes and to encode shape details with the global code. To capture the richness of geometry for generality, some research subdivides 3D space according to surface occupancy and encodes each part separately, using voxel grids [3-5], k-nearest neighbors [6-7], or octrees [8-10]. In the context of open surface reconstruction (e.g., clothes), unsigned distance field (UDF) based surface reconstruction has attracted increasing attention in recent years [11-12].

Supervised learning-based reconstruction involves learning the SDF or occupancy field using a dataset containing precomputed ground-truth field values.
The neural model adjusts its weights as input data is fed into it until the model is appropriately fitted. Thanks to the data prior provided by the ground-truth data, these methods generally produce impressive reconstruction results. 
Early works encode a global shape into a fixed-length latent code and recover the underlying surface through a decoding operation~\cite{DeepSDF, ONet}.
While these methods can encode the overall representation for a group of similar shapes, they struggle to generalize from training examples to unseen shapes and to encode shape details with the global code. To capture the richness of geometry for generality, some research subdivides 3D space according to surface occupancy and encodes each part separately,
using voxel grid~\cite{jiang2020local, DeepLS, Peng20ConvONet}, k-nearest neighbors~\cite{Points2surf, POCO}, or octrees~\cite{tang2021octfield, wang2022dual, Galerkin}.
In the context of open surface reconstruction (e.g., clothes), unsigned distance field (UDF) based surface reconstruction has attracted increasing attention in recent years~\cite{NDF, GIFS}.

\paragraph{Fitting-based Implicit Neural Representation}
To increase generalization ability, directly fitting the implicit representation from raw point clouds has been extensively studied in recent years, enabling end-to-end prediction of the target surface. Most methods apply different regularization techniques to accomplish this task. SAL/SALD~\cite{SAL, SALD} performs sign agnostic regression to obtain a signed version from the unsigned distance function.
IGR~\cite{IGR} directly applies the Eikonal term to encourage the predicted implicit function to have unit gradients.
Inspired by the definition of the SDF, Neural-Pull~\cite{NeuralPull} trains a neural network to predict the signed distance and gradients simultaneously so that a query point can be pulled to the closest point on the underlying surface.
Based on Neural-Pull, PredictableContextPrior~\cite{PCP} first trains a local context prior and then
specializes it in the predictive context prior to learning predictive queries at inference time.
OnSurfacePrior~\cite{OnSurface} improves the reconstruction quality for sparse point clouds with the help of a pre-trained unsigned distance network. Recently, CAP-UDF~\cite{CAP-UDF} extends Neural-Pull to learn UDFs directly from raw point clouds. 
SIREN~\cite{SIREN} shows that ReLU-MLPs bias toward low-frequency implicit representations, and then it introduces periodic activation functions to preserve high frequencies.
Based on SIREN, Iso-Points~\cite{yifan2020isopoints} constrains the hybrid representation grouped by explicit point clouds and implicit neural representation simultaneously to improve the reconstruction quality.
% IDF~\cite{IDF} decomposes 3D geometry into a smooth base surface and a continuous high-frequency implicit displacement field to recover details by offset. 
DiGS~\cite{DiGS} incorporates Laplacian energy as a soft constraint of the SDF to enable unoriented point cloud reconstruction for SIREN. 
In general, most existing approaches require oriented normals to produce high-quality results. However, the absence of normals may compromise their performance, resulting in over-smoothed shapes lacking rich geometric details.

\begin{figure}[!t]
    \centering
    \includegraphics[width=\linewidth]{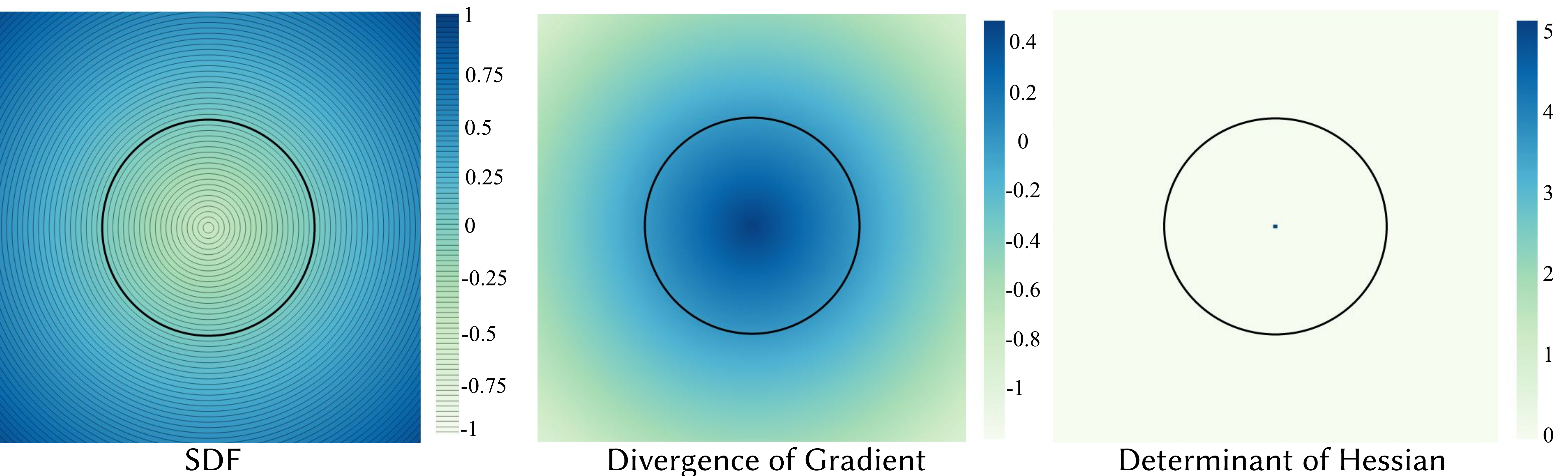}
    \caption{Visualizing the differential properties of an SDF for a circle. 
    In contrast to the divergence of the gradient (i.e., Laplacian energy) utilized by~\cite{DiGS}, the determinant of the Hessian matrix of the SDF remains consistently zero. This is due to the mathematical fact that, in the differentiable region near the surface, there exists a zero eigenvalue associated with an eigenvector aligned with the gradient.
    }
    \label{fig:2d_det}
\end{figure}
\section{PRELIMINARIES}
%\subsection{Implicit surfaces}
Given an unoriented point cloud~$\mathcal{P}$,
our goal is to find an implicit representation~${f}:\mathbb{R}^3 \mapsto \mathbb{R}$
such that the zero level set of~${f}$ accurately encodes the target surface~$\mathcal{S}$:
\begin{equation}
    \mathcal{S} = \left\{\boldsymbol{x} \in \mathbb{R}^3 \mid f(\boldsymbol{x})=0\right\}.
\end{equation}
The discrete surface representation can be extracted from~${f}$ using contouring algorithms.
It is important to note that our work requires~$f$ to be~$C^2$ continuous.
In previous research, it has been considered a suitable choice to encourage the neural implicit function~$f$  to approximate the signed distance function (SDF). To achieve this, $f$ must satisfy three boundary conditions:
(1) Dirichlet condition: $f(p) = 0$ for $p\in \mathcal{P}$, which encourages any given point to lie on the target surface.
(2) Eikonal condition: $\lVert \nabla f \rVert = 1$, which enforces $f$ to have a unit gradient norm or, at the very least, not to vanish as much as possible.
(3) Neumann condition: $\nabla f = \mathcal{N}$, which aligns the gradients with the normal field~$\mathcal{N}$ if the normals are available.
Existing implicit neural functions leverage the above boundary conditions as constraints, either explicitly~\cite{IGR, SIREN} or implicitly~\cite{NeuralPull, SAL}. 
In this paper, our method leverages the same network architecture as SIREN~\cite{SIREN}.
% explicit~\cite{IGR, SIREN} or implicit~\cite{NeuralPull, SAL} leverage the above boundary conditions as the constraints.
% Our method leverages the same network architecture of SIREN~\cite{SIREN}.

\begin{figure}[!t]
    \centering
    \includegraphics[width=\linewidth]{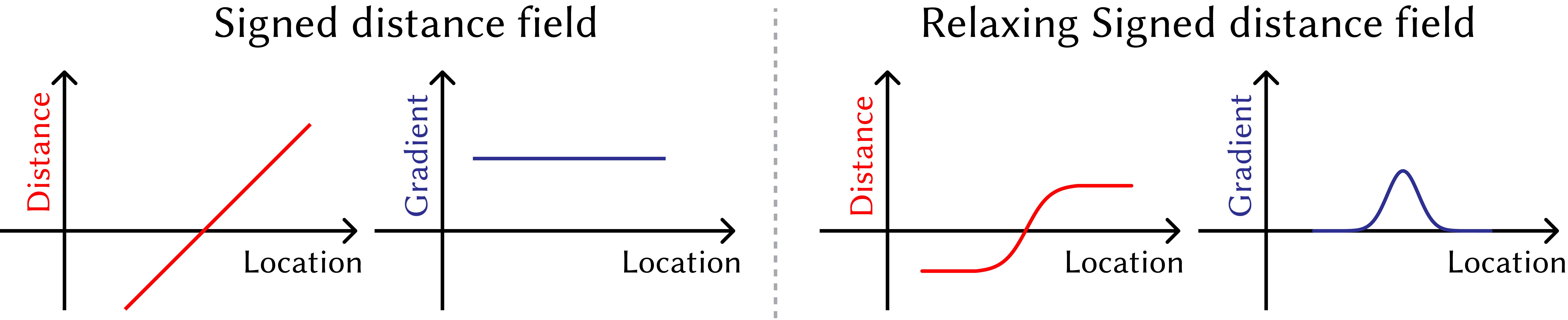}
    \caption{The illustration of the difference between the signed distance field and the relaxing signed distance field in terms of distance sign and gradient.}
    \vspace{-3mm}
    \label{fig:realx_sdf}
\end{figure}

For the point cloud~$\mathcal{P}$, 
let $\mathcal{Q}_\text{far}$ be a query point set uniformly sampled from its bounding box.
SIREN formulates these requirements into loss terms as below.
% \SQ{please left-align}
%Given point cloud $\mathcal{P}:\Omega\rightarrow\left[-1,1\right]$, SIREN defines the loss function as below:
\begin{align}
&L_{\text{manifold}} = \int_{\mathcal{P}} \lVert f(x) \rVert_1 dx
\\&L_{\text{non-manifold}} = \int_{\mathcal{Q}_\text{far}} \exp(-\alpha \lVert f(x) \rVert_1) dx, \alpha = 100 
\\&L_{\text{Eikonal}} = \int_{\mathcal{P} \cup \mathcal{Q}_\text{far}} \lVert \lVert \nabla f(x) \rVert_2 - 1 \rVert_1 dx
\\&L_{\text{Neumann}} = \int_{\mathcal{P}} (1 - \langle \nabla f(x), \mathcal{N}(x) \rangle) dx.
\end{align}
% where $\lVert \cdot \rVert_p$ is the $p$-norm and $\mathcal{Q}_\text{far}$ is a query point set uniformly sampled from the bounding box of $\mathcal{P}$.
The overall loss is thus given by 
\begin{equation}
\begin{split}
    L^{\text{oriented}}_\text{SIREN}=& \lambda_\text{manifold} L_{\text{manifold}} +  \lambda_\text{non-manifold} L_{\text{non-manifold}} + \\ &\lambda_\text{Eikonal} L_\text{Eikonal} + \lambda_\text{Neumann} L_{\text{Neumann}},
\end{split}
\end{equation}
where the four weights are respectively 3000, 100, 50, and 100.
If the normals are not available, the loss reduces to
\begin{equation}
\begin{split}
    L^{\text{unoriented}}_\text{SIREN}=& \lambda_\text{manifold} L_{\text{manifold}} +  \lambda_\text{non-manifold} L_{\text{non-manifold}} + \\ & \lambda_\text{Eikonal} L_\text{Eikonal}.
    %+ \\ & \lambda_\text{Neumann} L_{\text{Neumann}}.
\end{split}
\end{equation}
% In summary, it forms the SIREN loss with the weights $(\lambda_\text{Dirichlet}, \lambda_\text{Push}$, $\lambda_\text{Eikonal}, \lambda_\text{Neumann} )=\left(3000, 100, 50, 100\right)$:

% Although the loss terms of SIREN define meaningful constraints,
% SIREN is extremely difficult to optimize.
% As IDF~\cite{IDF} pointed out, SIREN may lead to many surplus parts even with given normals.
% The situation becomes even worse for unoriented inputs. 
% If $\mathcal{P}$ is of poor quality, 
% it is notoriously difficult to estimate the reliable normals/orientations~\cite{Rui2023GCNO}.
Despite the fact that the loss terms of SIREN establish meaningful constraints, optimizing SIREN presents a significant challenge. As indicated by IDF~\cite{IDF}, even when provided with normals, SIREN may generate numerous surplus parts. This problem is compounded when working with unoriented inputs. If the quality of the point cloud is poor, accurately estimating reliable normals or orientations can be notoriously difficult~\cite{Rui2023GCNO}.
In addressing the challenges associated with optimizing SIREN, it is important to recognize that there are an infinite number of candidate Eikonal solutions when enforcing the eikonal constraint at a limited number of sampled points, as discussed in DiGS~\cite{DiGS}. 
Only eikonal constraint may result in the neural network optimization becoming trapped in a local minimum with the emergence of ghost geometry, which is far removed from the optimal solution. 
Additionally, over-parametrized neural networks~\cite{sagun2017empirical} often possess a large number of parameters, which can further complicate the optimization process. Some approaches employ a form of smoothness energy, such as Dirichlet energy~\cite{phase}, Laplacian energy~\cite{DiGS} and Hessian energy~\cite{SSD, schroers2014HessianIMLS, zhang2022critical}, to guide the neural implicit function towards simplicity, enabling it to adapt to the inherent complexity of the geometry and topology of the given point cloud. However, enforcing smoothness can result in the loss of geometric detail.
This motivates us to define a new loss function that facilitates the identification of the optimal solution, rather than relying on smoothness energy.

\begin{figure}[!t]
    \centering
    \includegraphics[width=\linewidth]{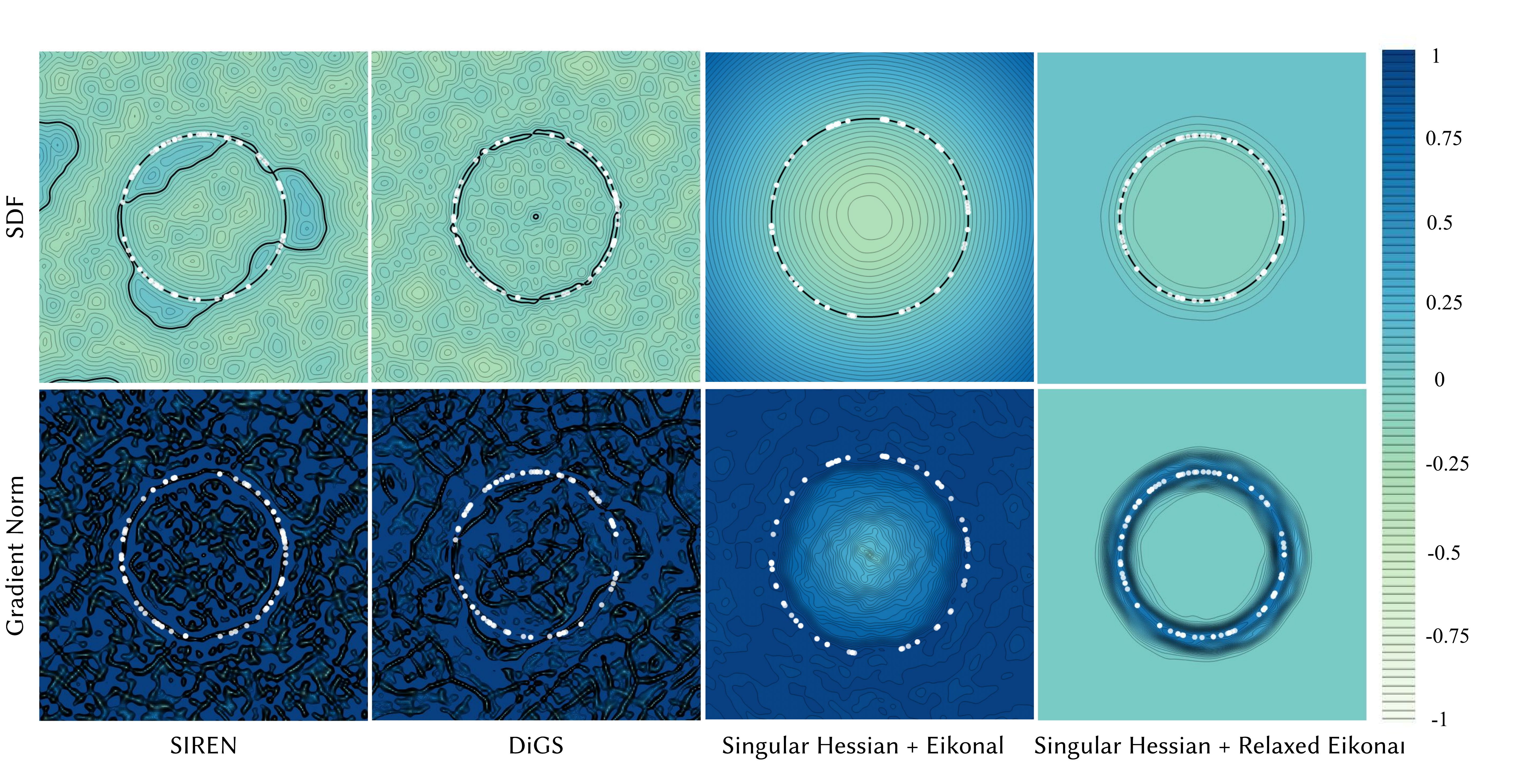}
    \caption{
    For SIREN~(only Eikonal) and DiGS~(Eikonal + Laplacian Energy), 
    the resulting neural implicit function has unnecessary critical points, leading to ghost geometry near the underlying circle (black). 
    Our singular-Hessian term can suppress critical points near the surface, thus allowing for high-fidelity reconstruction either combined with Eikonal loss or Relaxed Eikonal loss.
    It's worth noting that there are 100 data points (white).
    }
    \vspace{-3mm}
    \label{fig:2d_vis_siren}
\end{figure}
\section{Neural Singular Hessian}

% Existing methods often leverage smooth-based energy such as Direct Dirichlet energy~\cite{phase}, Laplacian energy~\cite{DiGS} and Hessian energy~\cite{SSD, schroers2014HessianIMLS, zhang2022critical}, to avoid the occurrence of ghost geometry.
% % However, utilizing smooth results directly is easy resulting in an over-smoothed result without rich geometric details. 
% This paper focuses on the other type of regularization terms, based on the following two observations.
% First, the unnecessary variations of ${f}$ lead to ghost geometry.
% And unnecessary variations about topology related to ghost geometry are affected by the critical points.
% Second, the smoothing energy easily results in an over-smoothed result without rich geometric details. 
% Hence, in this paper, the overall spirit of designing regularization terms lies in mitigating the redundant fluctuations through the suppressing critical points of the function ${f}$, instead of incorporating tricky smooth energy.
% Therefore, in this paper, the overall spirit of designing the regularization terms lies in suppressing the unnecessary variations by removing the non-degenerate critical points of ${f}$ with degeneration operation as much as possible rather than adding tricky smooth energy.

\subsection{Singular Hessian Term}

Given that the neural implicit function~${f}:\mathbb{R}^3 \mapsto \mathbb{R}$
is utilized to approximate the SDF and our primary interest lies in the zero level set of~$f$,
our focus is on learning~$f$ in the vicinity of the surface where the function value is close to~0, rather than approximating  the SDF everywhere. This is illustrated in Fig.~\ref{fig:realx_sdf}.
The real SDF may not be differentiable for a smooth surface at all points.
However, there must exist a narrow thin-shell space surrounding the surface within which the SDF is differentiable.
This differentiable region is denoted by~$\Omega$.
Consider a point~$x\in\Omega$ whose projection onto the surface is~$x'$.
The Hessian matrix of the SDF at~$x$, $\mathbf{H}_\text{SDF}(x)$, has a zero eigenvalue, with the gradient at~$x$, $\boldsymbol{g}_\text{SDF}(x)$, being the corresponding eigenvector. Additionally, the direction aligns with the normal vector at~$x'$. As a result, we have $\mathbf{H}_\text{SDF}(x) \boldsymbol{g}_\text{SDF}(x)  = \boldsymbol{0}$ and $\text{Det}(\mathbf{H}_\text{SDF}(x)) = 0$, which holds for any point in the differentiable region~$\Omega$, see Fig.~\ref{fig:2d_det}. This property can be easily derived with the Eikonal condition by differentiating both sides of the Eikonal equation $ \lVert \nabla f \rVert = 1$, refer to the reader~\cite{mayost2014applications}. This observation has led us to regularize the neural implicit function by enforcing a singular Hessian.

Suppose that~$f$ is $C^2$ smooth, its Hessian~$\mathbf{H}_f(x)$ is defined as the Jacobian of the gradient of~$f$:
\begin{equation}
\mathbf{H}_f(x)=\left[\begin{array}{lll}
f_{x x}(x) & f_{x y}(x) & f_{x z}({x}) \\
f_{y x}({x}) & f_{y y}({x}) & f_{y z}({x}) \\
f_{z x}({x}) & f_{z y}({x}) & f_{z z}({x})
\end{array}\right].
\end{equation}
In general,
it is desirable for
$f$ to approach the real SDF as accurately as possible, at least in~$\Omega$.
Therefore, 
it is necessary to enforce $\mathbf{H}_f(x) \boldsymbol{g}_\text{SDF}(x)  = \boldsymbol{0}$ or $\text{Det}(\mathbf{H}_f(x)) = 0$ for any point in~$\Omega$. 
We utilize $\text{Det}(\mathbf{H}_f(x)) = 0$ for points in close proximity to the surface,
since (1)~the ground-truth gradient~$\boldsymbol{g}_\text{SDF}(x)$ is unknown and (2)~$\mathbf{H}_f(x) \boldsymbol{g}_\text{SDF}(x)  = \boldsymbol{0}$
implies $\text{Det}(\mathbf{H}_f(x)) = 0$ but not vice versa.
%\ZX{A can infer B, B can not infer A. Is that mean we should use A to ensure both A and B are satisfied?}
% $\boldsymbol{g}_\text{SDF}(x)$ is not available, 
% we require $\text{Det}(\mathbf{H}_f(x)) = 0$ for points in close proximity to the surface. \ZX{Is that mean gt $g$ can not access? A question is why not use the gradient of $f$ as $\boldsymbol{g}_\text{SDF}(x)$} 
In implementation, we use $\mathcal{Q}_\text{near}$ to denote the query point set near the surface (sampling details will be given in Section~\ref{sec:exp}). The singular-Hessian loss is formulated as follows:
\begin{equation}
    L_\text{singularH} =  \int_{\mathcal{Q}_\text{near}} 
\lVert\text{Det}(\mathbf{H}_f(x))\lVert_1  dx.
\label{eq:critical}
\end{equation}

\begin{figure}[!t]
    \centering
    \includegraphics[width=.9\linewidth]{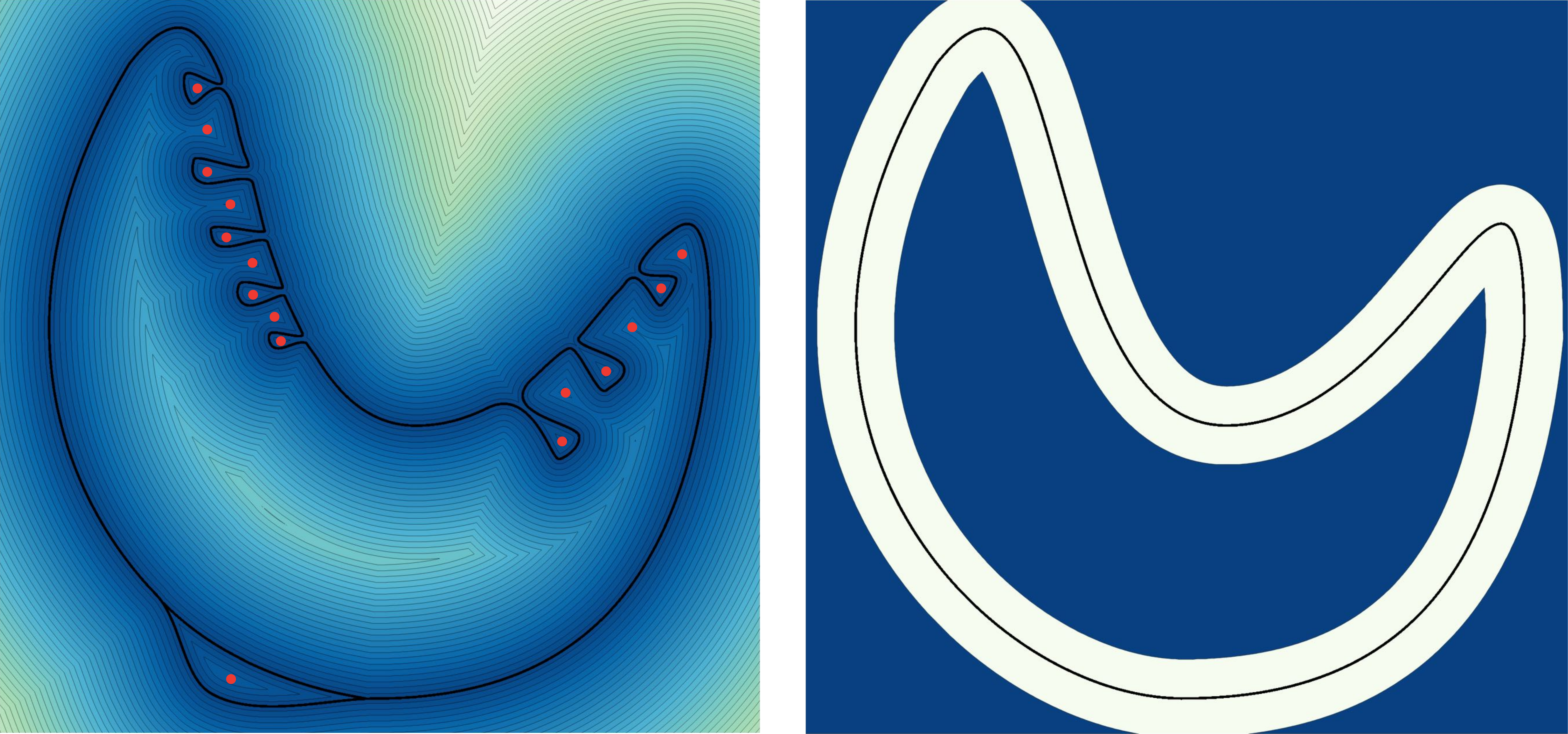}
    \caption{
    Unexpected critical points~(red points) near the surface, result in undesired surface variations~(left). To address this, we utilize the singular-Hessian term to suppress these variations and eliminate ghost geometry~(right).
    }
    \label{fig:morse}
    \vspace{-3mm}
\end{figure}
\subsection{How singular Hessian works}

\paragraph{Algebraic viewpoint.} According to the Taylor expansion,
\begin{equation}
    f(x) = f(x_0) +  \boldsymbol{g}_f^\text{T}(x-x_0) + \frac{1}{2}(x-x_0)^\text{T}\mathbf{H}_f(x)(x-x_0),
\label{eq:taylor}
\end{equation}
the utilization of  $\text{Det}(\mathbf{H}_f(x)) = 0$ permits variation in the second-order term~$\frac{1}{2}(x-x_0)^\text{T}\mathbf{H}_f(x)(x-x_0)$,
but with fewer degrees of freedom.
It is important to note that the Hessian energy is defined as $\|\mathbf{H}_f(x)\|_2^2$.
When the Hessian energy is reduced to 0, all entries of~$\mathbf{H}$ become 0, causing $f$ to become a linear field and diminishing its ability to accurately represent geometric details. In contrast to the Hessian energy, our singular Hessian term still allows~$f$ to be sufficiently expressive even if the term is reduced to~0. See Fig.~\ref{fig:2d_vis_siren} for an illustration. 

\paragraph{Morse theory.}
The points in~$\Omega$ (a thin-shell space surrounding the underlying surface) can be classified as either regular or critical, based on whether the gradient of function~$f$ vanishes.
The critical points can be further divided into 
(1) minimum points ($\mathbf{H}_f(x)$ does not have negative eigenvalues),
(2) 1-saddle points (one negative eigenvalue),
(3) 2-saddle points (two negative eigenvalues), and
(4) maximum points (three negative eigenvalues). 
Let the numbers be respectively
$c_\text{min}, c_\text{1-saddle}$, $c_\text{2-saddle}$, and $c_\text{max}$.
By taking $f$ as a Morse function defined in~$\Omega$, 
the Euler characteristic of~$\Omega$ is given by 
\begin{equation}
    \chi(\Omega) = c_\text{min} - c_\text{1-saddle} + c_\text{2-saddle}  - c_\text{max}.
\end{equation}
Unnecessary surface variations are generally caused by an overly complicated function~$f$ containing redundant critical points in~$\Omega$, see Fig.~\ref{fig:morse}.
The enforcement of $\text{Det}(\mathbf{H}_f(x)) = 0$ 
drives~$f$ toward simplicity by removing critical points in~$\Omega$, %\ZX{R4:the relationship between the critical points and the Hessian matrix is unclear.}
but has no side effects for regular points satisfying $\text{Det}(\mathbf{H}_f(x)) = 0$.
In other words,
$\text{Det}(\mathbf{H}_f(x)) = 0$
helps eliminate a minimum point and a 1-saddle point,
or a maximum point and a 2-saddle point,
or a minimum point and a maximum point,
or a 1-saddle point and a 2-saddle point,
until the minimum number of critical points exist and precisely conforms to the Euler characteristic of~$\Omega$.
In contrast, the enforcement of the Hessian energy 
has side effects on regular points and 
results in a linear field whose Euler characteristic may be independent of~$\Omega$.
Therefore, $\text{Det}(\mathbf{H}_f(x)) = 0$ is 
a more relaxed constraint that still preserves the topology of~$\Omega$.

\begin{table}[!htp]
\centering
\caption{Quantitative comparison on Surface Reconstruction Benchmark~\cite{DGP}. Note that the methods marked with ‘*’ require point normals.
% the methods marked with ‘$^+$’ are supervision based. 
In each column, the \under{best} scores are highlighted in bold with underline,
while the \textbf{second best} scores are highlighted with bold.}
\label{table:srb}
\resizebox{.9\linewidth}{!}{%
\begin{tabular}{l|cccc} 
\toprule
        & \multicolumn{2}{c}{Chamfer~$\downarrow$} & \multicolumn{2}{c}{F-Score~$\uparrow$}  \\
        & mean          & std.                 & mean           & std.               \\ 
\midrule
SPSR$^*$~\footnotesize{\cite{SPR}} & 4.36          & 1.56                & 75.87          & 18.57             \\ 
DGP$^*$~\footnotesize{\cite{DGP}} & 4.87 & 1.64 & 73.34 & 18.56 \\
\midrule
SIREN~\footnotesize{\cite{SIREN}}   & 18.24         & 17.09               & 38.74          & 31.26             \\
SAP~\footnotesize{\cite{SAP}}     & 6.19          & 1.75                & 57.21          & \under{11.66}    \\
iPSR~\footnotesize{\cite{iPSR}}    & 4.54          & 1.78                & 75.07          & 19.18             \\
PCP~\footnotesize{\cite{PCP}}     & 6.53          & 1.75                & 47.97          & 14.50             \\
CAP-UDF~\footnotesize{\cite{CAP-UDF}} & 4.54          & 1.82                & 74.75          & 18.84             \\
DiGS~\footnotesize{\cite{DiGS}}    & \textbf{4.16}          & \textbf{1.44}                & \textbf{76.69}          & 18.15             \\ 
\midrule
\textbf{Ours}    & \under{3.76} & \under{0.98}       & \under{81.38} & \textbf{13.73}             \\
\bottomrule
\end{tabular}
}
\end{table}

\begin{figure}[!htp]
    \centering
    \includegraphics[width=.99\linewidth]{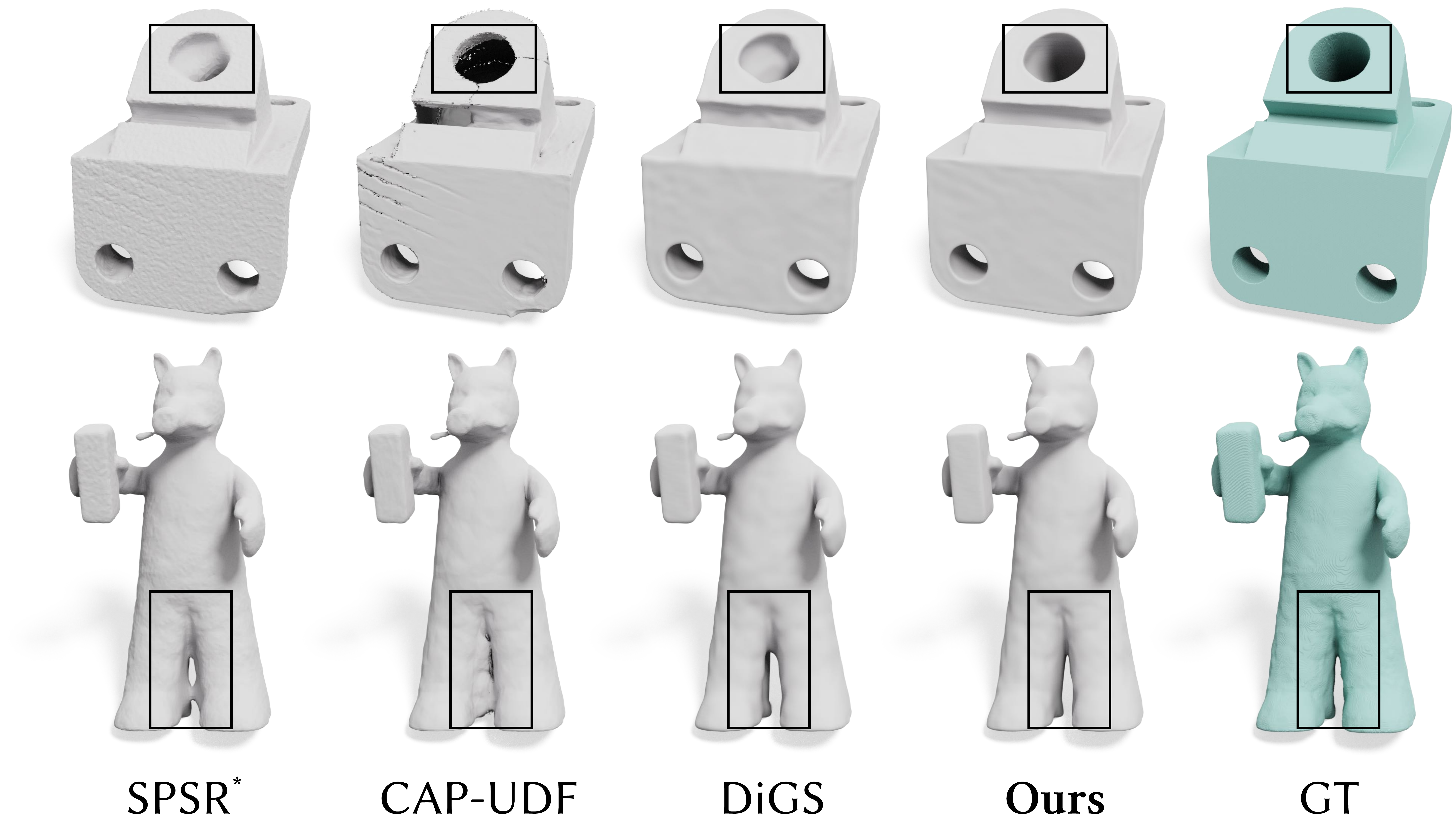}
    \caption{
    We select two point clouds from Surface Reconstruction Benchmark~\cite{DGP} and give a visual comparison
    among SPSR~\cite{SPR}, CAP-UDF~\cite{CAP-UDF}, DiGS~\cite{DiGS} and ours. 
    `*' means that the approach requires normals. 
    It can be seen from the highlighted differences 
    that our algorithm can produce a surface with high fidelity.
    % of     Qualitative results~(anchor and lord quas) of surface reconstruction on the Surface Reconstruction Benchmark~\cite{DGP}.
    % \SQ{to revise..}
    }
    \label{fig:srb}
\end{figure}

\begin{figure*}[!ht]
    \centering
    \includegraphics[width=.9\textwidth]{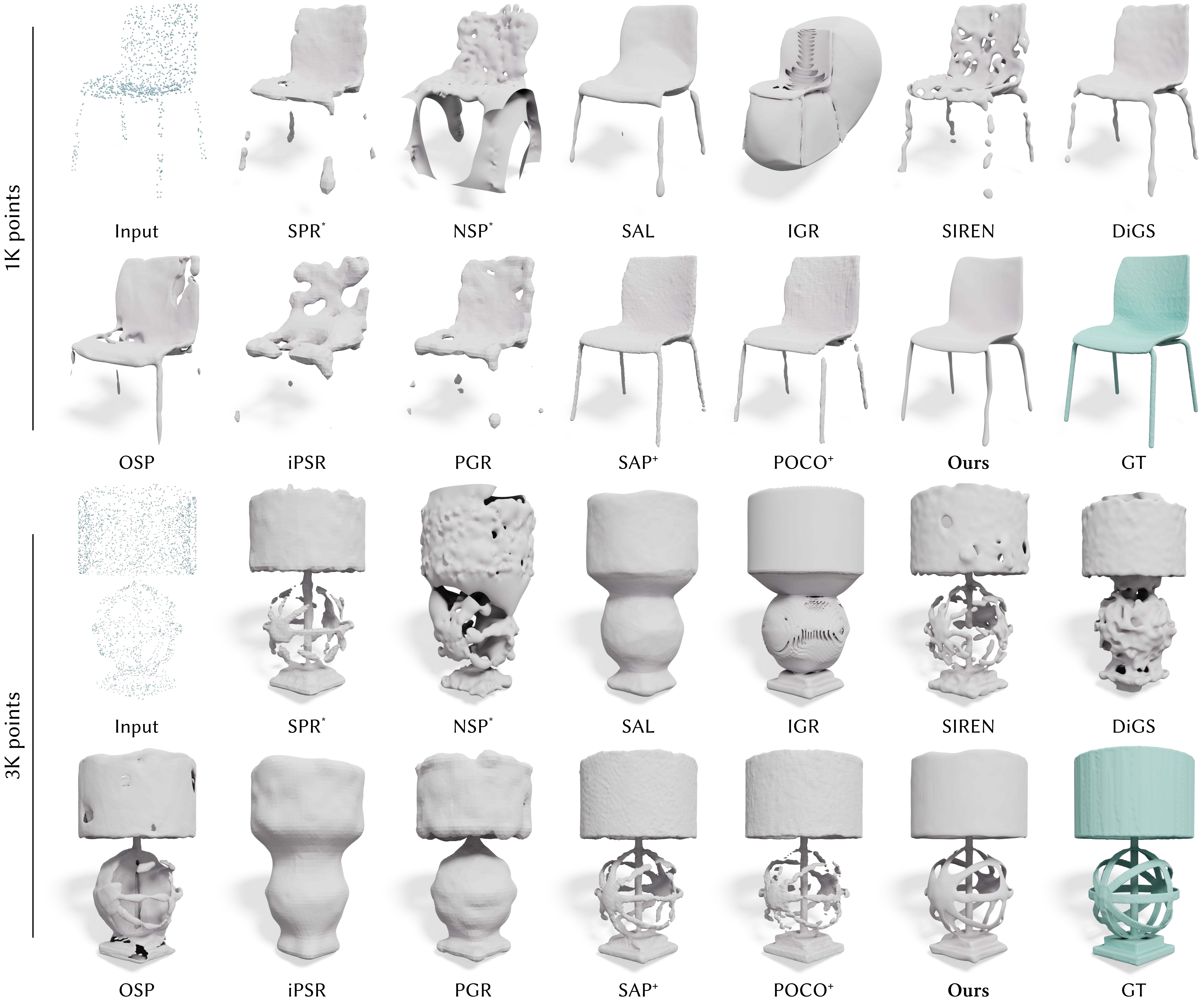}
    \caption{Visual comparison of surface reconstruction under ShapeNet~\cite{ShapeNet}. 
    The methods marked with `$^*$' require normals, 
    and the methods marked with `$^+$' are supervision-based. Our method can faithfully recover the thin plates/tubes
    even if the input point cloud is only with 1K points.
    }
    \label{fig:shapenet_main}
\end{figure*}

\subsection{Relaxing Eikonal Term}
The Eikonal equation, denoted as 
$\|\nabla {f}\|_2=1$, is commonly  used to characterize the 
 first-order property of the SDF. 
Several research works have mimicked this condition by requiring~$\|\nabla {f}\|_2=1$.
However, 
for the surface reconstruction problem, 
the focus is solely on the zero-level set of~$f$.
%whether~$f$ owns a unit gradient norm is a must.
It is sufficient to require that the gradient of~$f$ does not vanish near the surface. 
Therefore, we relax the Eikonal constraint of $\|\nabla {f}\|=1$ 
into $\|\nabla {f}\|>\sigma_{\min}$:
\begin{equation}
    L_{\text{Eikonal}}^\text{relax} = \int_{\mathcal{P}}\text{ReLU}\left(-(\|\nabla f(x)\| - \sigma_{\min})\right) dx,
    \label{eq:relax}
\end{equation}
where ReLU is the operator of $\max(0, \cdot)$ and $\sigma_{\min}$
represents the minimum gradient norm that must be retained by the field.
%is the minimum gradient norm the field must retain. 
We set $\sigma_{\min} = 0.8$ by default.
Additionally, the condition is enforced exclusively for points belonging to the input point cloud.

\noindent\textbf{Remark:} 
Although $f$ is intended to approach the real SDF,
they cannot be exactly identical.
Relaxing the Eikonal constraint accommodates a wider range of possible candidates, allowing for the identification of the most desirable solution. In summary, by relaxing this condition, other terms can play a more significant role, providing~$f$ with sufficient expressiveness.

% Also, shapes such as cylinders exist degenerate critical points that $\|\nabla {f}\| = 0$ in the field, hence forcing all points to satisfy $\|\nabla {f}\|=1$ may hinder the ability of the representation.
% The intention of the relaxation 
% lies in finding a non-degenerate implicit function to facilitate the extraction of zero level-set surfaces
% rather than insist on precisely the signed distance function. 
% The Fig.~\ref{fig:realx_sdf} shown the difference between SDF and Relaxing SDF.

% \subsection{Morse term}
% \label{subsec:Morse}

% \begin{definition}
% % \noindent{\bf Definition. }
% Given a closed and orientable surface $\mathcal{S}$,
% the {\em $\delta$-safe region} of $\mathcal{S}$
% is the $\delta$-wide shell in which there does not exist a singularity point.
% \end{definition}
% \begin{figure}[!t]
%     \centering
%     \includegraphics[width=\linewidth]{figures/2d_vis_critical_points.pdf}
%     \caption{
%     A 2D visual Comparison is made among SIREN~\cite{SIREN}, DiGS~\cite{DiGS}, and ours.
%     Our approach effectively reduces the number of unnecessary critical points through iterative processes, resulting in an accurate representation of the curve.
%     }
%     \label{fig:2d_vis}
% \end{figure}

\subsection{Total loss}
To this end, our total loss is formulated below:
\begin{equation}
    \begin{split}
        L_\text{ours}=& \lambda_\text{manifold} L_{\text{manifold}} +  \lambda_\text{non-manifold} L_{\text{non-manifold}}+ \\ &\lambda_\text{Eikonal}^{\text{relax}} L_\text{Eikonal}^{\text{relax}} +  \lambda_{\text{singularH}} \tau L_{\text{singularH}}.
    \end{split}
   \label{eq:total}
\end{equation}
where the parameter $\tau$ is an annealing factor,
allowing for a coarse-to-fine learning process that learns geometric details gradually. 
% The annealing mechanism is followed by DiGS~\cite{DiGS} with linear annealing:
% \begin{equation}
% \tau\left(t_0, t_1, \tau_1\right)= \begin{cases}1 & t<t_0 \\ 1+\left(\tau_1-1\right) \frac{\left(t-t_0\right)}{t_1-t_0} & t_0 \leq t \leq t_1 \\ \tau_1 & t>t_1\end{cases}.
% \end{equation}
% We refine the loss weights $\left(\lambda_\text{Dirichlet}, \lambda_\text{Push}, \lambda_\text{Eikonal}^{\text{relax}}, \lambda_\text{Morse} \right)=(7000, $600, $50, 3)$.
The annealing factor $\tau$ remains~1 during the first 20\% iterations, then linearly decreases to 0.0003 during the 20\% to 40\% iterations, and finally decreases to 0.00003 at the termination.
We tune weights to our preferred setting ($\lambda_\text{manifold}$= $7000$, $\lambda_\text{non-manifold}$ = $600$, $\lambda_\text{Eikonal}^\text{relax}$ = $50$) and use consistent hyperparameters over the different datasets.
% The annealing factor $\tau$ remains~1 during the first 20\% iterations, then linearly decreases to 0.0003 during the 20\% to 40\% iterations, and then gradually decreases to 0 at the final, that is $(t_0, t_1, \tau_1, t_2, \tau_2) = (0.2, 0.4, 0.0003, 1, 0)$.

\begin{table*}[!ht]
\centering
\caption{Quantitative comparison on ShapeNet~\cite{ShapeNet}.
Note that the methods marked with ‘*’ require point normals, and
the methods marked with ‘$^+$’ are supervision based. 
In each column, the \under{best} scores are highlighted in bold,
while the \textbf{second best} scores are highlighted in bold with underlining.
% In each column, the \under{best} scores are textbfd and highlighted in bold,
% while the \textbf{second best} scores are highlighted in bold but not textbfd.
% Results of the \under{best} and \textbf{second best} methods are highlighted in each column. 
}
\label{table:shapenet}
 \resizebox{.9\textwidth}{!}{%
\begin{tabular}{l|cccccc|cccccc} 
\toprule
\multicolumn{1}{c|}{} & \multicolumn{6}{c|}{1K points}                                                                                                & \multicolumn{6}{c}{3K points}                                                                                                 \\ 
\cmidrule{2-13}
\multicolumn{1}{c|}{} & \multicolumn{2}{c}{Normal C.~$\uparrow$} & \multicolumn{2}{c}{Chamfer~$\downarrow$} & \multicolumn{2}{c|}{F-Score~$\uparrow$} & \multicolumn{2}{c}{Normal C.~$\uparrow$} & \multicolumn{2}{c}{Chamfer~$\downarrow$} & \multicolumn{2}{c}{F-Score~$\uparrow$}  \\
\multicolumn{1}{c|}{} & mean           & std.                    & mean          & std.                     & mean           & std.                   & mean           & std.                    & mean          & std.                     & mean           & std                    \\ 
\midrule
SPSR$^*$~\footnotesize{\cite{SPR}}               & 91.89          & 4.76                    & 9.35          & 7.66                     & 46.91          & 26.06                  & 95.50          & 3.30                    & 4.66          & 4.64                     & 75.28          & 25.76                  \\
NSP$^*$~\footnotesize{\cite{NSP}}               & 87.05          & 6.05                    & 12.51         & 7.29                     & 36.17          & 21.56                  & 90.74          & 5.48                    & 8.85          & 6.96                     & 52.42          & 28.55                  \\ 
\midrule
SAL~\footnotesize{\cite{SAL}}                   & 82.99          & 11.11                   & 47.46         & 50.57                    & 18.16          & \textbf{19.15}                  & 86.69          & 9.66                    & 29.98         & 31.86                    & 25.76          & 22.43                  \\
IGR~\footnotesize{\cite{IGR}}                   & 79.26          & 12.27                   & 77.68         & 59.55                    & 22.48          & 32.08                  & 80.85          & 11.88                   & 62.54         & 48.44                    & 26.28          & 35.91                  \\
SIREN~\footnotesize{\cite{SIREN}}                 & 79.91          & 8.87                    & 38.04         & 46.02                    & 25.02          & 23.52                  & 83.79          & 10.20                   & 34.19         & 46.77                    & 32.34          & 30.13                  \\
DiGS~\footnotesize{\cite{DiGS}}                  & 92.67          & 6.03                    & 7.01          & 5.52                     & 58.72          & 29.77                  & 95.82          & 4.44                    & 4.59          & 4.94                     & 78.87          & 27.34                  \\
OSP~\footnotesize{\cite{OnSurface}}                   & 91.89          & 5.52                    & 8.77          & 6.76                     & 47.57          & 23.45                  & 94.73          & 3.94                    & 6.80          & 6.61                     & 59.12          & 25.82                  \\
iPSR~\footnotesize{\cite{iPSR}}                  & 87.88          & 7.26                    & 13.16         & 11.78                    & 39.36          & 25.11                  & 93.22          & 5.26                    & 5.95          & 5.97                     & 68.42          & 26.36                  \\
PGR~\footnotesize{\cite{PGR}}                   & 89.53          & 5.35                    & 10.49         & 6.52                     & 37.61          & \under{19.14}         & 91.90          & 4.93                    & 7.34          & 4.81                     & 51.44          & 23.12                  \\ 
\midrule
SAP$^+$~\footnotesize{\cite{SAP}}               & \textbf{94.92}          & \under{3.60}                    & 4.64          & \textbf{3.71}                     & 73.50          & 25.02                  & 96.33          & \textbf{3.24}                    & 3.75          & \textbf{4.16}                     & 84.79          & \textbf{20.94}                  \\
POCO$^+$~\footnotesize{\cite{POCO}}              & 94.79          & 4.15                    & \textbf{4.53}          & 4.05                     & \textbf{75.56}          & 26.46                  & \textbf{96.41}          & 3.53                    & \textbf{3.62}          & 4.21                     & \textbf{85.42}          & 23.13                  \\ 
\midrule
\textbf{Ours}                  & \under{95.10} & \textbf{4.04}           & \under{4.26} & \under{3.11}            & \under{80.51} & 21.48                  & \under{97.05} & \under{2.91}           & \under{3.08} & \under{2.64}            & \under{90.71} & \under{16.28}         \\
\bottomrule
\end{tabular}
}
    % \vspace{-2mm}
\end{table*}

\begin{figure*}[!ht]
    \centering
    \includegraphics[width=.98\textwidth]{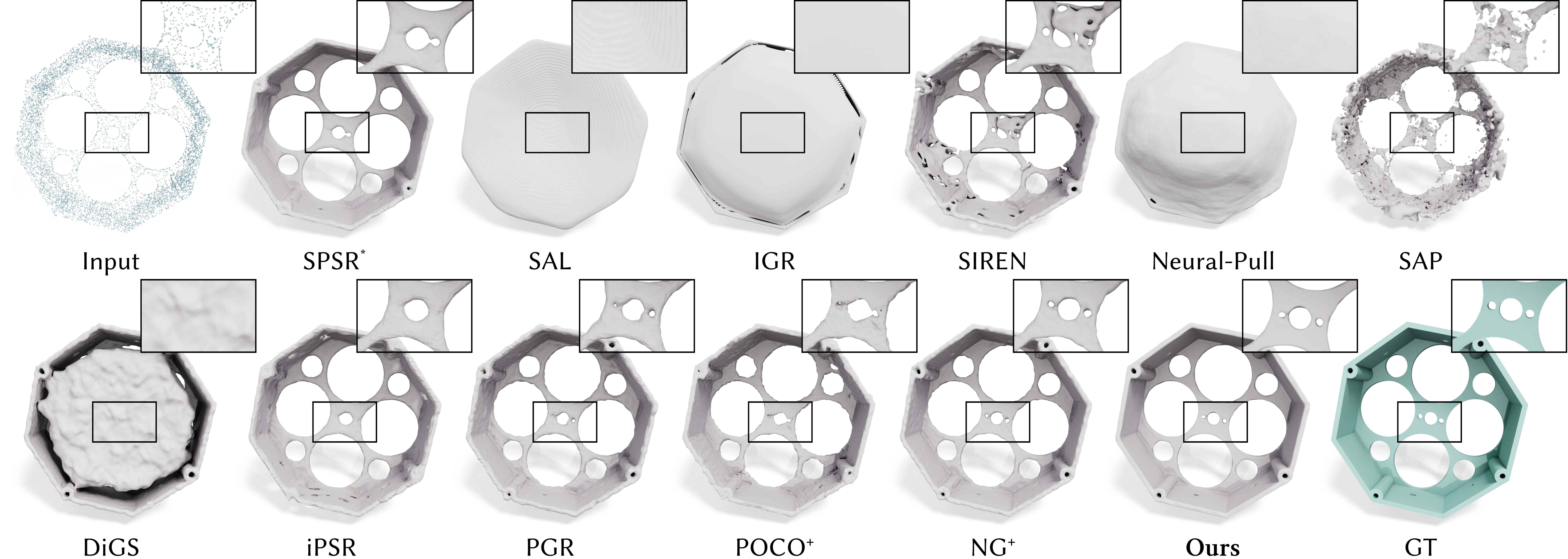}
    \caption{Visual comparison of surface reconstruction under ABC~\cite{ABC}. Our method can effectively recover the CAD features (e.g., small holes and thin plates).}
    \label{fig:abc_main}
\end{figure*}

\section{EXPERIMENTS}
\label{sec:exp}

In this section, the details regarding parameter selection are first presented, followed by an explanation of the metrics and evaluation protocol, more details can be checked in our supplementary material.
The proposed approach is then evaluated on various datasets, including both synthetic and real scans. Additionally, the approach is tested in shape space learning for human body scans.

% In this section, we first give the details about the choice of parameters.
% After that, we show the metrics and evaluation protocol. 
% Further, we evaluate the proposed approach
% on various datasets, including synthetical and real scans.
% We also test the proposed approach in shape space learning for human body scans.
% the detailed parameters and settings used in our experiments. 
% Our method is evaluated on multiple datasets with different types, i.e., watertight shapes, open cloth shapes, and shape space learning for the human body scans.
% More experimental results and details are available in the supplemental material.
%we provide full implementation details, extended results, ablations, additional experiments, and additional visualizations.

\subsection{Implementation Details} 
\label{subsec:Implementation}
Experiments were conducted using an NVIDIA GeForce RTX 3090 graphics card with 24GiB video memory and an AMD EPYC 7642.
A SIREN-based MLP with its initialization method was used for the network~\cite{SIREN}.
Inputs were first normalized to the range $\left[-1, 1\right]^3$ for SIREN-based MLP.
$\mathcal{Q}_\text{far}$ is uniformly sampled within the bounding box of the input point cloud $\mathcal{P}$, 
while $\mathcal{Q}_\text{near}$ is sampled following~\cite{IGR, NeuralPull, CAP-UDF} based on Gaussians distribution centered on each input point.
For a point in $p\in\mathcal{P}$, 
the Gaussian function is centered at $p_i$
with the mean and standard deviation set to the distance to its $k$-th~($k=50$ by default).
We sample one point for each distribution.
The number of points in both $\mathcal{Q}_\text{near}$ and $\mathcal{Q}_\text{far}$ complies with the batch size (15K by default).

\subsection{Metrics}
Comparison indicators include normal consistency, chamfer distances, and F-Score.
Normal consistency (expressed as a percentage and abbreviated as ‘Normal C.’) reflects 
the degree of agreement between the normals of the reconstructed surface and those of the ground-truth surface.
Chamfer distance (scaled by $10^3$ and using $L_1$-norm) measures the fitting tightness between the two surfaces, 
and F-Score~(expressed as a percentage) indicates the harmonic mean of precision and recall~(completeness).
The default threshold for F-Score is set to 0.005. All meshes are uniformly scaled to $\left[-0.5, 0.5\right]^3$, with 100K points sampled from each mesh for evaluation.

% is sampled from a sum of Gaussians centered at $\mathcal{P}$ with standard deviation equal to the distance to the kth nearest neighbor~(we used k = 50).
% the batch size
% The first examination point set $\mathcal{Q}_\text{far}$, is uniformly sampled within the bounding box of the input point cloud $\mathcal{P}$. 
% The query point set $\mathcal{Q}_\text{near}$ nearing the surface is sampled from a sum of Gaussians centered at $\mathcal{P}$ with standard deviation equal to the distance to the kth nearest neighbor~(we used k = 50).

\begin{table*}[!t]
\centering
\caption{Quantitative comparison on ABC~\cite{ABC} and Thingi10K~\cite{Thingi10K}. Each raw point cloud has 10K points.
The methods marked with ‘*’ require point normals, 
and the methods marked with ‘$^+$’ are supervision-based.
In each column, the \under{best} scores are highlighted in bold,
while the \textbf{second best} scores are highlighted in bold with underlining.
% In each column, the \under{best} scores are textbfd and highlighted in bold,
% while the \textbf{second best} scores are highlighted in bold but not textbfd.
}
\label{table:abc}
 \resizebox{.9\textwidth}{!}{%
\begin{tabular}{l|cccccc|cccccc} 
\toprule
\multicolumn{1}{c|}{} & \multicolumn{6}{c|}{ABC}                                                                                                      & \multicolumn{6}{c}{Thingi10K}                                                                                                 \\ 
\cmidrule{2-13}
\multicolumn{1}{c|}{} & \multicolumn{2}{c}{Normal C.~$\uparrow$} & \multicolumn{2}{c}{Chamfer~$\downarrow$} & \multicolumn{2}{c|}{F-Score~$\uparrow$} & \multicolumn{2}{c}{Normal C.~$\uparrow$} & \multicolumn{2}{c}{Chamfer~$\downarrow$} & \multicolumn{2}{c}{F-Score~$\uparrow$}  \\
\multicolumn{1}{c|}{} & mean           & std.                    & mean          & std.                     & mean           & std.                   & mean           & std.                    & mean          & std.                     & mean           & std                    \\ 
% \hline
\midrule
SPSR$^*$~\footnotesize{\cite{SPR}}               & 95.16          & 4.48                    & 4.39          & 3.05                     & 74.54          & 26.65                  & 97.15          & 2.95                    & 3.93          & 1.79                     & 77.03          & 23.71                  \\ 
\midrule
% \cmidrule{1-2}\cline{3-3}\cmidrule{4-4}\cline{5-7}\cmidrule{8-8}\cline{9-9}\cmidrule{10-10}\cline{11-11}\cmidrule{12-12}\cline{13-13}
SAL~\footnotesize{\cite{SAL}}                   & 86.25          & 8.39                    & 17.30         & 14.82                    & 29.60          & 18.04                  & 92.85          & 5.01                    & 13.46         & 7.97                     & 27.56          & 14.64                  \\
IGR~\footnotesize{\cite{IGR}}                   & 82.14          & 16.12                   & 36.51         & 40.68                    & 43.47          & 40.06                  & 90.20          & 10.61                   & 27.80         & 34.8                     & 54.28          & 39.99                  \\
SIREN~\footnotesize{\cite{SIREN}}                 & 82.26          & 9.24                    & 17.56         & 15.25                    & 30.95          & 22.23                  & 88.30          & 6.53                    & 17.69         & 13.47                    & 26.20          & 19.74                  \\
Neural-Pull~\footnotesize{\cite{NeuralPull}}           & 94.23          & 4.57                    & 6.73          & 5.15                     & 42.67          & \under{10.75}         & 96.15          & 2.80                    & 5.89          & 1.12            & 46.44          & \under{8.53}          \\
SAP~\footnotesize{\cite{SAP}}                   & 81.59          & 10.61                   & 15.18         & 16.60                    & 45.88          & 33.67                  & 92.60          & 7.03                    & 10.61         & 13.84                    & 53.32          & 31.91                  \\
DiGS~\footnotesize{\cite{DiGS}}                 & 94.48          & 6.12                    & 6.91          & 6.94                     & 66.22          & 32.01                  & 97.25          & 3.30                    & 5.36          & 5.59                     & 74.45          & 27.11                  \\
iPSR~\footnotesize{\cite{iPSR}}                  & 93.15          & 7.47                    & 4.84          & 4.06                     & 71.59          & 24.96                  & 96.46          & 3.57                    & 4.41          & 2.94                     & 74.88          & 22.72                  \\
PGR~\footnotesize{\cite{PGR}}                   & 94.11          & 4.63                    & 4.52          & 2.13                     & 68.91          & 27.86                  & 96.80          & 3.25                    & 4.22          & 2.01                     & 72.86          & 22.98                  \\ 
% \hline
\midrule
POCO$^+$~\footnotesize{\cite{POCO}}              & 92.90          & 7.00                    & 6.05          & 6.80                     & 68.29          & 26.05                  & 95.16          & 5.00                    & 5.61          & 9.42                     & 73.92          & 25.79                  \\
NG$^+$~\footnotesize{\cite{Galerkin}}   & \textbf{95.88}       &  \textbf{3.88}              &  \textbf{3.60}                       &      \under{1.38}         &    \textbf{81.38}                & 20.39               &   \textbf{97.71}                     & \textbf{2.68}               &   \textbf{3.16}                      &  \under{1.07}             &  \textbf{86.29}                        &      17.39                                 \\ 

\midrule
% \cmidrule{1-2}\cline{3-3}\cmidrule{4-4}\cline{5-5}\cmidrule{6-6}\cline{7-7}\cmidrule{8-8}\cline{9-9}\cmidrule{10-10}\cline{11-11}\cmidrule{12-12}\cline{13-13}
\textbf{Ours}                  & \under{97.42} & \under{2.37}           & \under{3.27} & \textbf{1.78}            & \under{88.62} & \textbf{13.87}                  & \under{98.23} & \under{1.98}           & \under{3.00} & \textbf{2.62}                     & \under{93.50} & \textbf{11.84}                  \\
\bottomrule
\end{tabular}
}
\end{table*}

\begin{figure*}[!htp]
    \centering
    \includegraphics[width=\textwidth]{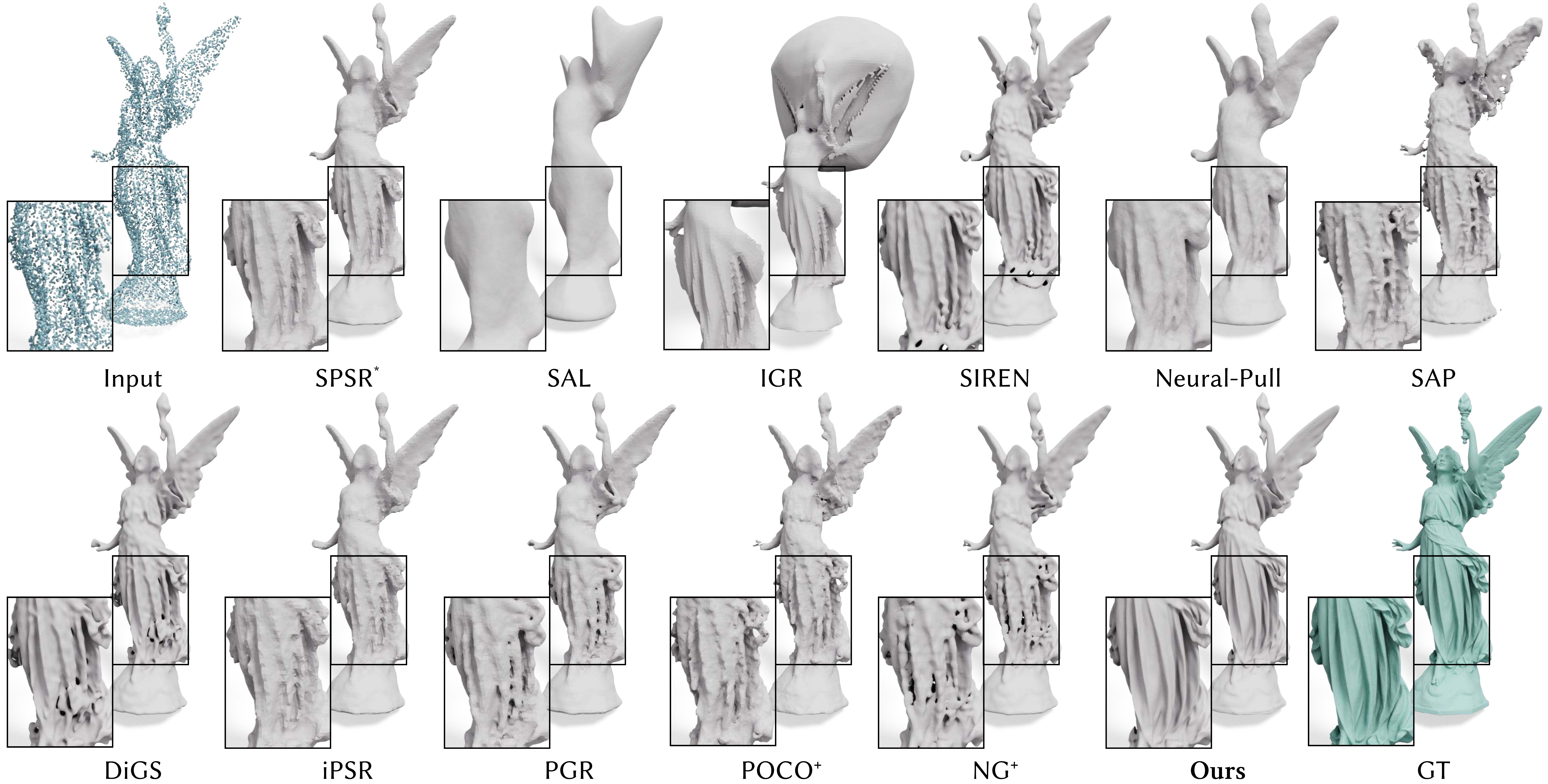}
    \caption{Visual comparison of surface reconstruction under Thingi10K~\cite{Thingi10K}. Our method can recover 
    high-fidelity geometric details.}
    \label{fig:thingi_main}
\end{figure*}

\subsection{Overfitting Surface Reconstruction}
A SIREN network consisting of 4 hidden layers with 256 units was used to conduct experiments. The discrete mesh of the zero-level set of the implicit function was extracted using the marching cubes algorithm~\cite{lewiner2003efficient} from scikit-image~\cite{scikit-image} with $256^3$ grids. For overfitting experiments, the Adam optimizer~\cite{Adam} was used  with a learning rate of $5\times10^{-5}$ and a total of 10K iterations by default.

\subsubsection{Surface Reconstruction Benchmark (SRB)}
The Surface Reconstruction Benchmark (SRB)~\cite{DGP} comprises five shapes, 
each with challenging features such as missing parts and rich details.
Approaches for comparison
include screened Poisson surface reconstruction (SPSR)~\cite{SPR}, SIREN~\cite{SIREN}, DGP~\cite{DGP}, Shape as points~(SAP)~\cite{SAP}, iPSR~\cite{iPSR}, Predictive Context Priors~(PCP)~\cite{PCP}, CAP-UDF~\cite{CAP-UDF} and DiGS~\cite{DiGS}.
It should be noted that SPSR and DGP leverages normals provided by the input scans.
%Note that SPSR leverages normals given by the input scans.
% For baselines, we choose the screened Poisson surface reconstruction (SPSR)~\cite{SPR}, SIREN~\cite{SIREN}, Shape as points~(SAP)~\cite{SAP}, iPSR~\cite{iPSR}, Predictive Context Priors~(PCP)~\cite{PCP}, CAP-UDF~\cite{CAP-UDF}, DiGS~\cite{DiGS}.
% The input is already with normals, and we provide it for SPSR.
As demonstrated  in Tab.~\ref{table:srb} and Fig.~\ref{fig:srb}, our method outperforms existing methods in terms of both Chamfer distance and F-score. 
In particular, the visual comparison in Fig.~\ref{fig:srb}
shows that our method can recover the hole feature of the Anchor model despite the lack of points on the inner wall
and preserve the nearby gaps of the Lord Quas model.
% By contrast, 
% Compared with other methods, our results effectively recover the features~(holes) of the ``anchor'' and do not suffer from shallow gap issues that wrongly close gaps.

\begin{figure}[!t]
    \centering
    \includegraphics[width=\linewidth]{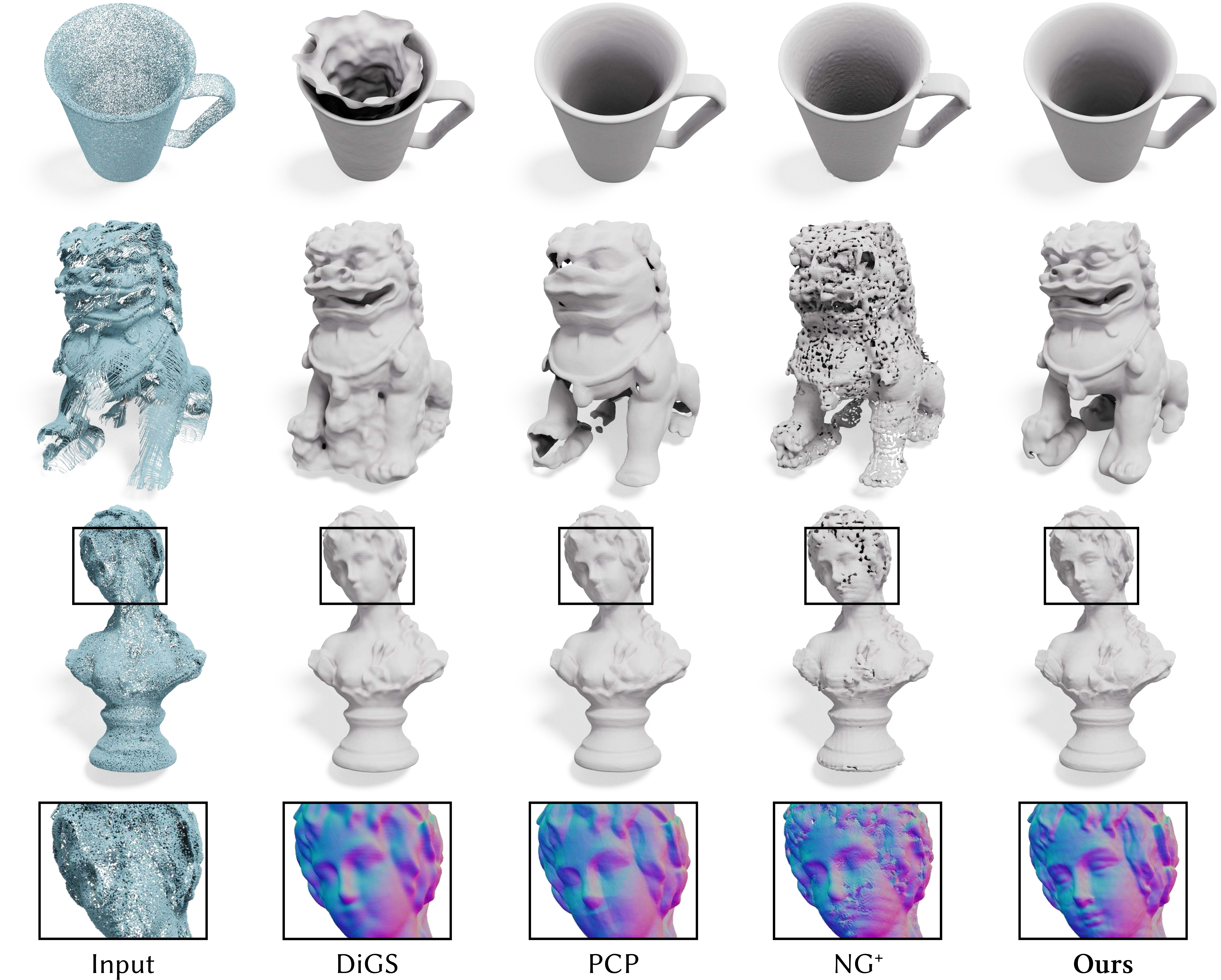}
    \caption{
    Visual comparison of real raw scans from~\cite{Huangsurvey} and Aim@Shape Shape Repository.
    Our method can deal with real scans and recover the details.
    }
    \label{fig:real_main}
\end{figure}

\subsubsection{ShapeNet}
The ShapeNet~\cite{ShapeNet} comprises  a diverse range of CAD models.
We follow the splitting of~\cite{NSP} for the 13 categories of shapes with totally 260 shapes.
% which of~\cite{NSP}, who evaluate 20 shapes in each of 13 categories and preprocess to make the normals consistent and internal structure to be manifold meshes. 
% In each category, 20 shapes are manifold and own the normal consistency.
We perform the comparison under two settings, i.e., ``1K points'' and ``3K points''.
Baseline approaches include the screened Poisson surface reconstruction~(SPSR)~\cite{SPR}, NSP~\cite{NSP}, SAL~\cite{SAL}, IGR~\cite{IGR}, SIREN~\cite{SIREN}, DiGS~\cite{DiGS}, OnSurfacePrior~(OSP)~\cite{OnSurface}, iPSR~\cite{iPSR} and PGR~\cite{PGR}.
 It should be noted that SPSR and NSP need to input normals. 
To make the comparison more convincing, we provide ground-truth normals to those methods that require normals.
Additionally, learnable baselines such as Points~(SAP)~\cite{SAP} and POCO~\cite{POCO} are included for comparison and re-trained from scratch on the ShapeNet dataset with 1K points and 3K points settings,  respectively. 
% They are re-trained from scratch.
% and learnable baselines including Shape as Points~(SAP)~\cite{SAP} and POCO~\cite{POCO}.
% Since SPSR and NSP need the input equipped with oriented normals, we provide the ground-truth normal.
% All learnable baselines are re-trained from scratch.
% The comparisons are shown in Tab.~\ref{table:shapenet} and visualized in Fig.~\ref{fig:shapenet_main}. \ZX{Should cite this figure?}
Based on the quantitative comparison in Tab.~\ref{table:shapenet}
and the visual comparison in Fig.~\ref{fig:shapenet_main},
 it is evident that the majority of existing methods are unable to effectively handle data sparsity. This issue is particularly pronounced when thin-walled plates and tubes are present, as it significantly increases the difficulty of reconstruction. In contrast, our method is able to effectively suppress unnecessary variations near the surface and adapt the implicit representation to the inherent complexity encoded by the point cloud. Furthermore, our reconstruction accuracy surpasses even that of supervision-based methods SAP and POCO.
% Note that SAP and POCO infer the implicit fields based on the local prior.
% Without learnable modules, most methods can not yield reliable results with sparse input, including SPSR and NSP, which leverage the ground-truth normals.
% However, our method effectively suppresses unnecessary variations leading to faithful results.
% Moreover, our reconstruction accuracy even surpasses the supervision methods SAP and POCO, where the implicit fields are inferred with the local prior.

\subsubsection{ABC and Thingi10K}
The ABC dataset~\cite{ABC} comprises a diverse collection of CAD meshes, while the Thingi10K dataset~\cite{Thingi10K} contains a variety of shapes with intricate geometric details.
% highly-detailed geometrical shapes.
% In this section, we conduct experiments under the ABC~\cite{ABC} dataset and Thingi10K~\cite{Thingi10K} dataset. 
% ABC dataset consists of a large number and variety of high-quality CAD meshes, and Thingi10K consists of highly-detailed geometrical shapes.
% We use the split of~\cite{Points2surf} and random sample 10K points from the meshes.
We follow~\cite{Points2surf} to perform splitting that 100 shapes for each dataset and randomly sample 10K points from each mesh. 
The baseline approaches include the screened Poisson surface reconstruction (SPSR)~\cite{SPR}, SAL~\cite{SAL}, IGR~\cite{IGR}, SIREN~\cite{SIREN}, Neural-Pull~\cite{NeuralPull}, SAP~\cite{SAP}, DiGS~\cite{DiGS}, iPSR~\cite{iPSR} and PGR~\cite{PGR}.
It is important to note that we find the supervision version of SAP does not generalize well on the shape not present on the trainset~(ShapeNet) when using the global PointNet-based encoder.
As such, we compare against the unsupervised version of SAP.
Additionally, 
We include supervision methods POCO~\cite{POCO} and Neural Galerkin~(NG, the version without normals)~\cite{Galerkin} for comparison.
In order to assess the generalization capabilities of the supervised methods, we retrained them using a setting of 10K points on the ShapeNet dataset.
%and tested them under ABC and Thingi10K datasets.
% and learnable baselines including POCO~\cite{POCO} and Neural Galerkin~(NG, the version without normal)~\cite{Galerkin}.
% We provide ground-truth normals for SPSR.
% To validate the generalization ability of the supervision methods, we re-trained them with 10K points under ShapeNet and tested them under ABC and Thingi10K datasets.
The quantitative comparison statistics are recorded in Tab.~\ref{table:abc}.
%while the visual comparison is available in Fig.~\ref{fig:abc_main} and Fig.~\ref{fig:thingi_main}, respectively. 
% Results are compared in Tab.~\ref{table:abc} and visualized in Fig.~\ref{fig:abc_main} and Fig.~\ref{fig:thingi_main}.
% We could recover highly-detailed geometries such as holes, clothes, and thin concave structures.
Furthermore, visual comparisons conducted using the ABC dataset~\cite{ABC} (as shown in Fig.~\ref{fig:abc_main}) demonstrate that our method is capable of effectively recovering CAD features such as small holes and thin plates. Similarly, visual comparisons conducted using the Thingi10K dataset~\cite{Thingi10K} (as shown in Fig.~\ref{fig:thingi_main}) demonstrate that our method is capable of recovering high-fidelity geometric details.
% \subsubsection{Large Scene} 
% We train on the scene from SIREN~\cite{SIREN} and change the network to 8 layers with 512 units to adopt scene-level shape.
% Qualitative results for scene reconstruction are presented in Fig.~\ref{fig:scene}. 
% When training SIREN without normal supervision, we observe many ghost geometries (SIREN wo n). On the other hand, DiGS can reconstruct the scene without these, producing a smooth result. This is desirable in most planar regions, e.g., ceiling, floor, and table, however, this trait has the drawback of smoothing out fine details, e.g., sofa legs and picture frame.

\subsubsection{Real Scans}
We also evaluated our method on real scans with various artifacts from~\cite{Huangsurvey} and AIM@SHAPE-VISIONAIR. The point clouds from~\cite{Huangsurvey} were scanned using a SHINING 3D Einscan SE scanner and exhibit noise and non-uniform densities. The point clouds from AIM@SHAPE-VISIONAIR were scanned using a Kreon scanner and exhibit highly non-uniform line distributions and unnatural scanner noise. We included DiGS~\cite{DiGS}, PCP~\cite{PCP}, and supervised method Neural Galerkin~(NG, without normals)~\cite{Galerkin} as baselines for comparison. 
Qualitative results can be seen in Fig.~\ref{fig:real_main}. Our method effectively recovers the details and concave parts of the shapes, while other methods do not.
In particular, the supervised method Neural Galerkin performs poorly when applied to real scans that have point distributions not present in its training dataset~(as shown in the second row of Fig.~\ref{fig:real_main}).
% \subsubsection{Sparse LiDAR Point Clouds}
% Common sparse point clouds are scanned by LiDAR.
% We leverage the data from KITTI~\cite{KITTI} to conduct the experiments.
% We use point clouds in single frames to conduct a comparison to DiGS~\cite{DiGS} and another sparse-oriented method OSP~\cite{OnSurface}.

\subsubsection{Large Scans}
\begin{figure*}[!t]
    \centering
    \includegraphics[width=\textwidth]{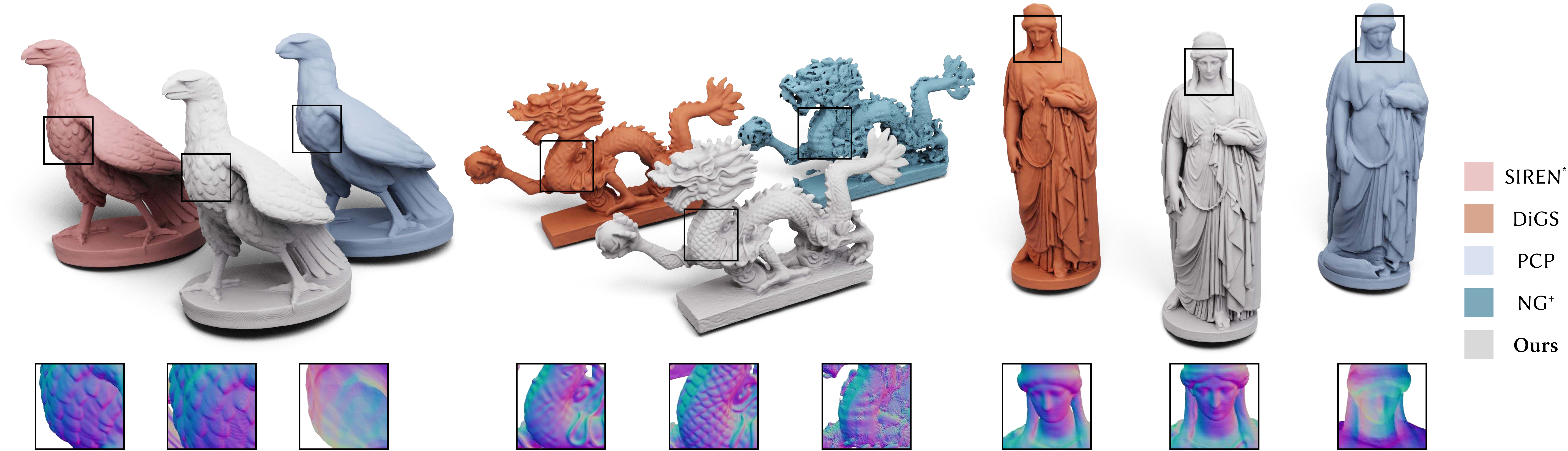}
    \caption{Tests on large scans~(about 30K points) from~ ThreedScans~\cite{three_scans}~(CC BY-NC-SA). 
    We include SIREN~\cite{SIREN}~(with ground-truth normals), DiGS~\cite{DiGS}, PCP~\cite{PCP} and NG~\cite{Galerkin} (supervision-based) for comparison. 
    % Our method is compared to SIREN~\cite{SIREN}~(with ground-truth normals), DiGS~\cite{DiGS}, PCP~\cite{PCP} and supervision method NG~\cite{Galerkin}. 
    In contrast to the SOTA methods,
    our approach can recover high-fidelity geometry details 
    from unoriented point clouds,
    and our results are comparable to SIREN trained with ground-truth normals.
    }
    \label{fig:large}
\end{figure*}

We evaluated the ability of our method to handle large-sized point clouds using three shapes from the ThreedScans dataset~\cite{three_scans}, randomly sampling approximately 300K points from each shape. For comparison, we include SIREN~\cite{SIREN} with ground truth oriented normals, PCP~\cite{PCP}, DiGS~\cite{DiGS}, and the learnable method of Neural Galerkin (NG, without normals)~\cite{Galerkin}. For this experiment, the final mesh is extracted at a resolution of $512^3$ rather than $256^3$.  Additionally, we sampled 1M points to analyze the distance between two surfaces and set the F-Score threshold to 0.001 (denoted as F-Score$^\vartriangle$). Quantitative comparison statistics are provided in Tab.~\ref{table:large}, while visual comparisons can be seen in Fig.~\ref{fig:large}. It is important to note that we trimmed surplus parts from SIREN’s results. PCP and DiGS tend to produce smooth results without geometric details, with DiGS’ smoothing energy weakening geometric details. Neural Galerkin is a supervised method and thus exhibits weaker generalization capabilities. In summary, for large-sized point clouds, our method outperforms existing unoriented approaches and is even superior to the ``with normals'' version of SIREN.

\begin{figure}[!htp]
    \centering
    \includegraphics[width=\linewidth]{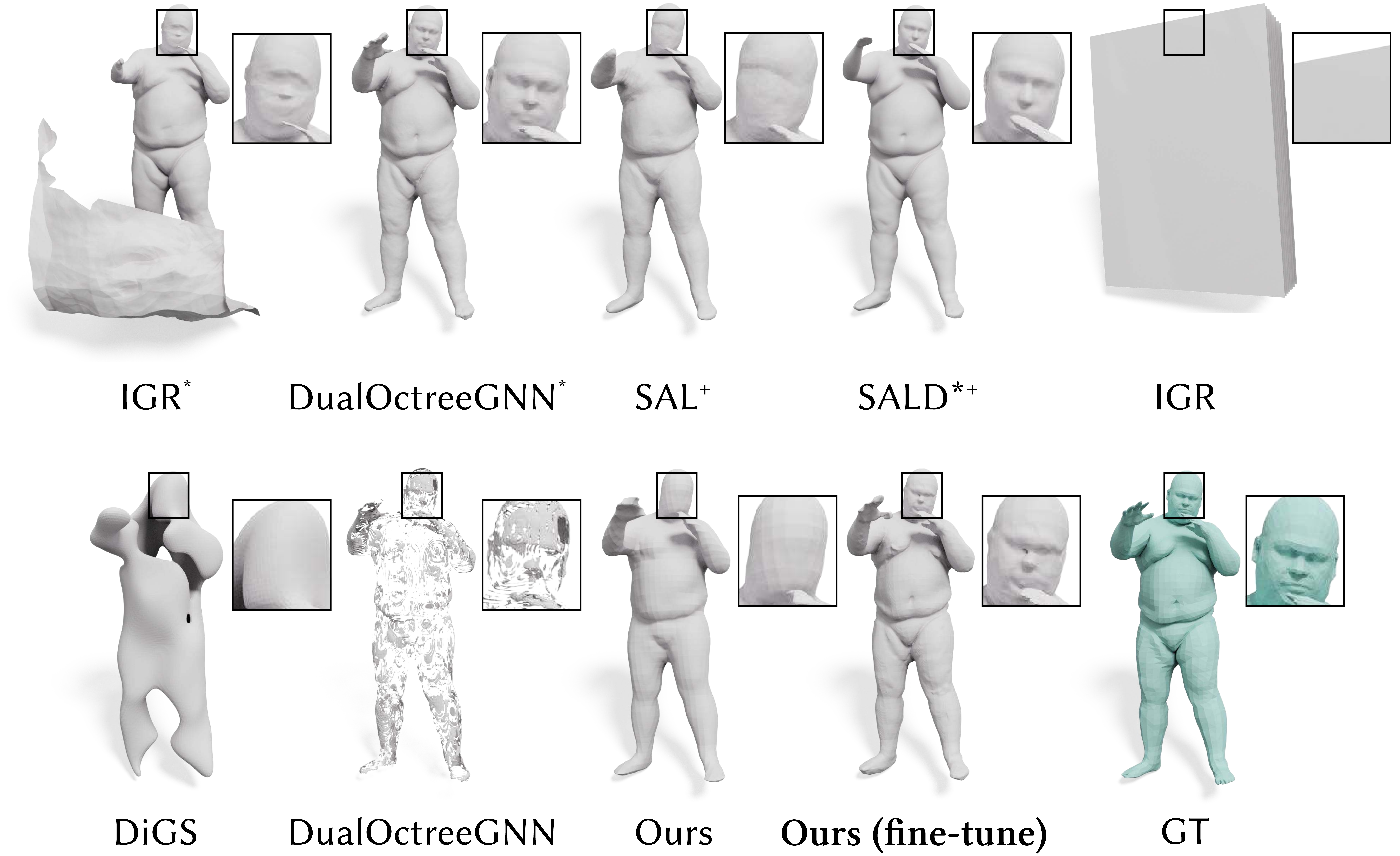}
    \caption{Visual comparison of shape space learning under DFAUST~\cite{dfaust:CVPR:2017}. 
    Our method can learn shape space without normals or additional supervision.
    By a fine-turning operation at the inference stage, our method can produce faithful shapes with great details.
    }
    \label{fig:dfaust_main}
\end{figure}

\begin{table}[!htp]
\centering
\caption{Quantitative comparison on large scans~\cite{three_scans}.
The methods marked with `$^*$' require normals, 
and the methods marked with `$^+$' are supervision based.}
\label{table:large}
\resizebox{\linewidth}{!}{%
\begin{tabular}{l|cccccc} 
\toprule
                     & \multicolumn{2}{c}{Normal C.~$\uparrow$}  & \multicolumn{2}{c}{Chamfer~$\downarrow$} & \multicolumn{2}{c}{F-Score$^{\vartriangle}~\uparrow$}  \\
                     & mean           & std.          & mean          & std.                     & mean           & std.                                 \\ 
\midrule
SIREN$^*$~\footnotesize{(trimmed)~\cite{SIREN}} & 98.38          & 0.30          & 0.96          & 0.17                     & 63.61          & 19.62                                \\ 
\midrule
PCP~\footnotesize{\cite{PCP}}                 & 94.32          & 2.58          & 4.44          & 1.41                     & 11.83          & \under{9.41}                        \\ 
DiGS~\footnotesize{\cite{DiGS}}                & 97.41          & 0.62          & 0.92          & 0.22                     & 63.78          & 21.91                                \\
\midrule
NG$^+$~\footnotesize{\cite{Galerkin}}             & 85.97          & 9.10          & 3.02          & 2.09                     & 31.01          & 20.67                                \\ 
\midrule
\textbf{Ours}        & \under{98.44} & \under{0.15} & \under{0.74} & \under{0.09}            & \under{79.77} & 9.90                                 \\
\bottomrule
\end{tabular}
}
\end{table}

\subsection{Learning Shape Space}

\paragraph{Dataset}
The D-Faust dataset~\cite{dfaust:CVPR:2017} contains high-resolution raw scans (triangle soups) of 10 humans in various poses. We followed the methodology outlined in DualOctreeGNN~\cite{wang2022dual} to perform splitting, using 6K scans for training and 2K scans for testing. The raw point clouds used as input are incomplete and noisy due to occlusion during the scanning process and the limited precision of scanners.

\paragraph{Training details}
During the training phase, we utilized an encoder based on Convolutional Occupancy Networks~\cite{Peng20ConvONet} to encode shapes that enable shape space learning. Specifically, we projected sparse on-surface point features obtained using a modified PointNet~\cite{PointNet} onto a regular 3D grid and used a convolutional module to propagate these features to the off-surface area. The query feature was then obtained using trilinear interpolation. Additionally, we employed FiLM conditioning~\cite{piGAN2021}, which applies an affine transformation to the network’s intermediate features, as SIREN struggles with handling high-dimensional inputs~\cite{piGAN2021, mehta2021modulated}. Furthermore, inspired by~\cite{SA-ConvONet}, we fine-tuned the network during the inference phase to perform accurate geometry learning for high-fidelity surface reconstruction using our novel loss. Our models were trained for 200 epochs using the AMSGrad optimizer~\cite{adamgrad} with an initial learning rate of $10^{-4}$, which was decayed to $10^{-6}$ using cosine annealing~\cite{loshchilov2017sgdr}. The training set was divided into mini-batches containing 32 different shapes (with accumulated gradients), with each shape being randomly sampled as 10K points.

% In the training phase, we adopt the encoder following Convolutional Occupancy Networks~\cite{Peng20ConvONet} to encode the shapes to enable shape space learning.
% Specifically, we project the sparse on-surface point features obtained using a modified PointNet~\cite{PointNet} onto a regular 3D grid, then use a convolutional module to propagate sparse on-surface point features to the off-surface area, and finally obtain the query feature using bilinear interpolation. 
% Besides, we adopt the FiLM conditioning~\cite{piGAN2021} that 
% applies an affine transformation to the network's intermediate features since SIREN is weak in handling high-dimensional inputs~\cite{piGAN2021, mehta2021modulated}. 
% Furthermore, inspired by~\cite{SA-ConvONet},
% we fine-tune the network in the inference phase to perform accurate geometry learning for high-ﬁdelity surface reconstruction based on our novel loss. 
% We train our models for $200$ epochs using AMSGrad optimizer~\cite{adamgrad} with an initial learning rate of $10^{-4}$ and decay to $10^{-6}$ using cosine annealing~\cite{loshchilov2017sgdr}.
% We divided the training set into mini-batches: a batch contains
% 32 different shapes (accumulate gradients), where each shape is randomly sampled as 10K points.

\begin{table}[!htp]
\centering
\caption{Quantitative comparison on DFAUST~\cite{dfaust:CVPR:2017}. The methods marked with `$^*$' require normals, 
and the methods marked with `$^+$' are supervision based. 
In each column, the \under{best} scores are highlighted in bold,
while the \textbf{second best} scores are highlighted in bold with underlining.}
% In each column, the \under{best} scores are textbfd and highlighted in bold,
% while the \textbf{second best} scores are highlighted in bold but not textbfd.}
\resizebox{\linewidth}{!}{%
\label{table:dfuast}
\begin{tabular}{l|cccccc} 
\toprule
\multirow{2}{*}{}                  & \multicolumn{2}{c}{Normal C.~$\uparrow$} & \multicolumn{2}{c}{Chamfer~$\downarrow$} & \multicolumn{2}{c}{F-Score~$\uparrow$}  \\
                                   & mean  & std.                  & mean  & std.                             & mean  & std.                            \\ 
\midrule
IGR$^*$~\footnotesize{\shortcite{IGR}}                             &   92.02     &      3.34                  &   29.01      &        33.61                          &  73.32     &    14.05                             \\

DualOctreeGNN$^*$~\footnotesize{\shortcite{wang2022dual}}                         & \under{97.65} & \under{0.34}                  & \under{1.78}  & 3.70                             & \under{97.48} & 1.03                            \\ 
\midrule
SAL$^+$~\shortcite{SAL}                                & 96.77 & 0.81                  & 2.82  & 4.67                             & 91.35 & 9.15                            \\
SALD$^{*+}$~\shortcite{SALD}                                & 97.04 & 0.92                  & 3.06  & \textbf{1.32}                            & 88.56 & 12.73                           \\ 
\midrule
IGR~\footnotesize{\shortcite{IGR}}                                & 57.93 & 3.40                  & 48.56 & 2.35                             & 6.54  & \under{0.11}                           \\
DiGS~\footnotesize{\shortcite{DiGS}}  & 87.60      &  1.91                     &  11.87     &   4.56                               &   37.77    & 7.11     \\ 
DualOctreeGNN~\footnotesize{\shortcite{wang2022dual}}                         & 92.42 & \textbf{0.45}                  & 3.02  & 2.38                             & 85.77 & 3.50                            \\ 
\midrule
Ours             & 96.22      &  0.55                     &  2.50     &   0.51                               &   94.45    & 1.48                                \\
\textbf{{Ours (fine-tune)}} &   \textbf{97.05}    &   0.50                    & \textbf{1.96}      &       \under{0.27}                           &   \textbf{96.46}    &      \textbf{0.93}                           \\
\bottomrule
\end{tabular}
}
 % \vspace{-2mm}
\end{table}

\paragraph{Results}
The baseline approaches we compared against include IGR~\cite{IGR}, SAL~\cite{SAL}, SALD~\cite{SALD}, DualOctreeGNN~\cite{wang2022dual}, and DiGS~\cite{DiGS}. We also demonstrated the results of IGR and DualOctreeGNN trained without normals. As shown in Fig.~\ref{fig:dfaust_main} and Tab.~\ref{table:dfuast}, while IGR is capable of generating detailed results, it also produces spurious planes away from the input. SAL, which is supervised using unsigned distance, can only produce smooth results. SALD, with the support of normal supervision, can generate more detailed results but its reconstruction accuracy is worse than that of SAL due to a large systematic misalignment that does not respect input poses. In contrast, DualOctreeGNN produces the most impressive results due to its well-designed octree network’s ability to capture local priors for details. However, the performance of both IGR and DualOctreeGNN is compromised without normals. 
In particular, the resulting surfaces of DualOctreeGNN for point clouds without input normals are not watertight, despite the small Chamfer distances.
While DiGS also being based on SIREN, it cannot produce reliable results. In summary, our method is capable of learning shape space without requiring input normals or additional supervision and still produces faithful shapes. It is important to note that we did not fine-tune other auto-encoder-based methods (except for IGR and DiGS, which leverage auto-decoder and have to be optimized during the inference stage) as our method consistently outperformed unoriented input even without fine-tuning.

\begin{table}[!htp]
\centering
\caption{Comparison about different gradient constraints.}
\label{table:abl_eikonal}
 \resizebox{\linewidth}{!}{%
\begin{tabular}{l|cccccc} 
\toprule
                               & \multicolumn{2}{c}{Normal C.~$\uparrow$}                    & \multicolumn{2}{c}{Chamfer~$\downarrow$}        & \multicolumn{2}{c}{F-Score~$\uparrow$}  \\
                               & mean                    & std.                   & mean                   & std.                   & mean                    & std.                        \\ 
\midrule
$\text{Eikonal}_\text{all}$                  & 97.33                   & 2.75                   & 3.74                   & 2.86                   & 88.04                   & 15.36                       \\ 
$\text{Eikonal}_\text{half}$                   & 97.52                   & 2.98                   & 3.42                   & 2.80                   & 89.07                   & 15.22                       \\ 
\midrule
$\sigma_\text{min} = {0.6}$ & 97.55                   & 2.84                   & 3.34                   & 2.47                   & 89.92                   & 14.58                       \\
$\sigma_\text{min} = {0.8}$ (\textbf{Ours}) & \textbf{\textbf{97.82}} & \textbf{\textbf{2.18}} & \textbf{\textbf{3.13}} & \textbf{\textbf{2.20}} & \textbf{\textbf{91.06}} & \textbf{\textbf{12.85}}     \\
$\sigma_\text{min} = {1.0}$ & 97.48                   & 3.18                   & 3.35                   & 2.57                   & 89.42                   & 14.70                       \\
\bottomrule
\end{tabular}
}
\end{table}

\subsection{Ablation Studies}
% Our ablation studies were conducted using the ABC~\cite{ABC} and Thingi10K~\cite{Thingi10K} datasets, with each dataset containing 100 shapes discretized into 10K points. Additionally, we used the SRB~\cite{DGP} dataset and five shapes from Stanford 3D Scanning Repository~(Armadillo, Bunny, Dragon, Asian dragon, and Thai Statue) to observe the effects of hyper-parameters.
We conducted ablation studies using two datasets: ABC~\cite{ABC} and Thingi10K~\cite{Thingi10K}. Each dataset comprised 100 shapes, discretized into 10K points. To examine the effects of hyper-parameters, we also utilized the SRB~\cite{DGP} dataset and five shapes from the Stanford 3D Scanning Repository: Armadillo, Bunny, Dragon, Asian Dragon, and Thai Statue.
More ablation studies can be found in our supplementary material.
% We use the same metric setting as before.

\begin{table}[!htp]
\centering
\caption{Comparison about different smooth energy forms: Dirichlet Energy~($E_D$), Hessian Energy~($E_{H_2}$), and Hessian Energy based-$L_1$~($E_{H_1}$).}
\label{table:abl_energy}
 % \resizebox{\linewidth}{!}{%
\begin{tabular}{l|cccccc} 
\toprule
          & \multicolumn{2}{c}{Normal C.~$\uparrow$}  & \multicolumn{2}{c}{Chamfer~$\downarrow$} & \multicolumn{2}{c}{F-Score~$\uparrow$}  \\
          & mean           & std.          & mean          & std.                     & mean           & std.                                 \\ 
\midrule
$E_{D}$ & 94.42          & 3.71          & 8.13          & 6.75                     & 64.45          & 31.23                                \\
% Laplacian & 96.29          & 3.70          & 5.34          & 5.51                     & 80.89          & 25.03                                \\
$E_{H_2}$   & 97.62          & 2.58          & 3.33          & 2.28                     & 87.94          & 17.70                                \\
$E_{H_1}$  & 97.45          & 2.74          & 3.85          & 3.90                     & 87.92          & 18.48  \\
\textbf{Ours}      & \under{97.82} & \under{2.18} & \under{3.13} & \under{2.20}            & \under{91.06} & \under{12.85}                       \\
\bottomrule
\end{tabular}
% }
\end{table}

\begin{figure}
    \centering
    \includegraphics[width=\linewidth]{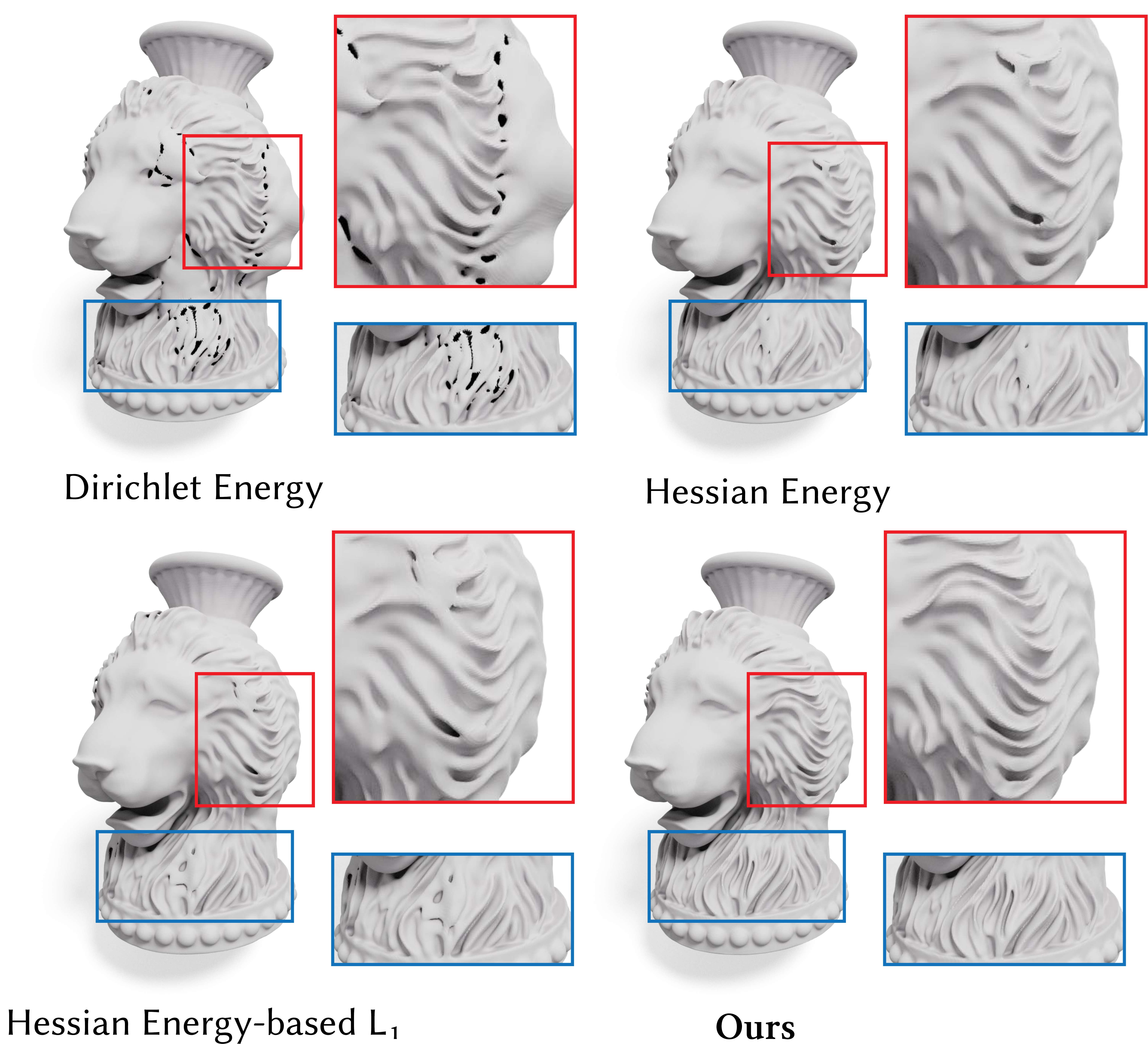}
    \caption{
    Visual comparison of different smooth energy forms. Our method is superior to the other approaches in terms of the ability to recover geometry details.
    }
    \label{fig:abl_energy}
\end{figure}

\subsubsection{Relaxing Eikonal Constraint}

In the following, we will demonstrate the effectiveness of relaxing the Eikonal constraint. 
Our relaxation technique is two-fold.
First, we hope that the gradients do not vanish,
rather than be exactly a unit vector. 
Second, the gradient constraint is specified 
around the surface rather than in the whole space. 
For the ablation study purpose,
we test the effect of substituting the traditional Eikonal condition for our relaxed gradient constraint. 
The first setting named $\text{Eikonal}_\text{all}$, requires the gradients at all points, 
i.e., the input points and the query points,
to be unit vectors. 
The second setting, named $\text{Eikonal}_\text{half}$,
is to specify the Eikonal constraint at the input points.
At the same time, we test different choices of $\sigma_{min}$ for our approach, i.e., $0.6, 0.8, \text{and}\ 1$.
The statistics in Tab.~\ref{table:abl_eikonal}
show that (1)~it seems better to specify the Eikonal constraint at the input points than at all the points, and (2)~the inequality of $\|\nabla f\|_2\geq \sigma_{min}$ is easier to solve compared with $\|\nabla f\|_2=1$.

\subsubsection{Comparison to Smooth Energy Forms}
The commonly used Dirichlet energy is as follows:
\begin{equation}
E_{D}=\frac{1}{2}  \int_{\mathcal{P} \cup \mathcal{Q}_\text{far}} \|\nabla f(x)\|_2^2 d x,
\end{equation}
where $\Omega$ is the space of the whole bounding box.
The commonly used Hessian energy is as follows:
\begin{equation}
E_{H_2}=  \int_{\mathcal{P} \cup \mathcal{Q}_\text{far}} \|\mathbf{H}_f(x)\|_2^2 d x.
\end{equation}

\citet{zhang2022critical} leveraged $L_1$-based Hessian:
\begin{equation}
E_{H_1}=  \int_{\mathcal{P} \cup \mathcal{Q}_\text{far}} \|\mathbf{H}_f(x)\|_1 d x.
\end{equation}

To ensure fairness in our evaluation, we combined the smoothness energy with relaxed Eikonal conditions (as described in Eq.~\ref{eq:relax}) to demonstrate the effectiveness of~$L_{\text{singularH}}$. As shown in Tab.~\ref{table:abl_energy} and Fig.~\ref{fig:abl_energy}, our method not only suppresses ghost geometry but also recovers high-fidelity geometric details. This demonstrates its superiority over both the Dirichlet energy, which produces ghost geometry, and the Hessian energy, which produces over-smoothed results possibly with adhesion in the area of sharp, thin features. It is important to note that when the Hessian energy becomes zero, all entries of~$\mathbf{H}$ become zero, causing the SDF to degenerate into a linear function and diminishing its ability to accurately represent geometric details. In contrast, our term~$L_{\text{singularH}}: \text{Det}(\mathbf{H}_f)=0$ is more conservative, allowing for flexibility and capacity to recover geometric details. DiGS~\cite{DiGS} utilizes another smoothness energy form, Laplacian energy of the Hessian matrix, we compare against it in the next subsection.

\begin{table}[!htp]
\centering
\caption{Quantitative comparison with DiGS~\cite{DiGS}. 
Based on the comparison,
our method exhibits a bigger advantage under different settings. 
% and our method consistently shows better results even though DiGS use the same settings as ours.
% \ZX{May some problems.}
}
\label{table:abl_digs}
 \resizebox{\linewidth}{!}{%
\begin{tabular}{l|ccc|cccccc} 
\toprule
\multirow{2}{*}{}     & \multirow{2}{*}{Ours Weights} & \multirow{2}{*}{SIREN Init.} & \multirow{2}{*}{Eikonal Relax} & \multicolumn{2}{c}{Normal C.~$\uparrow$}  & \multicolumn{2}{c}{Chamfer~$\downarrow$} & \multicolumn{2}{c}{F-Score~$\uparrow$}  \\
                      &                               &                              &                                & mean           & std.          & mean          & std.                     & mean           & std.                  \\ 
\midrule
\multirow{5}{*}{DiGS} &                               &                              &                                & 95.86          & 4.71          & 6.13          & 6.26                     & 70.34          & 29.56                 \\
                      &                               &                              & \checkmark                            & 96.15          & 4.94          & 5.08          & 4.65                     & 75.10          & 27.96                 \\
                      &                               & \checkmark                          &                                & 95.59          & 5.06          & 5.89          & 7.03                     & 76.02          & 30.48                 \\
                      &                               & \checkmark                          & \checkmark                            & 96.51               &       4.48        &   5.00            &    6.19                      &     81.40           &  26.11                     \\
                      & \checkmark                           & \checkmark                          & \checkmark                            &  96.32              &   3.35            &  4.71             &   5.27                       &  79.71              & 26.32                      \\ 
\midrule
\multirow{4}{*}{Ours} & \checkmark                           &                              & \checkmark                            & 95.36          & 4.39          & 5.79          & 4.81                     & 74.57          & 25.60                 \\
                      & \checkmark                           & \checkmark                          &                                & 97.33          & 2.75          & 3.74          & 2.86                     & 88.04          & 15.36                 \\
                      &                               & \checkmark                          & \checkmark                            & 96.55          & 4.60          & 4.52          & 5.06                     & 84.50          & 20.94                 \\
                      & \checkmark                           & \checkmark                          & \checkmark                            & \under{97.82} & \under{2.18} & \under{3.13} & \under{2.20}            & \under{91.06} & \under{12.85}        \\
\bottomrule
\end{tabular}
}
\end{table}

\subsubsection{Comparison with DiGS} 
\begin{figure}[!t]
    \centering
    \includegraphics[width=\linewidth]{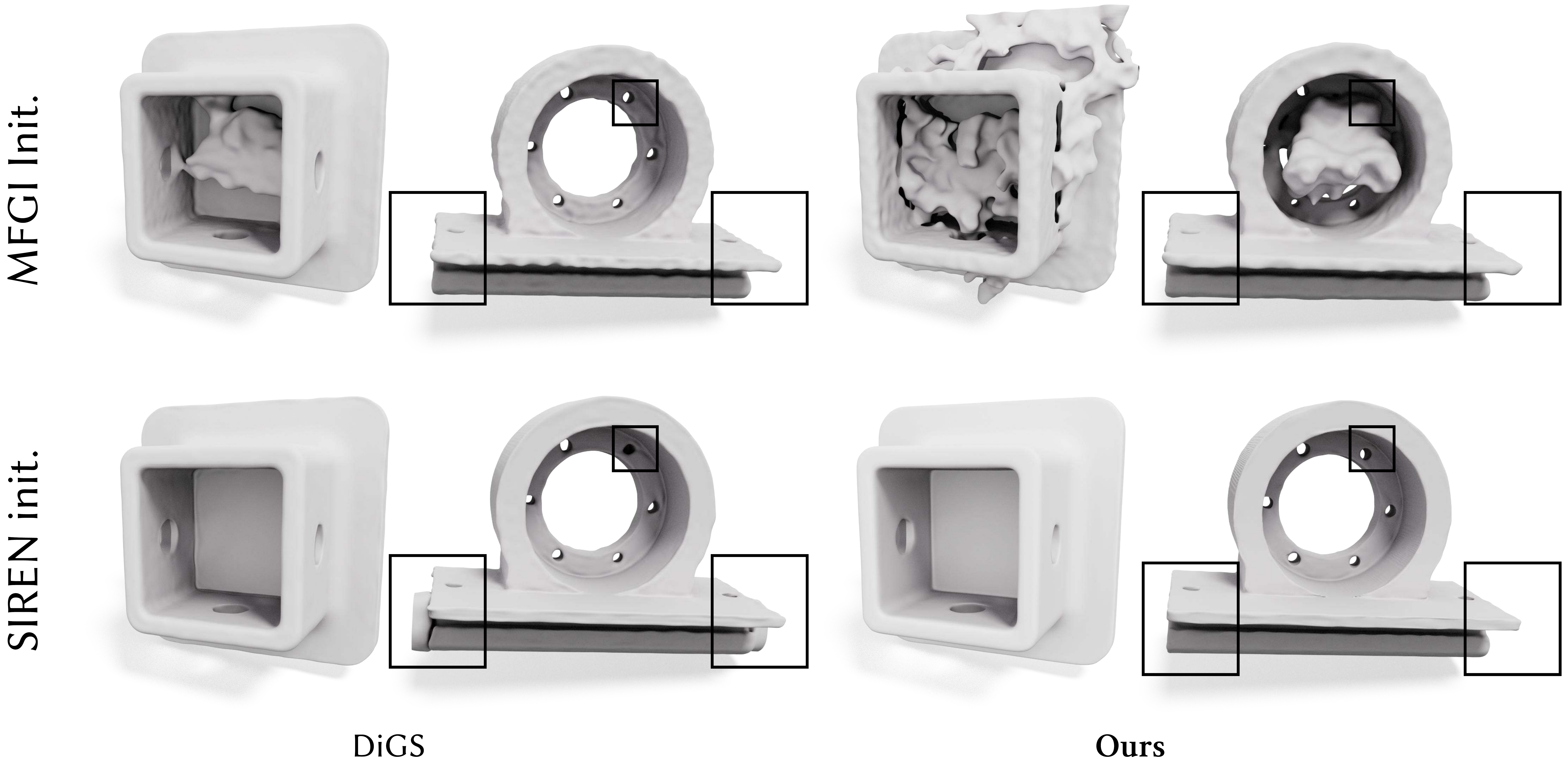}
    \caption{The ablation studies for initiation methods. DiGS~\cite{DiGS} is more sensitive to initiation. Both initiation methods cannot produce reliable results.}
    \label{fig:abl_init}
\end{figure}

DiGS~\cite{DiGS} is another unoriented point cloud reconstruction method based on SIREN~\cite{SIREN}. Notably, DiGS leverages the well-known smoothness of Laplacian energy based on the Hessian matrix for reconstruction. We conduct comprehensive experiments with DiGS and present the results in Tab.~\ref{table:abl_digs}, which demonstrate that our method outperforms DiGS. Our observations are three-fold. First, we relax the Eikonal constraint in DiGS using our gradient constraint and found that DiGS produced better results with our proposed gradient constraint. Second, as shown in Fig.~\ref{fig:abl_init}, the MFGI-based initialization, which initializes the SIREN network as an approximate sphere, is unable to reconstruct the concave parts of CAD shapes. In contrast, initializing SIREN directly appears to produce better results. However, unlike our approach, DiGS does not consistently yield better results when switching to direct SIREN initialization.

% DiGS~\cite{DiGS} is another unoriented point cloud reconstruction method based on SIREN~\cite{SIREN}. It is worth mentioning that DiGS leverages the well-known smoothness of Laplacian energy based on the Hessian matrix for reconstruction. We conducted comprehensive experiments with DiGS~\cite{DiGS} and present the statistics in Tab.~\ref{table:abl_digs}, which demonstrate that our method outperforms DiGS. Our observations are three-fold. First, we relaxed the Eikonal constraint in DiGS using our gradient constraint and found that DiGS produced better results with our proposed gradient constraint. Second, Fig.~\ref{fig:abl_init} shows that the MFGI-based initialization, which initializes the SIREN network as an approximate sphere, is unable to reconstruct the concave parts of CAD shapes. Instead, initializing SIREN directly appears to produce better results. However, unlike our approach, DiGS does not consistently yield better results when switching to direct SIREN initialization.
% Further, we do not leverage the MFGI initiation method proposed by DiGS to initiate the network to the signed distance function of the sphere.
% From Fig.~\ref{fig:abl_init}, we can observe that the methods initially with MFGI can not reconstruct the concave parts of CAD shapes, which damages the geometry of the results.
% Thus, both methods initiated with the initiation of SIREN introduced to get better results.
% However, DiGS yield bad results when changing to SIREN initiation for some shapes, which is why it can not get the same improvement as ours.
Finally, even when using the weighting scheme of DiGS~$(\lambda_\text{manifold}$, $\lambda_\text{non-manifold}, \lambda_\text{Eikonal} )=(3000, 100, 50)$,
our method still outperforms all variants of DiGS.
If the weights are adjusted  to our preferred setting ($\lambda_\text{manifold}$= $7000$, $\lambda_\text{non-manifold}$ = $600$, $\lambda_\text{Eikonal}^\text{relax}$ = $50$),
our method exhibits an even greater advantage, while the performance of DiGS variants diminishes.

% \subsubsection{Settings for $Q_{near}$}
% \usepackage{multirow}
% \usepackage{booktabs}

% \usepackage{multirow}
% \usepackage{booktabs}

\section{Conclusions}
Learning the implicit neural representation from an unoriented point cloud is a fundamental task. In this paper, we propose to regularize the implicit function by enforcing singular Hessian near the surface. Extensive experimental results demonstrate that our approach exhibits the superior ability to recover high-fidelity geometric details in the presence of various imperfections.

\appendix
\section{Additional Ablation Studies}
\begin{table}
\centering
\caption{Ablation studies on the loss functions.}
\label{tab:abl_siren}
\resizebox{\linewidth}{!}{%
\begin{tabular}{ccc|cccccc} 
\toprule
\multirow{2}{*}{Eikonal Relax} & \multirow{2}{*}{Singular Hessian Term} & \multirow{2}{*}{Our weights} & \multicolumn{2}{c}{Normal C.~$\uparrow$} & \multicolumn{2}{c}{Chamfer~$\downarrow$} & \multicolumn{2}{c}{F-Score~$\uparrow$}  \\ 
% \cline{4-9}
                               &                             &                              & mean          & std.                & mean         & std.               & mean          & std.               \\ 
\midrule
                               &                             &                              & 85.28         & 6.53                & 17.62        & 14.37              & 28.58         & 20.99              \\
\checkmark                     &                             &                              & 88.54         & 6.71                & 15.55        & 11.33              & 36.64         & 31.15              \\
                               & \checkmark                  &                              & 96.34         & 3.44                & 4.25         & 5.06               & 84.38         & 20.08              \\
\multicolumn{1}{c}{}           &                             & \checkmark                   & 93.74         & 4.85                & 9.45         & 8.35               & 61.68         & 32.83              \\
\multicolumn{1}{c}{\checkmark} & \checkmark                  &                              & 96.55         & 4.60                & 4.52         & 4.17               & 84.50         & 20.94              \\
\multicolumn{1}{c}{\checkmark} &                             & \checkmark                   & 94.85         & 3.54                & 6.53         & 3.99               & 70.92         & 25.21              \\
\multicolumn{1}{c}{}           & \checkmark                  & \checkmark                   & 97.33         & 2.75                & 3.74         & 2.86               & 88.04         & 15.36              \\
\checkmark                     & \checkmark                  & \checkmark                   & \under{97.82} & \under{2.18}        & \under{3.13} & \under{2.20}       & \under{91.06} & \under{12.85}      \\
\bottomrule
\end{tabular}
}
\end{table}

\begin{table}[!htp]
\centering
\caption{Comparison of the weights of $L_{\text{singularH}}$ and procedure of coarse-to-fine under SRB~\cite{DGP} dataset.}
\label{table:abl_weights}
 \resizebox{.8\linewidth}{!}{%
\begin{tabular}{l|cccc} 
\toprule
                                                 & \multicolumn{2}{c}{Chamfer~$\downarrow$} & \multicolumn{2}{c}{F-Score~$\uparrow$}  \\
                                                 & mean         & std.                      & mean          & std.                    \\ 
\midrule
$\lambda_\text{singularH} = 0.3$ no-decay            & 4.29         & 1.38                      & 75.36         & 15.99                   \\
$\lambda_\text{singularH} = 3$ no-decay              & 5.02         & 1.83                      & 70.76         & 15.12                   \\
$\lambda_\text{singularH} = 30$ no-decay             & 7.41         & 2.78                      & 56.90         & 11.56                   \\ 
\midrule
$\lambda_\text{singularH} = 0.3$ \& decay               & 3.88         & 1.12                      & 78.94         & 16.45                   \\
$\lambda_\text{singularH} = 3$ \& decay~(\textbf{Ours}) & \under{3.76} & 0.98~                     & \under{81.38} & \under{13.73}           \\
$\lambda_\text{singularH} = 30$  \& decay                & 4.01         & 1.37                      & 77.14         & 14.07                   \\
\bottomrule
\end{tabular}
}
\end{table}

\subsection{Loss functions} 
We conduct experiments to observe the effects of the loss functions in Tab.~\ref{tab:abl_siren}.
Compared with the original unoriented version of SIREN~\cite{SIREN}, our method differs in three main ways: relaxed Eikonal constraint, singular Hessian term, and our preferred settings for loss weights.
These three factors are mutually influenced and simply tuning the weights cannot improve the overall performance. 
%First, each individual operation results in progress in quantitative analysis. However, simply tuning the weights cannot improve the overall performance. 
Instead, enabling all three factors results in a significant improvement in performance. It is important to note that our method outperforms other fitting-based methods even with only $L_\text{singularH}$ (without $L_\text{Eikonal}^\text{relax}$) and our preferred weights settings.

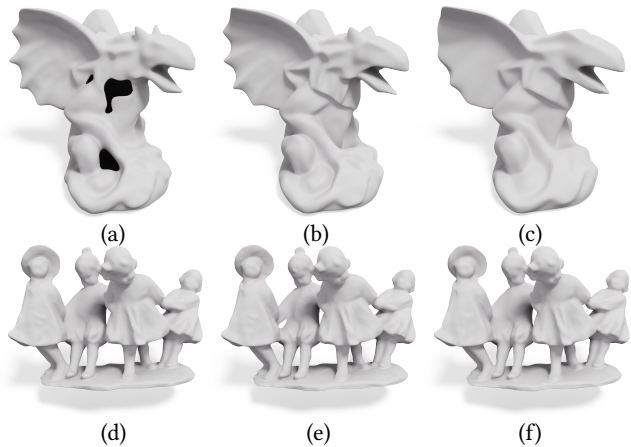
\begin{figure}[!htp]
\centering
\begin{overpic}[width=\linewidth]{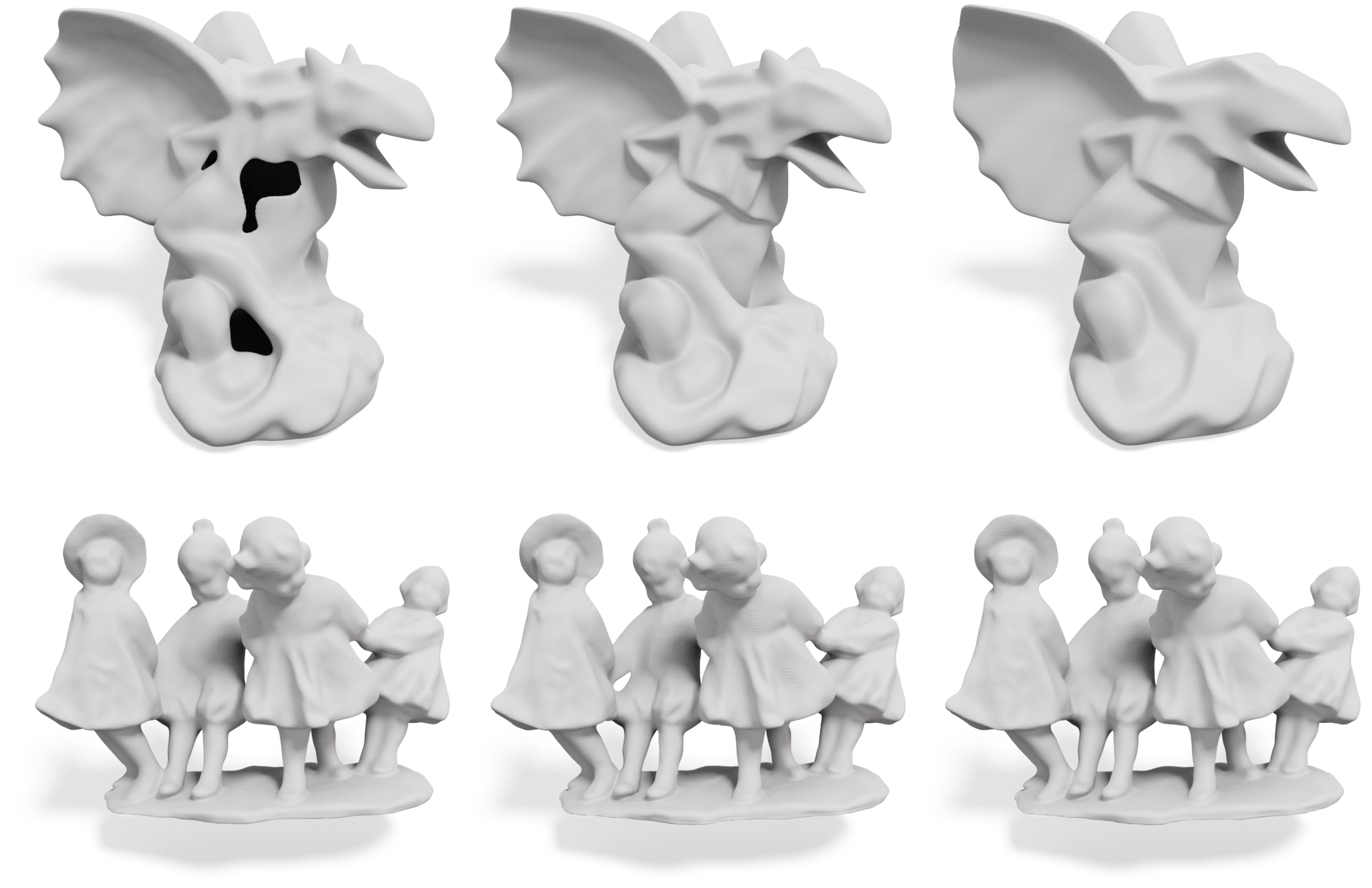}
    
    \put(17, 28.0){(a)}

    \put(48.5, 28.0){(b)}

    \put(82, 28.0){(c)}

    \put(17, -3.0){(d)}

    \put(49, -3.0){(e)}

    \put(82, -3.0){(f)}

    % \put(83, -3.0){(d)}
\end{overpic}

 \caption{
    Comparison about different $\lambda_\text{singularH}$ and coarse-to-fine training curriculum under SRB~\cite{DGP} that has different point clouds artifacts.
    From left to right: (a) $\lambda_\text{singularH} = 0.3 \& no-decay$, (b) $\lambda_\text{singularH} = 3 \& no-decay$ (c) $\lambda_\text{singularH} = 30 \& no-decay$, (d) $\lambda_\text{singularH} = 0.3 \& decay$, (e) $\lambda_\text{singularH} = 3 \& decay$, and $\lambda_\text{singularH} = 30 \& decay$ .
    }
    \label{fig:abl_weight}
\end{figure}

\subsection{Weight of $L_{\text{singularH}}$ and Coarse-to-fine training curriculum} 
\begin{figure}[!htp]
    \centering
    \includegraphics[width=\linewidth]{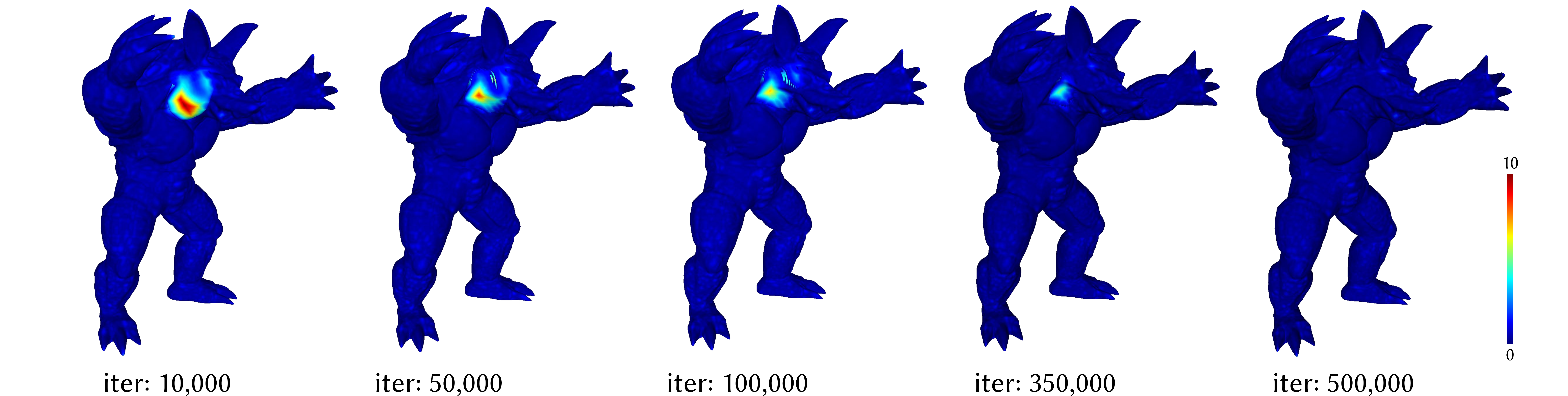}
    \caption{
    The reconstruction result of at different iterations with small weight~($0.01$). It requires a longer time to converge.
    % Reconstructions of a gecko model and serval unconnected~(butclose) torus of different radii. Each result is 
    }
    \label{fig:small_weight}
\end{figure}

We investigate the effects of several design choices made for $L_\text{singularH}$ over SRB~\cite{DGP} dataset with noise or missing parts shapes in Tab.~\ref{table:abl_weights}.
First, we examine the influence of different $L_\text{singularH}$ settings without annealing.
Fig.~\ref{fig:abl_weight} shows that larger weights lead to over-smooth results with topological errors, while smaller weights cannot fill the missing parts.
Additionally, all results exhibit  some topology errors without annealing.
We further test the effect of the annealing function $\tau$ with different $\lambda_\text{singularH}$ settings.
Our method achieves  the best performance with initialization $\lambda_\text{singularH}=3$, balancing geometric details and robustness to point cloud artifacts.
The remaining question is what are the effects if we keep a very small weight all the time?
In Fig.~\ref{fig:small_weight}, we show the results with weight $0.01$ at different iterations. It can be seen our method can also work well with a constant small weight, but it may take a longer time to converge without the coarse-to-fine training curriculum.

\begin{table}[!htp]
\centering
\caption{We compare the impact of the parameter $k$ on the sampling process of $Q_{near}$ using different numbers of points from the Stanford 3D Scanning Repository. The evaluation metric used in this comparison is the Chamfer distance.}
\label{tab:abl_k}
\begin{tabular}{l|cccc} 
\toprule
      & 500   & 1000 & 10000 & 100000  \\ 
\midrule
$k=5$   & 9.10  & 5.88 & 2.54  & 2.44    \\
$k=25$  & 9.40  & 5.64 & \under{2.51}  & 2.14    \\
$k=50$  & \under{8.86}  & 5.84 & 2.61  & \under{2.10}    \\
$k=75$  & 9.31  & 5.88 & 2.46  & 2.17    \\
$k=100$ & 9.41 & \under{5.56} & 2.55  & 2.14    \\
\bottomrule
\end{tabular}
\end{table}

\subsection{The Effect of $Q_{near}$ range}
Finally, we discuss the effect of the query location range. By default, we used the Gaussian destruction with the distance of $k=50$ neighbors as the standard deviation for sampling~$Q_{near}$.
Here, we use several candidates including $\{k=5, k=25, k=50, k=75, k=100\}$, to test the effects of different ranges of $Q_{near}$.
We report the results under five shapes from Stanford 3D Scanning Repository with different point clouds resolutions $\{500, 1000, 10000, 100000\}$ in Tab.~\ref{tab:abl_k}.
The comparison results indicate that there are no significant differences observed with different values of $k$. We choose to set $k=50$ for our study.
The comparison shows that a query location range that is either too small or too large will degrade surface reconstruction performance.

\subsection{Combined with Softplus}
\begin{figure}[!htp]
    \centering
    \includegraphics[width=\linewidth]{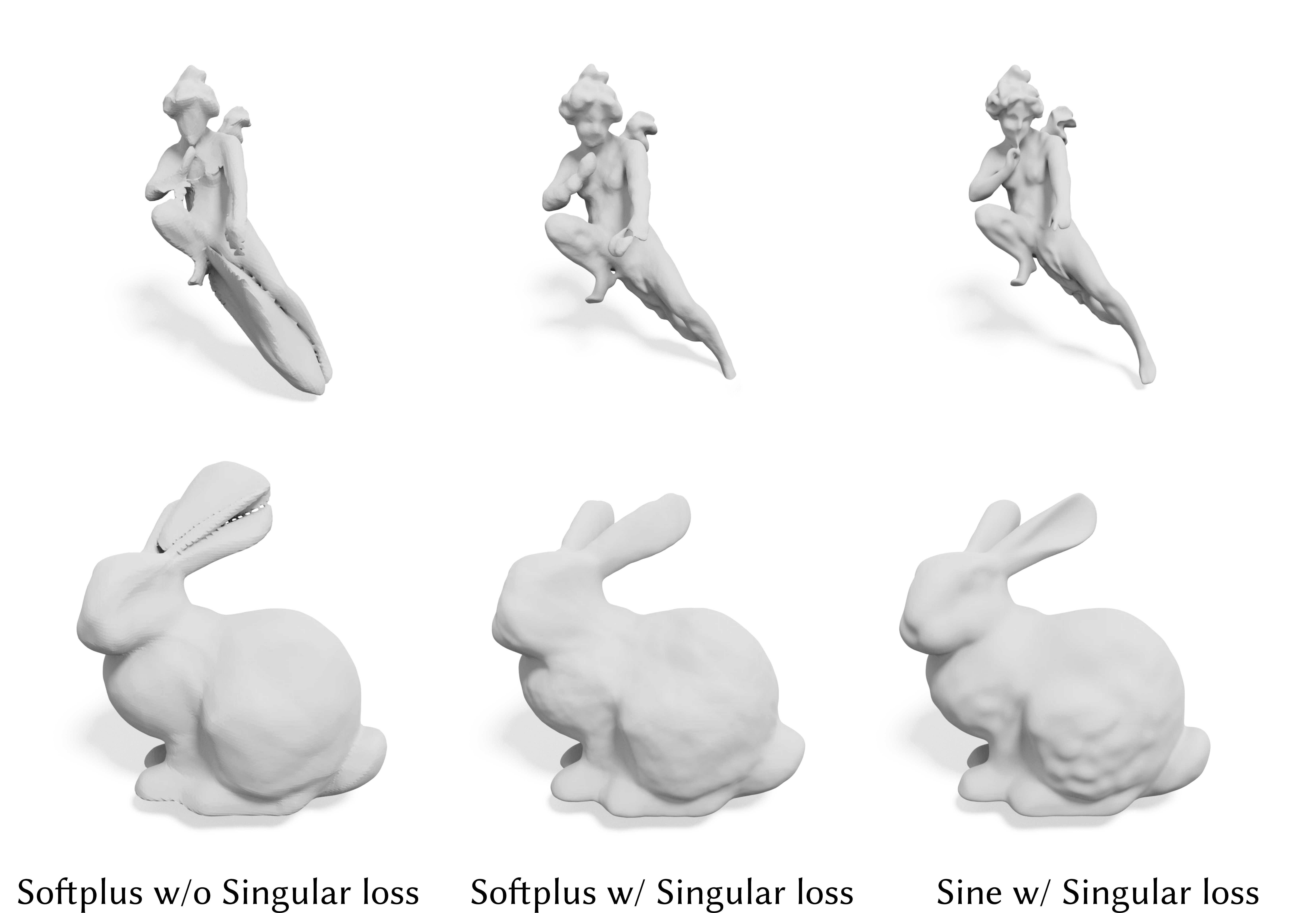}
    \caption{
    Our work consistently works well with other twice-differential activation functions though Softplus produce fewer details compared to Sine.
    % Reconstructions of a gecko model and serval unconnected~(butclose) torus of different radii. Each result is 
    }
    \label{fig:Abl_func}
\end{figure}

Our approach is general and can be applied to any network where second-order derivatives are defined across the entire domain. In our experiments, we employ Softplus, a smooth variant of ReLU, and initialize the network using GNI~\cite{SAL}. 
As depicted in Fig.~\ref{fig:Abl_func}, our method consistently yields reasonable results. However, similar to ReLU, Softplus also tends to produce low-frequency solutions when compared to sine functions, resulting in less detail.

% Since it is hard to use the query locations to probe the area around the surface of the query location range is too small, it is also hard to push the network to produce the accurate direction and distance to move the query locations to the surface of the
% query locations are too far away from the surface of query locations are too far away from the surface.

\subsection{Runtime Performance}
Second-order optimization increases the overhead of back-propagation. We included IGR~\cite{IGR}, SIREN~\cite{SIREN}, and DiGS~\cite{DiGS} for comparison. We set the batch size to 15K for all methods and utilized a network with four hidden layers and 256 units per layer for the SIREN-based methods, which is the default setting for our method. Tab.~\ref{tab:time} reports the time cost for a single iteration. Generally speaking, the time costs of DiGS and our method are higher than that of SIREN since DiGS and our method require second-order optimization. However, our method is more computationally efficient than IGR.

% The second-order optimization increases the overhead of the back-propagation.
% We include IGR~\cite{IGR}, SIREN~\cite{SIREN} and DiGS~\cite{DiGS} for comparison.
% We set the batch size to 15K for all the methods and utilized the network with four hidden layers, 256 units for each layer for the SIREN-based methods, which are the default setting for our method.
% Tab.~\ref{tab:time} reports the timing cost spent in a single iteration.
% Roughly speaking, the timing costs of DiGS and ours are higher than SIREN
% since DiGS and ours need a second-order optimization.
% However, ours is more computationally efficient than IGR.

\begin{table}[!htp]
\centering
\caption{Timing costs per iteration. 
The comparison is made among IGR~\cite{IGR}, SIREN~\cite{SIREN}, and DiGS~\cite{DiGS} without the supervision of normals. 
Timing statistics are reported in milliseconds (ms).}
\label{tab:time}
% \resizebox{\linewidth}{!}{%
\begin{tabular}{l|cccc} 
\toprule
              & IGR & SIREN & DiGS & Ours  \\
\midrule
\# parameters & 1.86M     &  264.4K     & 264.4K  & 264.4K      \\
time~[ms]     & 50.73      &  11.52           &  36.28    &  40.10     \\
\bottomrule
\end{tabular}
% }
\end{table}

\subsection{Illustrative Examples}

\begin{figure}[!htp]
    \centering
    \includegraphics[width=\linewidth]{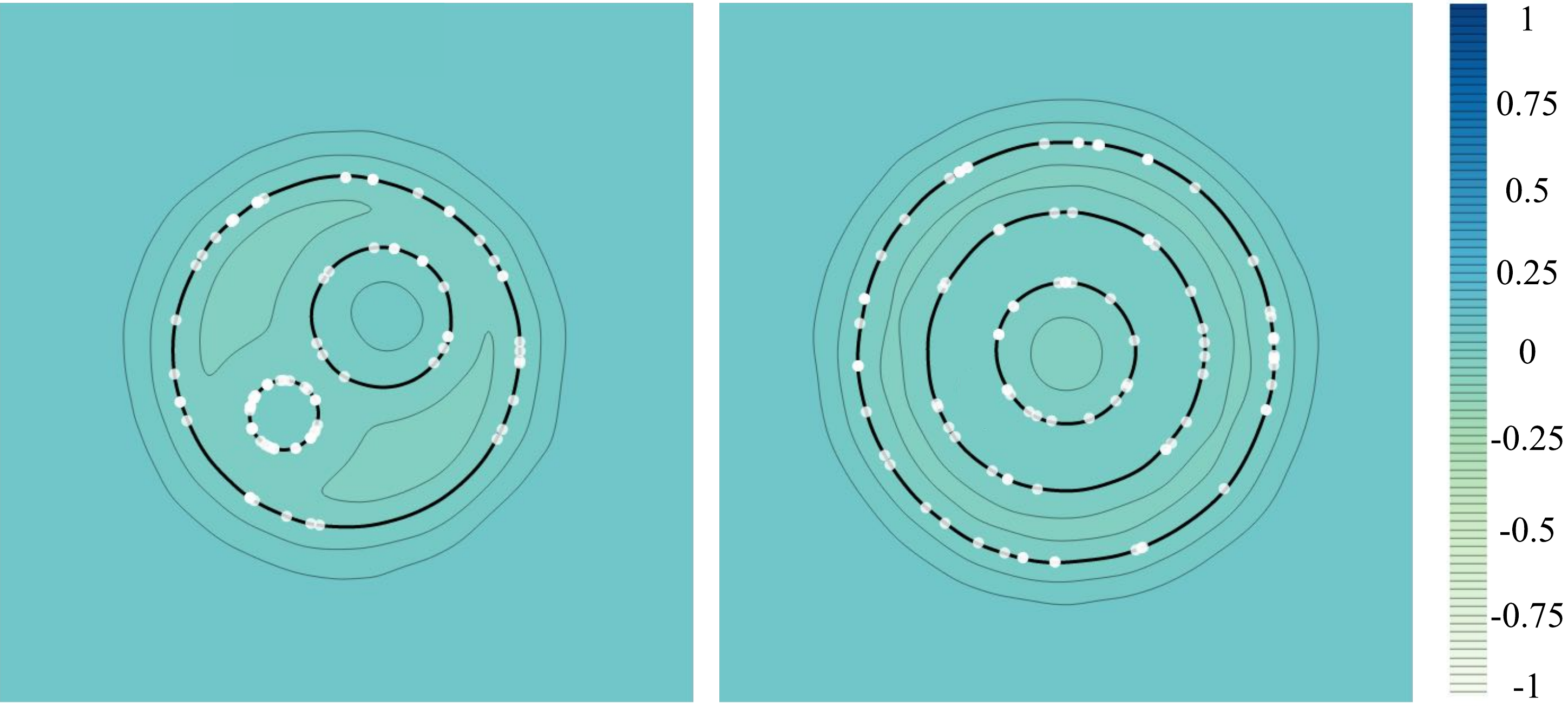}
    \caption{
    Our method can deal with nested surfaces. The left result is a disk with two holes, while the right shows three circles with different radii. Notably, there are 100 data points (white).}
    
     % Results of our method with nested surfaces. The left result is a disk with two holes, and the right is three different radii circles. It is worth noting that there are 100 data points~(white).
    \label{fig:2d_nest}
\end{figure}

\begin{figure}[!htp]
    \centering
    \includegraphics[width=.8\linewidth]{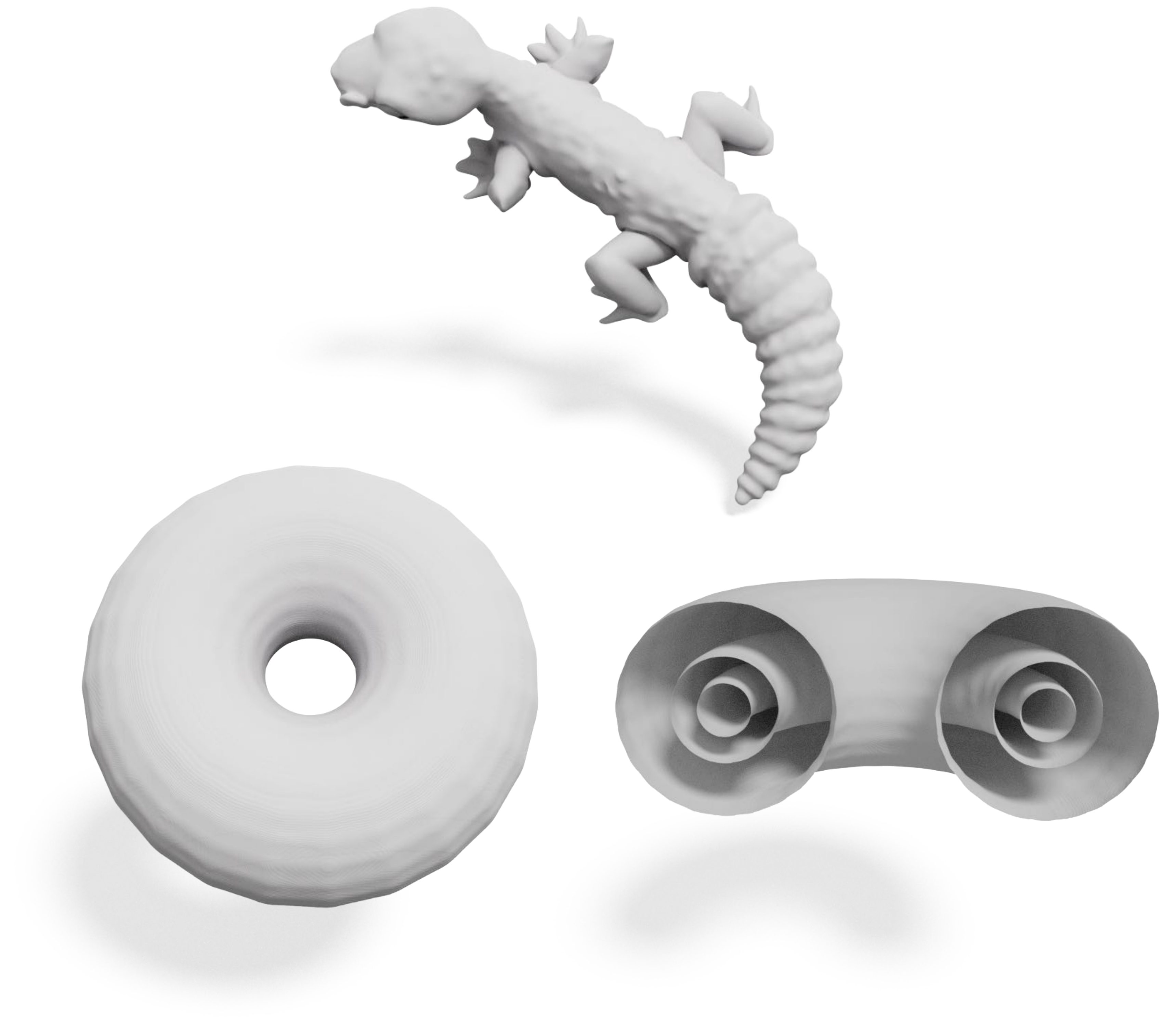}
    \caption{
    The reconstruction result of a gecko model and several separate tori with different radii. Each result is reconstructed with 10K input points.    
    % Reconstructions of a gecko model and serval unconnected~(but close) torus of different radii. Each result is reconstructed with 10K input.
    }
    \label{fig:3d_nest}
\end{figure}

\paragraph{Nested Surfaces}
Our method supports nested surfaces with multiple connected components. For easy visualization, we present 2D cases in Fig.~\ref{fig:2d_nest}.
Additionally, we present a 3D shape with several separate
tori with different radii to demonstrate our expressive ability.
We also visualize the gecko model in Fig.~\ref{fig:3d_nest}.
Our method works in a coarse-to-fine fashion and can eventually recover the true topology even for multi-surface shapes.

\begin{figure}[!htp]
    \centering
    \includegraphics[width=\linewidth]{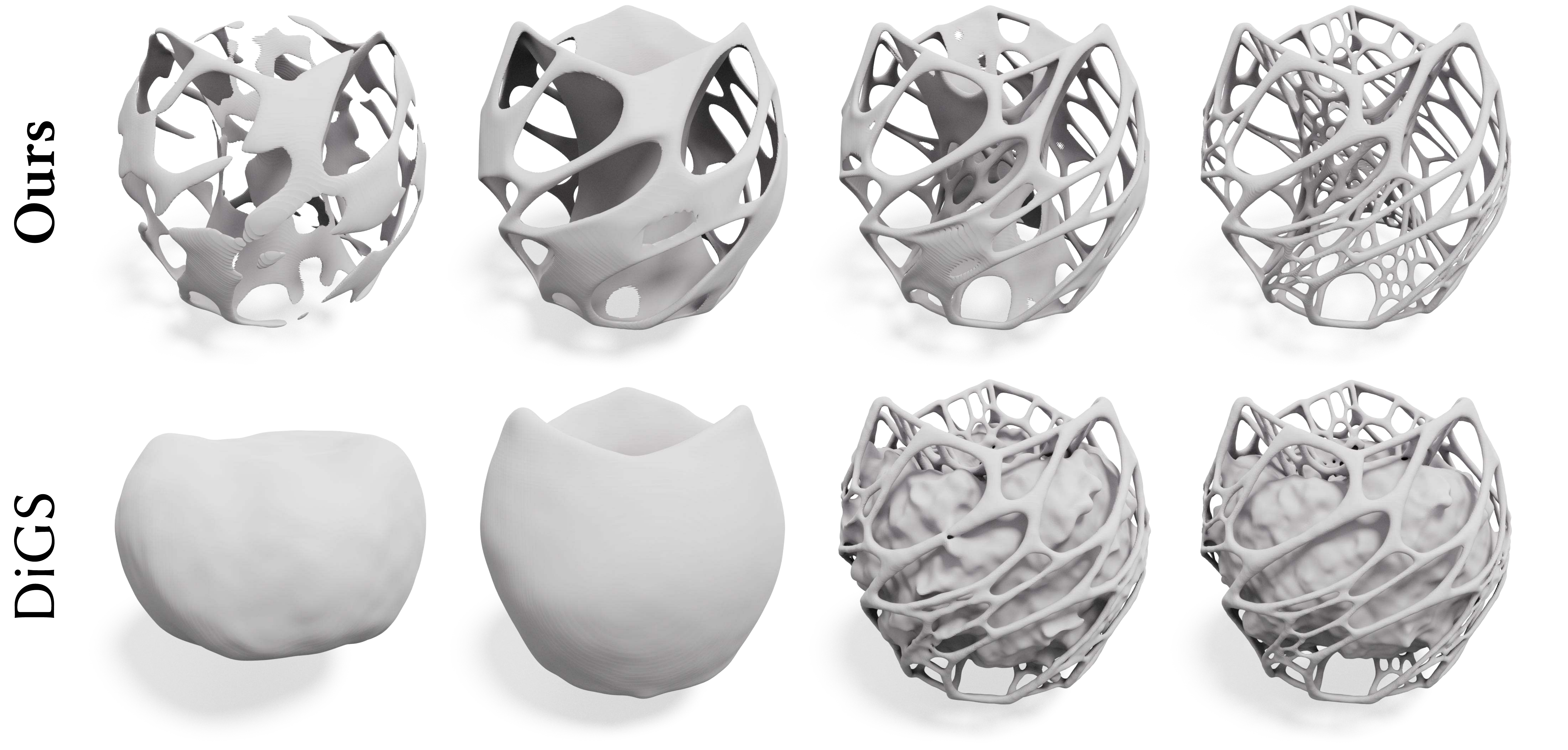}
    \caption{
    Comparing ours~(top) and DiGS~(bottom)~\cite{DiGS}
    by observing the four intermediate iterations under high genus shape.    
    The input is 100K points.
    }
    \label{fig:high_genus}
\end{figure}

\paragraph{High-genus Shapes}
Fig.~\ref{fig:high_genus} shows the iteration process of our method comparison to DiGS~\cite{DiGS} for high-genus shapes.
The input has 100K points with complex topology.
DiGS cannot incorrectly close the holes of the shapes.
Our method first suppresses the critical points that look forward to the coarse surface and then gradually recovers the real complex topology.

\begin{figure}[!htp]
\centering
\begin{overpic}[width=\linewidth]{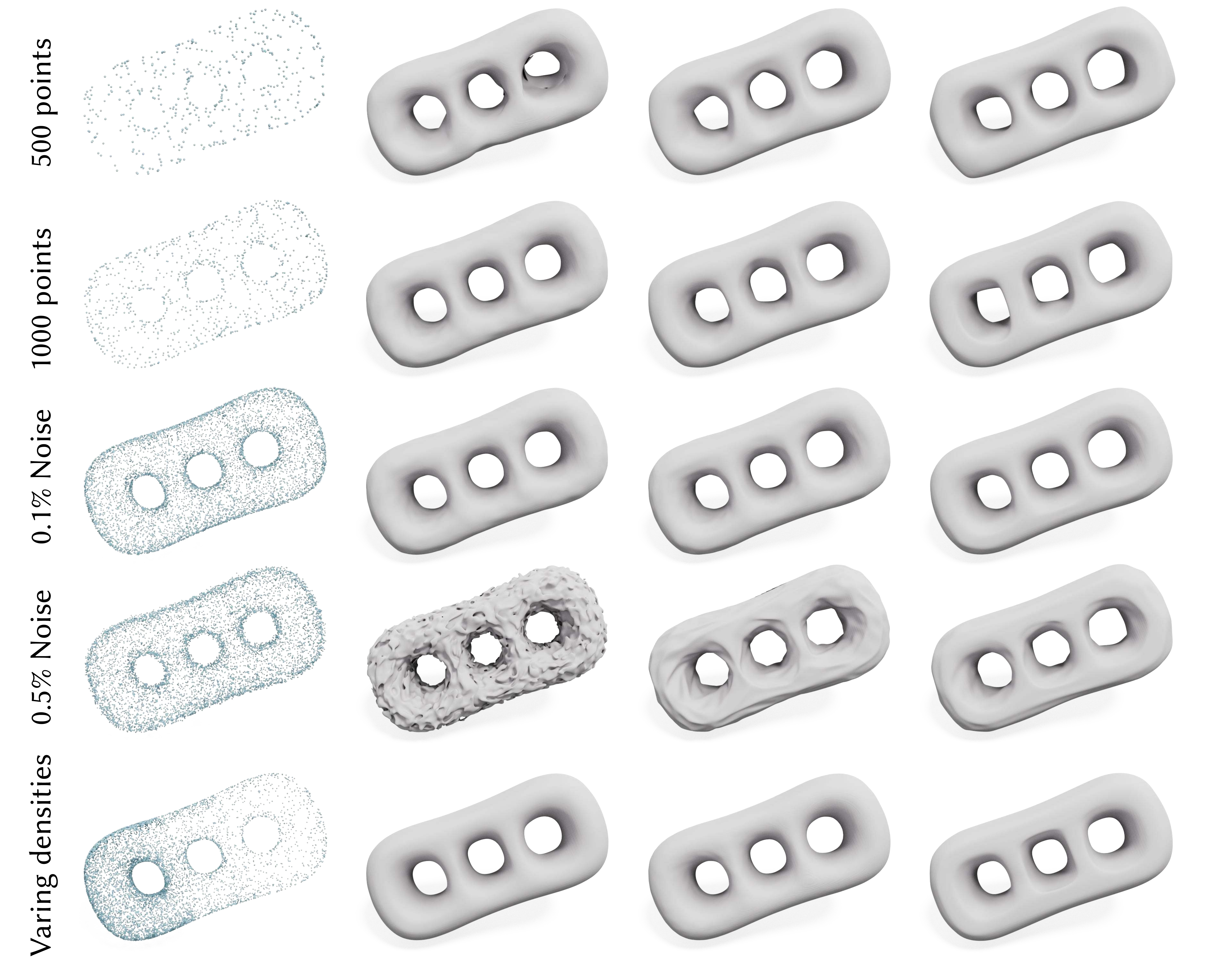}
    
    \put(15, -3.0){(a)}

    \put(37, -3.0){(b)}

    \put(60, -3.0){(c)}

    \put(83, -3.0){(d)}
\end{overpic}

 \caption{
    We test a point cloud of the genus-3 torus with different experimental configurations.
    From left to right: (a) Input, (b) $\lambda_\text{singularH} = 3 \& decay$, (c) $\lambda_\text{singularH} = 3 \& no-decay$, and (d) $\lambda_\text{singularH} = 30 \& no-decay$.
    }
    \label{fig:matrix}
\end{figure}
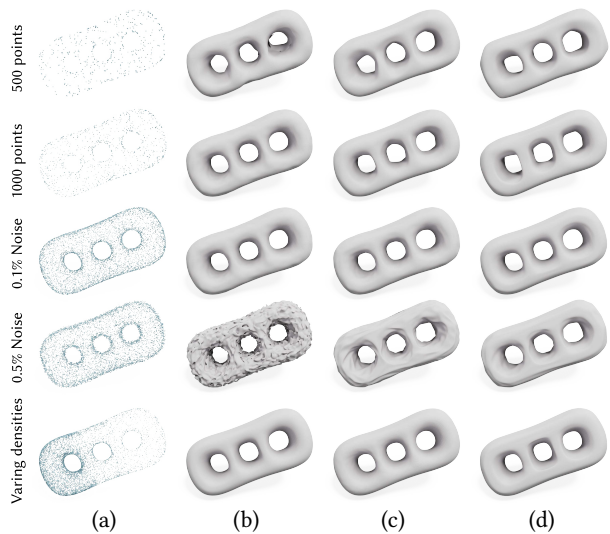

\paragraph{Varying Point Density, Data Sparsity, Noise}
In Fig.~\ref{fig:matrix}, we used three-hole shapes to test our methods with different point cloud artifacts, including varying point density, data sparsity, and noise. By default, we used 8K points except for data sparsity validation. We show the results under three different configurations: $\lambda_\text{singularH} = 3 \& decay$, $\lambda_\text{singularH} = 3 \& no-decay$ and $\lambda_\text{singularH} = 30 \& no-decay$.
We can observe that the default parameters $\lambda_\text{singularH} = 3 \& decay$ are robust to different point cloud artifacts. Additionally, large weights without decay are more robust to noisy and sparse inputs in terms of recovering the topology of the underlying shape, but they may yield over-smoothed results or deviate from the true surfaces.

% In Fig.~\ref{fig:matrix}, we use three-hole shapes to test our methods with different point cloud artifacts including varying point density, data sparsity, and noise.
% By default, we use 8K points except for data sparsity validation.
% We show the results under three different configurations: $\lambda_\text{singularH} = 3 \& decay$, $\lambda_\text{singularH} = 3 \& no-decay$ and $\lambda_\text{singularH} = 30 \& no-decay$.
% We can observe that the default parameters $\lambda_\text{singularH} = 3 \& decay$ are robust to different point cloud artifacts.
% Besides, the large weights without decay are more robust to noisy and sparse inputs in the capability of recovering the topology of the underlying shape, but they may yield over-smooth results or deviate from the real surfaces.
% On the other hand, our default parameters are the trade-off of the ability to recover the details of input.

\begin{figure}[!htp]
    \centering
    \includegraphics[width=\linewidth]{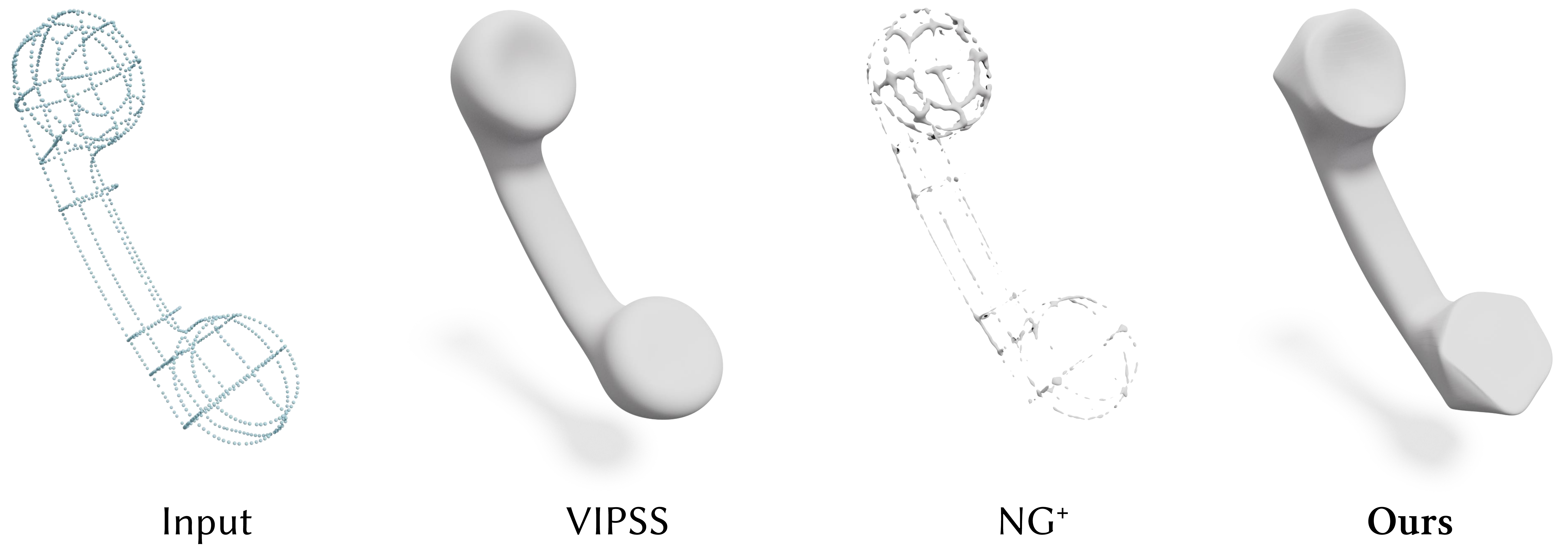}
    \caption{
    The reconstruction results from sketch points by different methods.
    }
    \label{fig:abl_sketch}
\end{figure}

\paragraph{Sketch Input}
It is interesting to determine whether our approach can transform a super-sparse 3D sketch point cloud into a meaningful shape. In VIPSS~\cite{VIPSS}, the authors provided a 3D sketch point cloud of approximately 1K points. We visualized the reconstructed results by VIPSS~\cite{VIPSS}, NG~\cite{Galerkin}, and our method in Fig.~\ref{fig:abl_sketch}. NG fails because its training set does not include sketch-type data. Our result is comparable to VIPSS, but VIPSS is severely limited by the number of points.

% It is interesting to make clear if our approach can transform a super-sparse 3D sketch point cloud into a meaningful shape. In VIPSS~\cite{VIPSS}, the authors gave a 3D sketch point cloud of about 1K points. 
% We visualize the reconstructed results by VIPSS~\cite{VIPSS}, NG~\cite{Galerkin}, and ours in Fig.~\ref{fig:abl_sketch}.
% It can be seen that NG fails since its training set does not include sketch-type data. 
% Our result is comparable to VIPSS but VIPSS seriously suffers from the number of points.

\begin{figure}[!htp]
    \centering
    \includegraphics[width=\linewidth]{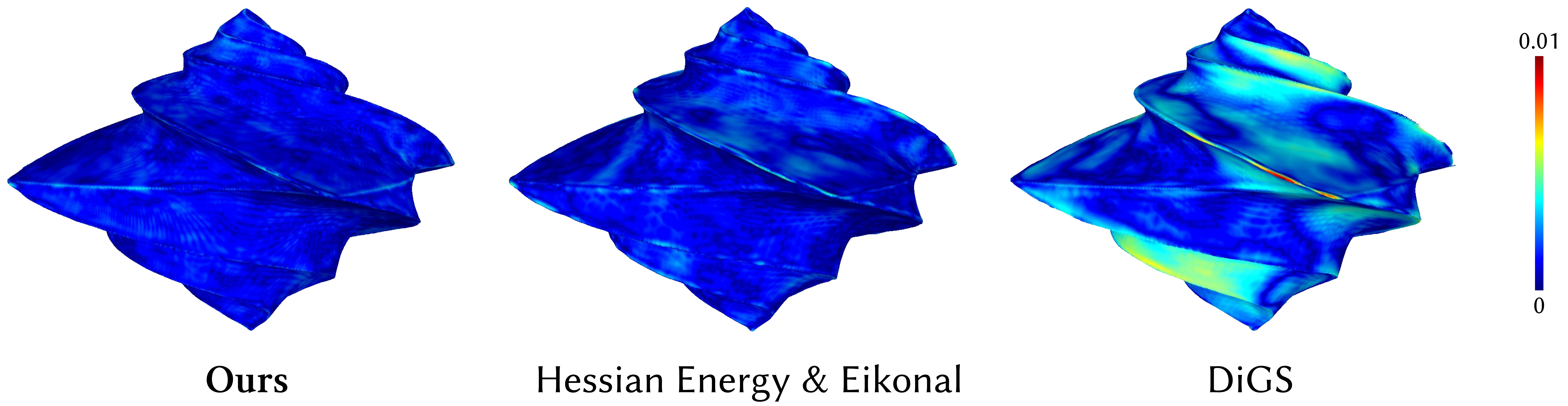}
    \caption{
    The error colormaps of the octa-flower (10K points) reconstructed by our method, Hessian energy, and DiGS~\cite{DiGS} indicate that our method has smaller error near sharp edges.    
    % Error colormaps of the octa-flower~(10K points) reconstructed by ours, Hessian energy, and DiGS~\cite{DiGS}. It indicates that our method has a lower error near the sharp edges.
    }
    \label{fig:abl_sharp}
\end{figure}

\paragraph{Sharp Features}
We utilize the octa-flower model (comprising 10K points) to evaluate our method’s capacity to preserve sharp edges (as depicted in Fig.~\ref{fig:abl_sharp}). The error colormaps demonstrate that our approach surpasses both the Hessian energy method and DiGS~\cite{DiGS}, which employs Laplacian energy. The primary cause of their suboptimal performance is the smooth energy and original Eikonal condition, which result in gradient inaccuracies near sharp edges.

% We use the octa-flower~(10K points) to test the ability
% of our method to retain sharp edges~(Fig.~\ref{fig:abl_sharp}). 
% The error colormaps show that our method outperforms Hessian energy and DiGS~\cite{DiGS} which is based on Laplacian energy.
% The main reason for their unsatisfying performance is the smooth energy and original Eikonal condition leading to the inaccuracy of gradient near the sharp edges.

% \paragraph{Different activation functions}
% It is interesting to explore our loss with other activation.
% Here we combined $L_{\text{singularH}}$ to IGR~\cite{IGR} and NeuS~\cite{wang2021neus}, which leverage SoftPlus serving as the smooth version of ReLU to show the power of our loss.
% Fig.~\ref{} shows that our $L_{\text{singularH}}$ is also useful for the SoftPlus network and benefits other implicit neural representations works, including point cloud input and multi-image input.

\begin{figure}[!htp]
\centering
\begin{overpic}[width=\linewidth]{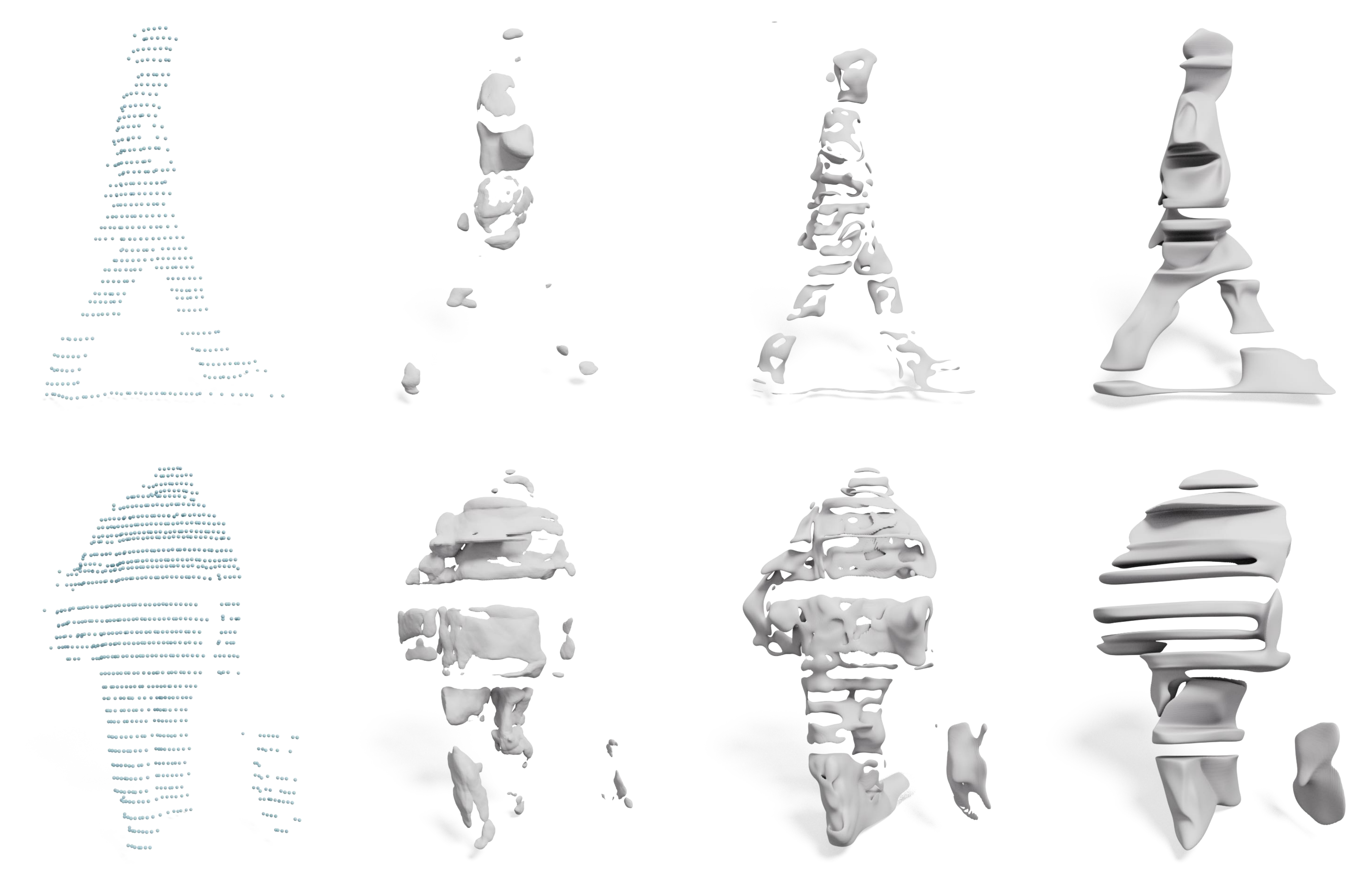}
    
    \put(15, -3.0){Input}

    \put(37, -3.0){OSP}

    \put(60, -3.0){DiGS}

    \put(83, -3.0){\textbf{Ours}}
\end{overpic}
\caption{
    The visual comparison under KITTI~\cite{KITTI}.
    }
    \label{fig:kitti}
\end{figure}

\section{Limitations}
\paragraph{LiDAR Input.}
In its present state, Neural-Singular-Hessian is not well-suited for processing LiDAR input with unique point distributions.
Unlike the point clouds from DoF or structural light cameras, LiDAR has its unique distribution characteristics, which include stripe distribution, sparsity, and a substantial presence of missing parts. In particular, most data derived from the KITTI dataset are partial scans. This presents significant challenges in closing the gaps inherent between the stripes and completing large missing parts.
Our results, in comparison with DiGS~\cite{DiGS} and OSP~\cite{OnSurface}~(specifically designed to handle sparse point clouds), are presented in Fig.~\ref{fig:kitti} within the context of the KITTI dataset~\cite{KITTI}. 
Despite its advantages, our algorithm is unable to effectively process such data, resulting in discrepancies between different levels of points.
%ur data term tends to interpolate each input point leading to a gap between different levels of points.

% \paragraph{Scene-level Reconstruction.}
% Note that our framework has two steps of refining normal vectors. It is necessary to test
% if our algorithm highly depends on normal consistency and accuracy. In Fig. 23, the top row shows the augmented edge points
% when one randomly flips 5%, 10% and 20% normal vectors, while the
% bottom row shows the augmented edge points when one adds white

% \begin{figure}
%     \centering
%     \includegraphics[width=\linewidth]{figures/2dvis_supp.pdf}
%     \caption{More 2D visualization compared with SIREN~\cite{SIREN} and DiGS. Our method results in reliable results even with sparse 100 points input.}
%     \label{fig:2dvis_supp}
% \end{figure}
% \subsection{More visualization in 2D}
% We show more visualization in 2D to demonstrate the benefits of our loss.
% From Fig.~\ref{fig:2dvis_supp}, we can find the unnecessary critical points nearing the original curve are damaging the geometry. But our method effectively suppresses the occurrence of critical points leading to faithful results.

\begin{figure}[!htp]
    \centering
    \includegraphics[width=\linewidth]{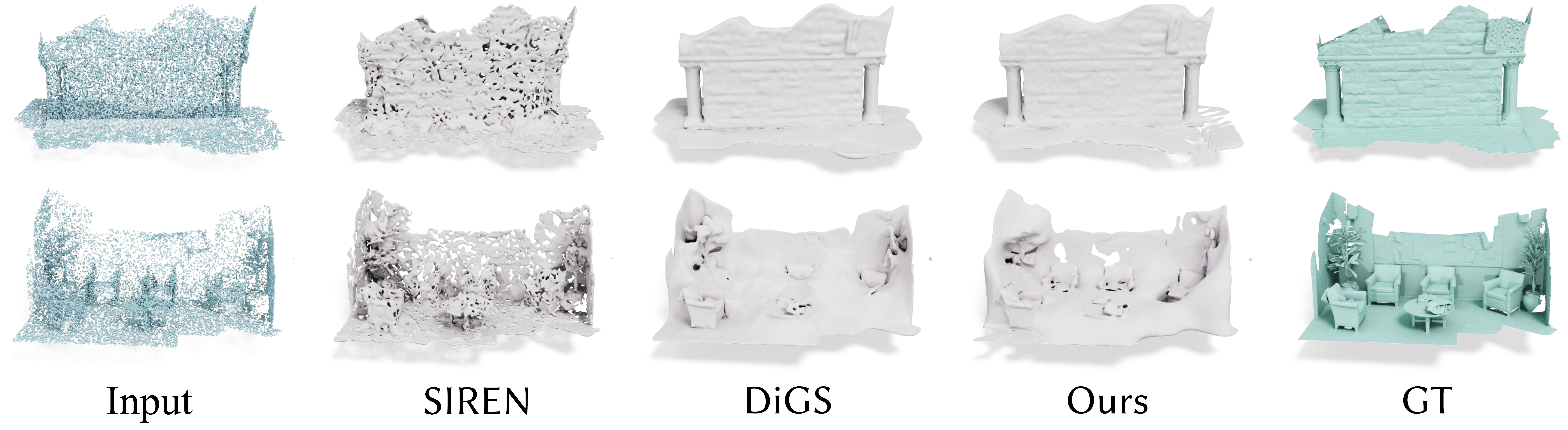}
    \caption{Scene-level reconstruction under 3DScene~\cite{3dscene}.}
    \label{fig:scene}
\end{figure}

\paragraph{Scene-level Reconstruction}
We also try our method to reconstruct the scene.
We propose experiments under 3DScene~\cite{3dscene} with 20K points for each scene.
The results in Fig.~\ref{fig:scene} show that it seems existing methods leveraging sine activation function without normal are weak in handling scene data due to the planes~(wall, floor).
It is interesting to extend our method for scenes in the future.

\section{Experimental Details}
\subsection{Evaluation metrics}
To compare the performance of different reconstruction methods, we use the same evaluation metrics as ConvONet~\cite{Peng20ConvONet}, i.e., Chamfer distances, F-Score, and Normal consistency.
We denote $M_g$ and $M_p$ as the ground-truth mesh (or point cloud) and the mesh of the predicted result, respectively. 
Let $P_1$ and $P_2$ be the randomly sampled points on
the ground-truth mesh (or point cloud) and the predicted mesh.
% these two meshes (or point clouds), respectively. 

\paragraph{Chamfer Distance} The Chamfer distance between two point clouds $P_1$, $P_2$ is defined as follows:

\begin{equation}
\begin{aligned}
\operatorname{Chamfer}\left(P_1, P_2\right)= & \frac{1}{2|P 1|} \sum_{p_1 \in P_1} \min _{p_2 \in P_2} d(p_1, p_2) \\
& +\frac{1}{2|P 2|} \sum_{p_2 \in P_2} \min _{p_1 \in P_1} d(p_1, p_2),
\end{aligned}
\end{equation}
where $d(p_1, p_2)$ is the straight-line distance between points $p1$, $p2$. 
We use the $L_1$ norm following ConvONet~\cite{Peng20ConvONet}.

\paragraph{F-Score.} The F-Score between the two point clouds $P_1$ and $P_2$ at a given threshold $t$ is given by:
\begin{equation}
\operatorname{F-Score}\left(t, P_1, P_2\right)=\frac{2 \text { Recall Precision }}{\text { Recall }+\text { Precision }},
\end{equation}
where
\begin{equation}
\begin{aligned}
\operatorname{Recall}\left(t, P_1, P_2\right)=\left|\left\{p_1 \in P_1 \text {, s.t. } \min _{p_2 \in P_2} d\left(p_1, p_2\right)<t\right\} \right| \\
\text { Precision }\left(t, P_1, P_2\right)=\left|\left\{p_2 \in P_2 \text {, s.t. } \min _{p_1 \in P_1} d\left(p_2, p_1\right)<t\right\}  \right|
\end{aligned}
\end{equation}

\paragraph{Normal consistency} The normal consistency between two point clouds $P_1$, $P_2$ is defined as follows:
\begin{equation}
\begin{aligned}
\mathrm{Normal C.} \left(P_1, P_2\right)= & \frac{1}{2|P 1|} \sum_{p_1 \in P_1} n_{p_1} \cdot n_{\text {closest }\left(p_1, P_2\right)} \\
& +\frac{1}{2|P 2|} \sum_{p_2 \in P_2} n_{p_2} \cdot n_{\text {closest }\left(p_2, P_1\right)},
\end{aligned}
\end{equation}
where
\begin{equation}
\operatorname{closest}(p, P)=\underset{p^{\prime} \in P}{\arg \min }\ d\left(p, p^{\prime}\right)
\end{equation}

\subsection{Surface reconstruction on SRB}

%\paragraph{Implementation details} 

We report the results of baselines using their official source code.
All methods leverage $256^3$ grids (SPSR~\cite{SPR} and iPSR~\cite{iPSR} use the octree of depth 8) to extract the final mesh.
We trained DiGS and SIREN with four hidden layers, each layer containing 256 units,
and the total number of iterations is set to 10K same as ours.
%for 10K iterations for DiGS and SIREN, same as ours.
%Other parameters for each method are used with default settings.
More parameters of each method are used with their default settings.

\begin{figure*}
    \centering
    \includegraphics[width=\textwidth]{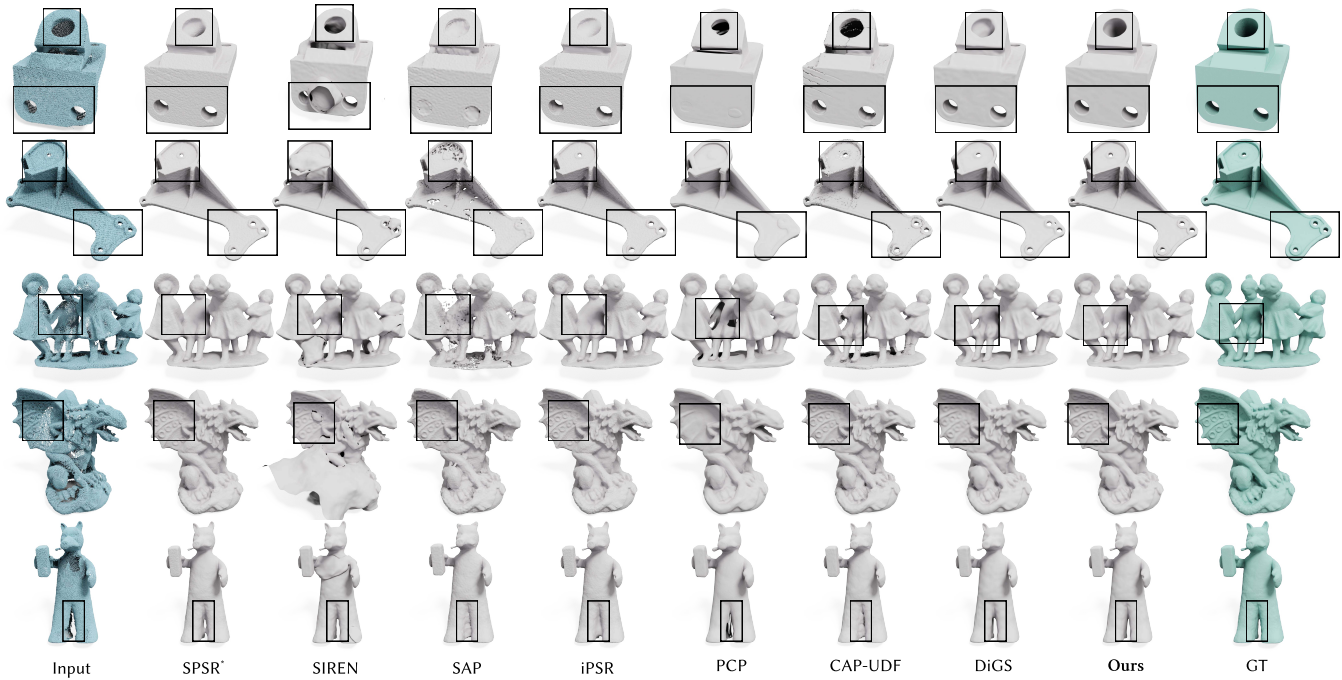}
    \caption{Visual comparison about various approaches on the Surface Reconstruction Benchmark dataset~\cite{DGP}.}
    \label{fig:srb_supp}
\end{figure*}

%\paragraph{Additional quantitative results}
% We provide additional quantitative results for surface reconstruction on the Surface Reconstruction Benchmark~\cite{DGP}. 
In Tab.~\ref{tab:add_srb}, we provide the relevant comparison statistics on the Surface Reconstruction Benchmark~\cite{DGP}. 
% Tab.~\ref{tab:add_srb} shows errors for each shape. 
% Our method achieves the best performance for each shape compared with the baselines, except for Daratech.
It can be seen that our method achieves the best score on all the shapes except  Daratech. Fig.~\ref{fig:srb_supp} shows the visual comparison. 
% \paragraph{Additional qualitative results} 
% We provide additional qualitative results for surface reconstruction on the Surface Reconstruction Benchmark. Fig.~\ref{fig:srb_supp} shows visualizations
% of the output reconstruction of different methods.

\begin{table*}
\centering
\caption{Comparison on Surface Reconstruction Benchmark.}
\label{tab:add_srb}
\resizebox{\textwidth}{!}{%
\begin{tabular}{l|cc|cc|cc|cc|cc|cc|cc} 
\toprule
              & \multicolumn{2}{c|}{Mean}                 & \multicolumn{2}{c}{Std.} & \multicolumn{2}{c|}{Anchor}               & \multicolumn{2}{c|}{Daratech}             & \multicolumn{2}{c|}{DC}                   & \multicolumn{2}{c|}{Gargoyle}             & \multicolumn{2}{c}{Lord Quas}              \\
              & Chamfer~$\downarrow$ & F-Score~$\uparrow$ & Chamfer & F-Score~       & Chamfer~$\downarrow$ & F-Score~$\uparrow$ & Chamfer~$\downarrow$ & F-Score~$\uparrow$ & Chamfer~$\downarrow$ & F-Score~$\uparrow$ & Chamfer~$\downarrow$ & F-Score~$\uparrow$ & Chamfer~$\downarrow$ & F-Score~$\uparrow$  \\ 
% \cmidrule{1-3}\cline{4-5}\cmidrule{6-15}
\midrule
SPSR*~\footnotesize{\cite{SPR}}        & 4.36                 & 75.87              & 1.56    & 18.57          & 6.93                 & 46.14              & 4.20                 & 83.22              & 3.40                 & 85.89              & 4.37                 & 70.65              & 2.85                 & 93.60               \\ 
DGP*~\footnotesize{\cite{DGP}}        & 4.87                 & 73.34              & 1.64    & 18.56          & 7.56                 & 43.04              & 3.85                 & 83.05              & 4.85                 & 78.79              & 4.84                 & 70.40              & 3.26                 & 91.45               \\ 
\midrule
% \cmidrule{1-3}\cline{4-5}\cmidrule{6-15}
SIREN~\footnotesize{\cite{SIREN}}        & 18.24                & 38.74              & 17.09   & 31.26          & 38.31                & 5.05               & 6.19                 & 52.30              & 46.24                & 75.47              & 35.50                & 7.25               & 6.53                 & 54.58               \\
SAP~\footnotesize{\cite{SAP}}          & 6.19                 & 57.21              & 1.75    & \textbf{11.66}          & 8.33                 & 46.73              & 7.76                 & 48.42              & 5.11                 & 60.34              & 4.27                 & 75.66              & 5.54                 & 54.61               \\
iPSR~\footnotesize{\cite{iPSR}}         & 4.54                 & 75.07              & 1.78    & 19.18          & 7.53                 & 44.29              & 4.20                 & 83.51              & 3.52                 & 84.36              & 4.49                 & 69.87              & 2.91                 & 93.53               \\
PCP~\footnotesize{\cite{PCP}}          & 6.53                 & 47.97              & 1.75    & 14.50          & 9.04                 & 37.63              & 7.23                 & 36.08              & 5.82                 & 45.09              & 6.17                 & 49.71              & 4.30                 & 72.09               \\
CAP-UDF~\footnotesize{\cite{CAP-UDF}}      & 4.54                 & 74.75              & 1.82    & 18.84          & 7.68                 & 43.92              & 3.96                 & 82.78              & 3.61                 & 84.03              & 4.40                 & 70.82              & 3.06                 & 92.19               \\
DiGS~\footnotesize{\cite{DiGS}}         & 4.16                 & 76.69              & 1.44    & 18.15          & 6.63                 & 46.52              & \textbf{3.62}        & \textbf{85.54}     & 3.32                 & 86.11              & 4.19                 & 73.34              & 3.04                 & 91.86               \\
\textbf{Ours} & \textbf{3.76}        & \textbf{81.38}     & \textbf{0.98}    & 13.73          & \textbf{5.31}        & \textbf{59.32}     & 3.75                 & 83.89              & \textbf{3.28}        & \textbf{87.05}     & \textbf{3.84}        & \textbf{80.09}     & \textbf{2.64}        & \textbf{96.46}      \\
\bottomrule
\end{tabular}
}
\end{table*}

\subsection{Surface reconstruction on ShapeNet}
%\paragraph{Implementation details} 
We report all baselines using their code.
% All methods leverage the $256^3$ grid (SPSR~\cite{SPR}, iPSR~\cite{iPSR} and PGR~\cite{PGR} use octree depth 8) to extract mesh.
All methods leverage $256^3$ grids (SPSR~\cite{SPR}, iPSR~\cite{iPSR}, and PGR~\cite{PGR} use the octree of depth 8) to extract the final mesh.
We trained DiGS and SIREN with four hidden layers, each layer containing 256 units.
The total number of iterations is set to 10K. 
We conduct 10K iterations for DiGS and SIREN, the same as ours,
and train SAL and IGR within 20K iterations and 15K iterations, respectively.
% For SAL and IGR, we trained them with 20K and 15K iterations, respectively.
% We trained using a SIREN network with four hidden layers, each containing 256 neurons for 10K iterations for DiGS and SIREN, the same as ours.
For NSP~\cite{NSP}, we follow the parameters used in its main paper (1024 input points with 1024 Nyström samples and no regularization) and set the Nyström samples to 1000 and 3000 for 1K and 3K input points, respectively, without regularization.%\SQ{what parameters?}\ZX{Nyström samples}
For PGR, we use the officially recommended parameters for sparse inputs (alpha: 2, wmin: 0.04). %\SQ{alpha? wmin?}\ZX{The parameters PGR used}
For the supervision methods SAP~\cite{SAP} and POCO~\cite{POCO}, we retrain them with 1K points and 3K points under ShapeNet~\cite{ShapeNet}, respectively.
More parameters of each method follow the default setting.

\begin{table*}
\centering
\caption{Class-by-class comparison of the surface reconstruction quality
on 1K-point clouds of ShapeNet.
}
\label{tab:add_shapenet_1k}
\resizebox{\linewidth}{!}{%
\begin{tabular}{l|l|ccccccccccccc|cc} 
\toprule
                                       &               & airplane       & bench          & cabinet        & car            & chair          & display        & lamp           & loudspeaker    & rifle          & sofa           & table          & telephone      & watercraft     & mean           & \multicolumn{1}{l}{std.}  \\ 
\midrule
\multirow{12}{*}{Normal C.~$\uparrow$} & SPSR$^*$~\footnotesize{\cite{SPR}}     & 90.79          & 87.45          & 92.86          & 91.30          & 90.14          & 94.32          & 91.06          & 93.79          & 94.18          & 91.58          & 88.97          & 96.87          & 91.46          & 91.89          & 4.76                      \\
                                       & NSP$^*$~\footnotesize{\cite{NSP}}          & 81.87          & 79.43          & 89.33          & 89.49          & 83.16          & 90.42          & 85.70          & 91.50          & 89.22          & 87.01          & 82.38          & 94.43          & 87.74          & 87.05          & 6.05                      \\ 
\cmidrule{2-17}
                                       & SAL~\footnotesize{\cite{SAL}}          & 75.33          & 73.27          & 89.04          & 88.55          & 75.20          & 88.61          & 79.65          & 93.75          & 80.43          & 85.85          & 73.44          & 92.25          & 82.12          & 82.99          & 11.11                     \\
                                       & IGR~\footnotesize{\cite{IGR}}          & 74.77          & 73.65          & 84.59          & 84.90          & 76.02          & 70.67          & 82.11          & 92.46          & 77.95          & 85.32          & 73.03          & 70.22          & 82.92          & 79.26          & 12.27                     \\
                                       & SIREN~\footnotesize{\cite{SIREN}}        & 84.70          & 76.29          & 74.42          & 75.44          & 80.57          & 83.56          & 85.47          & 73.42          & 83.78          & 72.63          & 78.82          & 88.86          & 80.94          & 79.91          & 8.87                      \\
                                       & DiGS~\footnotesize{\cite{DiGS}}         & 94.12          & 89.76          & 91.32          & 90.06          & 90.79          & 94.90          & 93.19          & 92.80          & 96.08          & 90.44          & 89.48          & 98.12          & 93.70          & 92.67          & 6.03                      \\
                                       & OSP~\footnotesize{\cite{OnSurface}}          & 92.16          & 85.66          & 90.72          & 90.64          & 91.67          & 94.19          & 91.38          & 93.48          & 94.17          & 91.13          & 92.26          & 96.91          & 90.24          & 91.89          & 5.52                      \\
                                       & iPSR~\footnotesize{\cite{iPSR}}         & 83.41          & 78.82          & 91.28          & 89.93          & 83.86          & 89.47          & 88.94          & 92.72          & 92.87          & 88.73          & 79.62          & 94.00          & 88.61          & 87.88          & 7.26                      \\
                                       & PGR~\footnotesize{\cite{PGR}}          & 83.16          & 83.91          & 91.80          & 90.25          & 88.21          & 93.24          & 88.06          & 93.29          & 89.36          & 91.02          & 87.13          & 95.92          & 88.54          & 89.53          & 5.35                      \\ 
\cmidrule{2-17}
                                       & SAP$^+$~\footnotesize{\cite{SAP}}      & 94.52          & \textbf{92.53} & 96.17          & 92.41          & 95.17          & 97.35          & 93.46          & \textbf{95.15} & 95.22          & 95.41          & \textbf{95.66} & 98.44          & 92.51          & 94.92          & \textbf{3.60}             \\
                                       & POCO$^+$~\footnotesize{\cite{POCO}}     & 93.65          & 91.96          & \textbf{96.20} & 91.29          & \textbf{95.23} & 97.38          & 93.18          & 95.06          & \textbf{95.91} & 95.77          & 95.59          & 98.62          & 92.45          & 94.79          & 4.15                      \\ 
\cmidrule{2-17}
                                       & \textbf{Ours} & \textbf{96.04} & 92.20          & 94.92          & \textbf{93.17} & 94.93          & \textbf{97.77} & \textbf{94.75} & 93.58          & 92.86          & \textbf{96.01} & 95.81          & \textbf{98.83} & \textbf{95.41} & \textbf{95.10} & 4.04                      \\ 
\midrule
\multirow{12}{*}{Chamfer~$\downarrow$} & SPSR$^*$~\footnotesize{\cite{SPR}}     & 6.09           & 10.29          & 9.63           & 9.53           & 12.34          & 9.81           & 10.35          & 9.05           & 3.50           & 10.52          & 14.26          & 6.33           & 9.86           & 9.35           & 7.66                      \\
                                       & NSP$^*$~\footnotesize{\cite{NSP}}          & 20.84          & 13.32          & 12.84          & 7.74           & 17.47          & 10.30          & 14.13          & 12.91          & 4.40           & 12.18          & 19.74          & 6.19           & 10.46          & 12.51          & 7.29                      \\
\cmidrule{2-17}
                                       & SAL~\footnotesize{\cite{SAL}}          & 64.90          & 56.83          & 25.53          & 20.98          & 92.43          & 30.14          & 99.07          & 20.59          & 38.40          & 28.57          & 71.88          & 24.37          & 36.80          & 47.46          & 50.57                     \\
                                       & IGR~\footnotesize{\cite{IGR}}          & 13.20          & 98.42          & 36.03          & 55.19          & 52.71          & 10.54          & 65.54          & 20.19          & 11.22          & 44.93          & 78.41          & 13.10          & 78.88          & 77.68          & 59.55                     \\
                                       & SIREN~\footnotesize{\cite{SIREN}}        & 27.12          & 49.85          & 26.99          & 44.64          & 17.21          & 41.93          & 28.31          & 38.66          & 84.89          & 38.38          & 25.90          & 33.19          & 37.39          & 38.04          & 46.02                     \\
                                       & DiGS~\footnotesize{\cite{DiGS}}         & 4.17           & 6.37           & 10.72          & 7.39           & 8.70           & 6.29           & 5.59           & 9.90           & 2.53           & 10.04          & 11.15          & 3.22           & 5.11           & 7.01           & 5.52                      \\
                                       & OSP~\footnotesize{\cite{OnSurface}}          & 7.36           & 8.89           & 9.81           & 10.29          & 9.49           & 9.37           & 7.57           & 8.93           & 5.57           & 9.18           & 8.56           & 6.02           & 12.97          & 8.77           & 6.76                      \\
                                       & iPSR~\footnotesize{\cite{iPSR}}         & 13.98          & 21.38          & 10.82          & 11.03          & 18.84          & 12.56          & 11.93          & 10.34          & 3.95           & 13.65          & 22.76          & 7.31           & 12.41          & 13.16          & 11.78                     \\
                                       & PGR~          & 10.13          & 11.59          & 10.54          & 10.85          & 11.23          & 8.72           & 13.99          & 10.12          & 6.11           & 11.26          & 13.25          & 6.69           & 11.90          & 10.49          & 6.52                      \\ 
\cmidrule{2-17}
                                       & SAP$^+$~\footnotesize{\cite{SAP}}      & 3.36           & \textbf{3.78}  & 4.58           & 6.19           & \textbf{4.71}  & 3.63           & 4.20           & 6.55           & 2.62           & 4.94           & 5.66           & 2.96           & 7.16           & 4.64           & 3.71                      \\
                                       & POCO$^+$~\footnotesize{\cite{POCO}}     & 4.06           & 4.42           & \textbf{4.48}  & 6.60           & 4.93           & 3.73           & 4.11           & \textbf{5.84}  & \textbf{2.25}  & 4.31           & 5.86           & \textbf{2.33}  & 6.03           & 4.53           & 4.05                      \\ 
\cmidrule{2-17}
                                       & \textbf{Ours} & \textbf{2.66}  & 4.06           & 5.57           & \textbf{4.68}  & 5.25           & \textbf{3.36}  & \textbf{3.45}  & 9.19           & 2.44           & \textbf{4.28}  & \textbf{4.69}  & 2.34           & \textbf{3.33}  & \textbf{4.26}  & \textbf{3.11}             \\ 
\midrule
\multirow{12}{*}{F-Score~$\uparrow$}   & SPSR$^*$~\footnotesize{\cite{SPR}}      & 54.28          & 36.69          & 39.63          & 49.11          & 34.55          & 38.94          & 51.89          & 44.16          & 76.85          & 38.52          & 23.76          & 71.35          & 50.08          & 46.91          & 26.06                     \\
                                       & NSP$^*$~\footnotesize{\cite{NSP}}           & 33.69          & 29.18          & 21.13          & 44.77          & 19.17          & 36.16          & 39.69          & 28.72          & 69.88          & 26.08          & 18.25          & 59.65          & 43.91          & 36.17          & 21.56                     \\ 
\cmidrule{2-17}
                                       & SAL~\footnotesize{\cite{SAL}}           & 6.79           & 9.88           & 20.30          & 24.47          & 7.92           & 24.92          & 8.48           & 23.80          & 13.67          & 25.31          & 7.04           & 53.29          & 23.07          & 18.16          & 19.15                     \\
                                       & IGR~\footnotesize{\cite{IGR}}           & 0.72           & 11.84          & 41.50          & 32.34          & 23.48          & 14.61          & 29.34          & 48.08          & 11.89          & 40.58          & 11.32          & 12.73          & 24.33          & 22.48          & 32.08                     \\
                                       & SIREN~\footnotesize{\cite{SIREN}}         & 34.92          & 23.14          & 18.97          & 14.56          & 32.50          & 22.94          & 40.34          & 14.60          & 30.34          & 16.00          & 22.52          & 32.29          & 22.16          & 25.02          & 23.52                     \\
                                       & DiGS~\footnotesize{\cite{DiGS}}          & 69.85          & 61.19          & 38.32          & 54.62          & 45.74          & 57.97          & 70.23          & 41.02          & 88.13          & 39.74          & 38.38          & 87.67          & 70.44          & 58.72          & 29.77                     \\
                                       & OSP~\footnotesize{\cite{OnSurface}}           & 40.72          & 51.27          & 46.06          & 35.57          & 42.16          & 47.60          & 58.39          & 43.93          & 47.13          & 52.54          & 43.26          & 81.23          & 28.54          & 47.57          & 23.45                     \\
                                       & iPSR~\footnotesize{\cite{iPSR}}          & 27.35          & 24.04          & 32.92          & 43.52          & 24.99          & 33.81          & 50.28          & 39.66          & 72.93          & 35.79          & 18.29          & 64.51          & 43.37          & 39.36          & 25.11                     \\
                                       & PGR~\footnotesize{\cite{PGR}}           & 35.49          & 30.33          & 35.22          & 40.82          & 31.70          & 37.00          & 34.25          & 38.51          & 52.15          & 29.30          & 22.80          & 65.87          & 35.51          & 37.61          & \textbf{19.14}            \\ 
\cmidrule{2-17}
                                       & SAP$^+$~\footnotesize{\cite{SAP}}       & 82.48          & 78.55          & 70.19          & 67.54          & 72.31          & 80.75          & 77.30          & 51.80          & 88.26          & 73.05          & 54.48          & 93.69          & 65.10          & 73.50          & 25.02                     \\
                                       & POCO$^+$~\footnotesize{\cite{POCO}}      & 78.56          & 74.93          & 74.35          & 68.06          & 72.23          & 83.71          & 80.48          & \textbf{60.07} & \textbf{91.82} & 77.32          & 57.29          & 94.48          & 68.97          & 75.56          & 26.46                     \\ 
\cmidrule{2-17}
                                       & \textbf{Ours} & \textbf{90.32} & \textbf{79.93} & \textbf{74.58} & \textbf{75.02} & \textbf{72.28} & \textbf{89.76} & \textbf{85.28} & 54.87          & 89.45          & \textbf{80.99} & \textbf{74.13} & \textbf{96.03} & \textbf{83.95} & \textbf{80.51} & 21.48                     \\
\bottomrule
\end{tabular}
}
\end{table*}

\begin{figure*}[!htp]
    \centering
    \includegraphics[width=\textwidth]{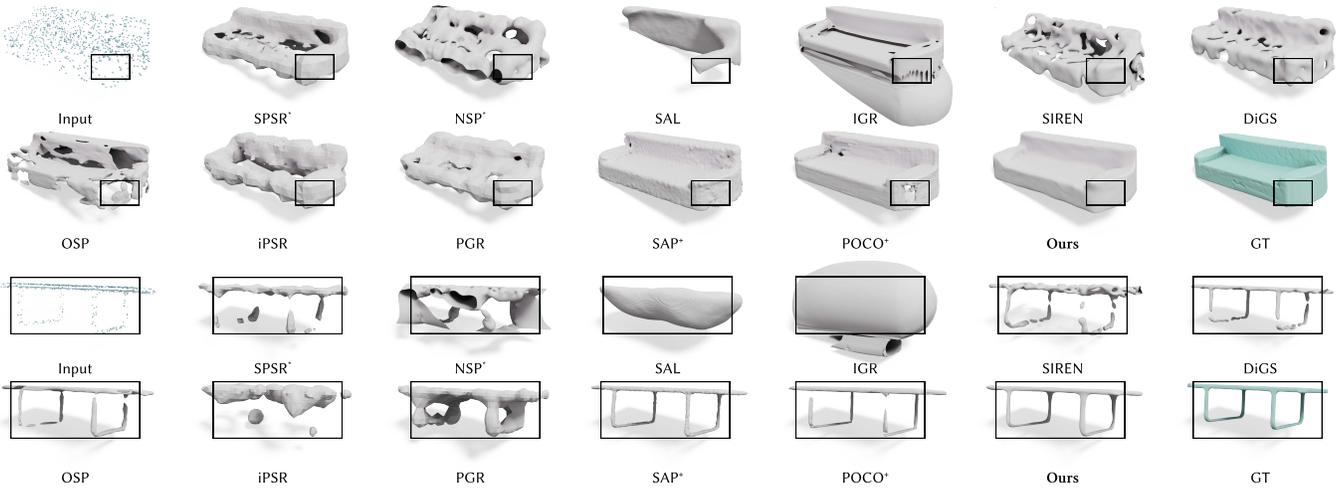}
    \caption{Visual comparison on ShapeNet~\cite{ShapeNet},
    where the input has 1K points.}
    \label{fig:shapenet_supp_1k}
\end{figure*}

\begin{table*}
\centering
\caption{
Class-by-class comparison of the surface reconstruction quality
on 3K-point clouds of ShapeNet.
}
\label{tab:add_shapenet_3k}
\resizebox{\linewidth}{!}{%
\begin{tabular}{l|l|ccccccccccccc|cc} 
\toprule
                                       &        & airplane       & bench          & cabinet        & car            & chair          & display        & lamp           & loudspeaker    & rifle          & sofa           & table          & telephone      & watercraft     & mean           & \multicolumn{1}{l}{std.}  \\ 
\midrule
\multirow{12}{*}{Normal C.~$\uparrow$}  & SPSR$^*$~\footnotesize{\cite{SPR}}   & 95.30          & 92.85          & 95.98          & 93.43          & 95.02          & 97.35          & 95.03          & 96.19          & 96.93          & 95.53          & 94.65          & 98.67          & 94.61          & 95.50          & 3.30                      \\
                                       & NSP$^*$~\footnotesize{\cite{NSP}}    & 86.04          & 85.37          & 91.30          & 91.08          & 87.83          & 93.70          & 90.45          & 92.95          & 94.94          & 90.18          & 87.66          & 96.87          & 91.30          & 90.74          & 5.48                      \\ 
% \cline{2-17}
\cmidrule{2-17}
                                       & SAL~\footnotesize{\cite{SAL}}    & 77.52          & 78.04          & 90.64          & 90.32          & 79.57          & 91.74          & 86.09          & 94.60          & 86.35          & 90.44          & 77.05          & 97.59          & 87.68          & 86.69          & 9.66                      \\
                                       & IGR~\footnotesize{\cite{IGR}}    & 74.48          & 73.98          & 88.47          & 86.00          & 75.23          & 76.79          & 83.46          & 92.61          & 78.43          & 82.07          & 73.97          & 82.29          & 83.22          & 80.85          & 11.88                     \\
                                       & SIREN~\footnotesize{\cite{SIREN}}  & 88.42          & 80.93          & 78.58          & 77.01          & 84.08          & 89.76          & 84.80          & 78.84          & 84.98          & 80.28          & 86.70          & 90.27          & 84.30          & 83.79          & 10.20                     \\
                                       & DiGS~\footnotesize{\cite{DiGS}}   & 97.19          & 93.86          & 94.61          & 93.18          & 93.98          & 97.50          & 95.59          & 96.05          & \textbf{98.12} & 96.11          & 94.18          & 99.02          & 96.28          & 95.82          & 4.44                      \\
                                       & OSP~\footnotesize{\cite{OnSurface}}    & 94.97          & 91.87          & 95.23          & 92.90          & 95.44          & 97.56          & 93.00          & 95.63          & 92.62          & 95.70          & 95.90          & 97.51          & 93.23          & 94.73          & 3.94                      \\
                                       & iPSR~\footnotesize{\cite{iPSR}}   & 92.45          & 88.80          & 93.26          & 92.51          & 92.64          & 95.35          & 93.70          & 94.54          & 96.39          & 92.41          & 89.66          & 97.56          & 92.63          & 93.22          & 5.26                      \\
                                       & PGR~\footnotesize{\cite{PGR}}    & 85.57          & 86.85          & 94.09          & 91.25          & 91.88          & 95.59          & 90.63          & 95.17          & 91.14          & 93.55          & 91.03          & 97.80          & 90.19          & 91.90          & 4.93                      \\ 
% \cline{2-17}
\cmidrule{2-17}
                                       & SAP$^+$~\footnotesize{\cite{SAP}}    & 96.10          & 94.26          & \textbf{97.36} & 93.79          & 96.93          & 98.01          & 95.35          & 96.84          & 96.14          & 97.34          & 97.09          & 99.05          & 94.01          & 96.33          & 3.24                      \\
                                       & POCO$^+$~\footnotesize{\cite{POCO}}   & 96.81          & 94.23          & 97.28          & 93.52          & 96.56          & 98.29          & 95.92          & \textbf{96.86} & 97.56          & 96.87          & 96.65          & 99.02          & 93.77          & 96.41          & 3.53                      \\ 
% \cline{2-17}
\cmidrule{2-17}
                                       & \textbf{Ours}   & \textbf{97.62} & \textbf{94.77} & 97.24          & \textbf{94.97} & \textbf{97.61} & \textbf{98.53} & \textbf{96.53} & 96.24          & 96.34          & \textbf{97.81} & \textbf{98.03} & \textbf{99.32} & \textbf{96.70} & \textbf{97.05} & \textbf{2.91}             \\ 
% \cmidrule{1-1}\cline{2-17}
\midrule
\multirow{12}{*}{Chamfer~$\downarrow$} & SPSR$^*$~\footnotesize{\cite{SPR}}  & 2.73           & 4.08           & 5.06           & 6.66           & 5.83           & 4.08           & 3.97           & 6.26           & 1.73           & 5.31           & 6.00           & 2.36           & 6.50           & 4.66           & 4.64                      \\
                                       & NSP$^*$~\footnotesize{\cite{NSP}}    & 19.44          & 7.61           & 9.57           & 6.07           & 11.75          & 7.08           & 7.33           & 13.65          & 2.45           & 8.24           & 10.14          & 3.82           & 7.96           & 8.85           & 6.96                      \\
\cmidrule{2-17}
                                       & SAL~\footnotesize{\cite{SAL}}    & 52.97          & 45.43          & 21.19          & 14.25          & 55.97          & 23.83          & 34.35          & 13.93          & 13.33          & 17.25          & 68.27          & 6.54           & 23.29          & 29.98          & 31.86                     \\
                                       & IGR~\footnotesize{\cite{IGR}}    & 12.44          & 69.77          & 34.50          & 24.30          & 58.89          & 91.40          & 55.51          & 19.87          & 68.71          & 46.77          & 75.63          & 83.02          & 60.22          & 62.54          & 48.44                     \\
                                       & SIREN~\footnotesize{\cite{SIREN}}  & 26.12          & 38.23          & 33.62          & 42.33          & 23.52          & 17.84          & 46.18          & 34.34          & 65.53          & 23.49          & 21.36          & 30.22          & 42.08          & 34.19          & 46.77                     \\
                                       & DiGS~\footnotesize{\cite{DiGS}}   & 2.44           & 3.87           & 8.50           & 4.82           & 7.69           & 4.45           & 3.83           & 5.95           & \textbf{1.35}  & 4.50           & 6.63           & 2.36           & 3.23           & 4.59           & 4.94                      \\
                                       & OSP~\footnotesize{\cite{OnSurface}}    & 5.85           & 4.40           & 6.02           & 9.28           & 5.97           & 4.20           & 11.91          & 6.63           & 9.09           & 5.09           & 5.74           & 5.15           & 9.06           & 6.80           & 6.61                      \\
                                       & iPSR~\footnotesize{\cite{iPSR}}   & 4.17           & 6.08           & 6.35           & 7.22           & 6.97           & 5.40           & 4.27           & 7.20           & 2.10           & 7.28           & 7.81           & 4.60           & 7.84           & 5.95           & 5.97                      \\
                                       & PGR~\footnotesize{\cite{PGR}}    & 7.27           & 8.17           & 7.19           & 8.41           & 7.69           & 6.22           & 8.52           & 7.78           & 4.93           & 7.45           & 8.85           & 3.70           & 9.28           & 7.34           & 4.81                      \\ 
% \cline{2-17}
\cmidrule{2-17}
                                       & SAP$^+$~\footnotesize{\cite{SAP}}    & 2.63           & 2.82           & 3.79           & 5.89           & 3.78           & 3.46           & 3.40           & 4.76           & 2.37           & 3.11           & 4.13           & 2.07           & 6.48           & 3.75           & 4.16                      \\
                                       & POCO$^+$~\footnotesize{\cite{POCO}}   & 2.00           & 3.64           & 3.76           & 5.30           & 4.02           & 2.93           & 2.47           & \textbf{4.13}  & 1.37           & 3.91           & 4.99           & 2.18           & 6.40           & 3.62           & 4.21                      \\ 
% \cline{2-17}
\cmidrule{2-17}
                                       & \textbf{Ours}   & \textbf{1.84}  & \textbf{2.61}  & \textbf{3.59}  & \textbf{3.61}  & \textbf{3.77}  & \textbf{2.84}  & \textbf{2.34}  & 7.46           & \textbf{1.35}  & \textbf{2.97}  & \textbf{2.93}  & \textbf{1.84}  & \textbf{2.90}  & \textbf{3.08}  & \textbf{2.64}             \\ 
% \cmidrule{1-1}\cline{2-17}
\midrule
\multirow{12}{*}{F-Score~$\uparrow$}   & SPSR$^*$~\footnotesize{\cite{SPR}} & 90.17          & 76.96          & 67.50          & 69.99          & 65.21          & 81.29          & 79.09          & 56.36          & 95.66          & 70.81          & 61.00          & 94.69          & 69.64          & 75.28          & 25.76                     \\
                                       & NSP$^*$~\footnotesize{\cite{NSP}}   & 50.61          & 46.86          & 38.65          & 57.83          & 30.18          & 54.11          & 60.14          & 28.52          & 91.31          & 44.29          & 33.47          & 81.70          & 63.80          & 52.42          & 28.55                     \\ 
\cmidrule{2-17}
                                       & SAL~\footnotesize{\cite{SAL}}    & 8.73           & 15.46          & 24.86          & 34.22          & 11.06          & 23.62          & 26.27          & 31.98          & 28.40          & 30.07          & 8.37           & 62.90          & 29.98          & 25.76          & 22.43                     \\
                                       & IGR~\footnotesize{\cite{IGR}}   & 1.77           & 10.28          & 56.52          & 47.66          & 12.76          & 15.74          & 31.11          & 60.90          & 4.44           & 34.99          & 14.17          & 21.77          & 29.49          & 26.28          & 35.91                     \\
                                       & SIREN~\footnotesize{\cite{SIREN}}  & 45.30          & 29.34          & 17.52          & 15.28          & 39.76          & 45.30          & 42.28          & 16.07          & 44.51          & 22.96          & 26.04          & 47.63          & 27.59          & 32.34          & 30.13                     \\
                                       & DiGS~\footnotesize{\cite{DiGS}}   & 93.08          & 83.44          & 53.57          & 77.66          & 68.33          & 77.55          & 85.41          & 62.75          & 98.57          & 76.56          & 68.14          & 95.84          & 84.40          & 78.87          & 27.34                     \\
                                       & OSP~\footnotesize{\cite{OnSurface}}   & 43.02          & 73.47          & 65.35          & 38.96          & 57.37          & 80.52          & 57.39          & 55.29          & 41.30          & 75.01          & 61.78          & 84.08          & 35.03          & 59.12          & 25.82                     \\
                                       & iPSR~\footnotesize{\cite{iPSR}}   & 72.59          & 62.60          & 58.69          & 66.97          & 59.01          & 73.75          & 80.01          & 53.79          & 93.43          & 62.92          & 50.46          & 91.29          & 64.98          & 68.42          & 26.36                     \\
                                       & PGR~\footnotesize{\cite{PGR}}    & 44.70          & 42.12          & 46.98          & 57.14          & 46.95          & 57.69          & 48.46          & 50.78          & 62.57          & 43.66          & 38.29          & 84.75          & 44.64          & 51.44          & 23.12                     \\ 
\cmidrule{2-17}
                                       & SAP$^+$~\footnotesize{\cite{SAP}}    & 91.88          & 90.83          & 82.41          & 75.20          & 83.54          & 90.65          & 87.47          & 68.85          & 91.63          & 88.56          & 81.13          & 98.21          & 71.93          & 84.79          & 20.94                     \\
                                       & POCO$^+$~\footnotesize{\cite{POCO}}   & 96.21          & 82.62          & 81.33          & 82.85          & 83.37          & 93.08          & \textbf{93.74} & 72.98          & \textbf{99.30} & 83.13          & 71.51          & 95.31          & 74.99          & 85.42          & 33.13                     \\ 
\cmidrule{2-17}
                                       & \textbf{Ours}   & \textbf{98.75} & \textbf{93.13} & \textbf{89.08} & \textbf{85.80} & \textbf{86.56} & \textbf{94.42} & 92.12          & \textbf{74.05} & 97.69          & \textbf{91.17} & \textbf{89.79} & \textbf{99.49} & \textbf{87.21} & \textbf{90.71} & \textbf{16.28}            \\
\bottomrule
\end{tabular}
}
\end{table*}

\begin{figure*}[!htp]
    \centering
    \includegraphics[width=\textwidth]{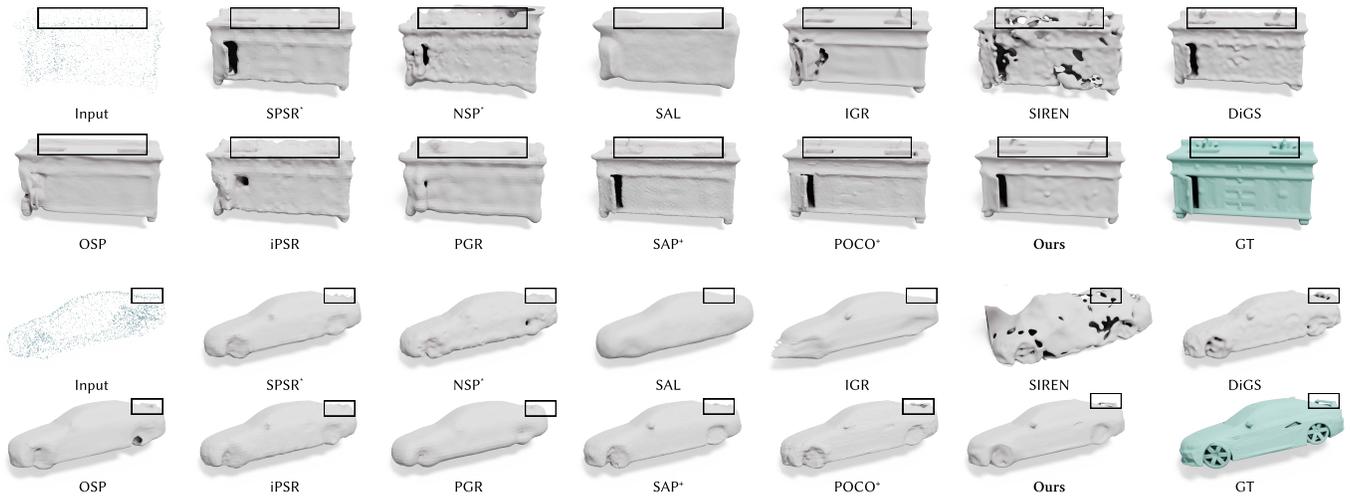}
    \caption{Visual comparison the ShapeNet~\cite{ShapeNet},
    where the input has 3K points.}
    \label{fig:shapenet_supp_3k}
\end{figure*}

%\paragraph{Results}
%Additional quantitative results}
We give the comparison statistics under the settings of 1K points and 3K points in Tab.~\ref{tab:add_shapenet_1k} and Tab.~\ref{tab:add_shapenet_3k}, respectively.
The corresponding visual comparison 
is given in Fig.~\ref{fig:shapenet_supp_1k} and Fig.~\ref{fig:shapenet_supp_3k},
respectively. 
% We show additional visualizations in Fig.~\ref{fig:shapenet_supp_1k} and Fig.~\ref{fig:shapenet_supp_3k}. 
Both qualitative and quantitative comparisons
show that our method can faithfully recover fine geometric details and thin structures, outperforming the other methods.

% As with the summary of all shape classes, we can see that we get better results among almost classes than with other methods.
% %\paragraph{Additional qualitative results}
% Our method can faithfully recover fine geometric details and thin structures.

\subsection{Surface Reconstruction on ABC and Thingi10K}

%\paragraph{Implementation details} 
% We report all baselines using their code.
We report the results of baselines using their source code.
All methods leverage $256^3$ grids, and SPSR~\cite{SPR}, iPSR~\cite{iPSR}, PGR~\cite{PGR}, and Neural Galerkin~\cite{Galerkin} use the depth 8 during the mesh extraction phase. %) to extract mesh.
For SAL~\cite{SAL} and IGR~\cite{IGR}, we trained them with 20K iterations and 15K iterations, respectively.
% We trained a SIREN network with four hidden layers, 
% each containing 256 neurons for 10K iterations for DiGS~\cite{DiGS} and SIREN~\cite{SIREN}, 
% the same as ours.
We conduct 10K iterations for DiGS and SIREN, the same as ours,
where the SIREN network has four hidden layers, 
each containing 256 neurons.
%and train SAL and IGR within 20K iterations and 15K iterations, respectively.
For PGR, we use the officially recommended parameters for the 10K-point input (alpha: 1.2, wk: 16).
For the supervision methods POCO~\cite{POCO} and Neural Galerkin~\cite{Galerkin} (without normals), we retrained them with 10K points under ShapeNet~\cite{ShapeNet} to validate their generalization ability.
Other parameters remain the same with the default settings.

\begin{figure*}[!htp]
    \centering
    \includegraphics[width=\textwidth]{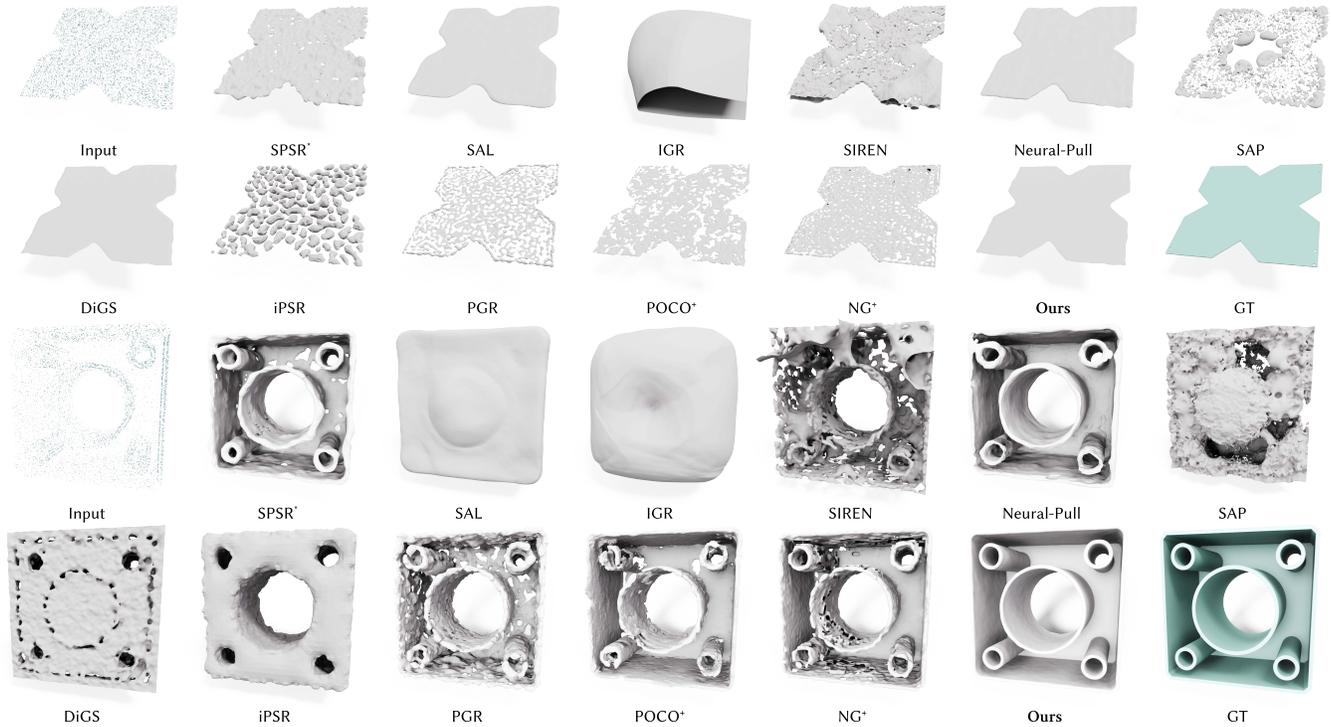}
    \caption{Visual comparison about different approaches on ABC~\cite{ABC} with 10K points. Our method is better at recovering thin geometry features.}
    \label{fig:abc_supp}
\end{figure*}

\begin{figure*}[!htp]
    \centering
    \includegraphics[width=\textwidth]{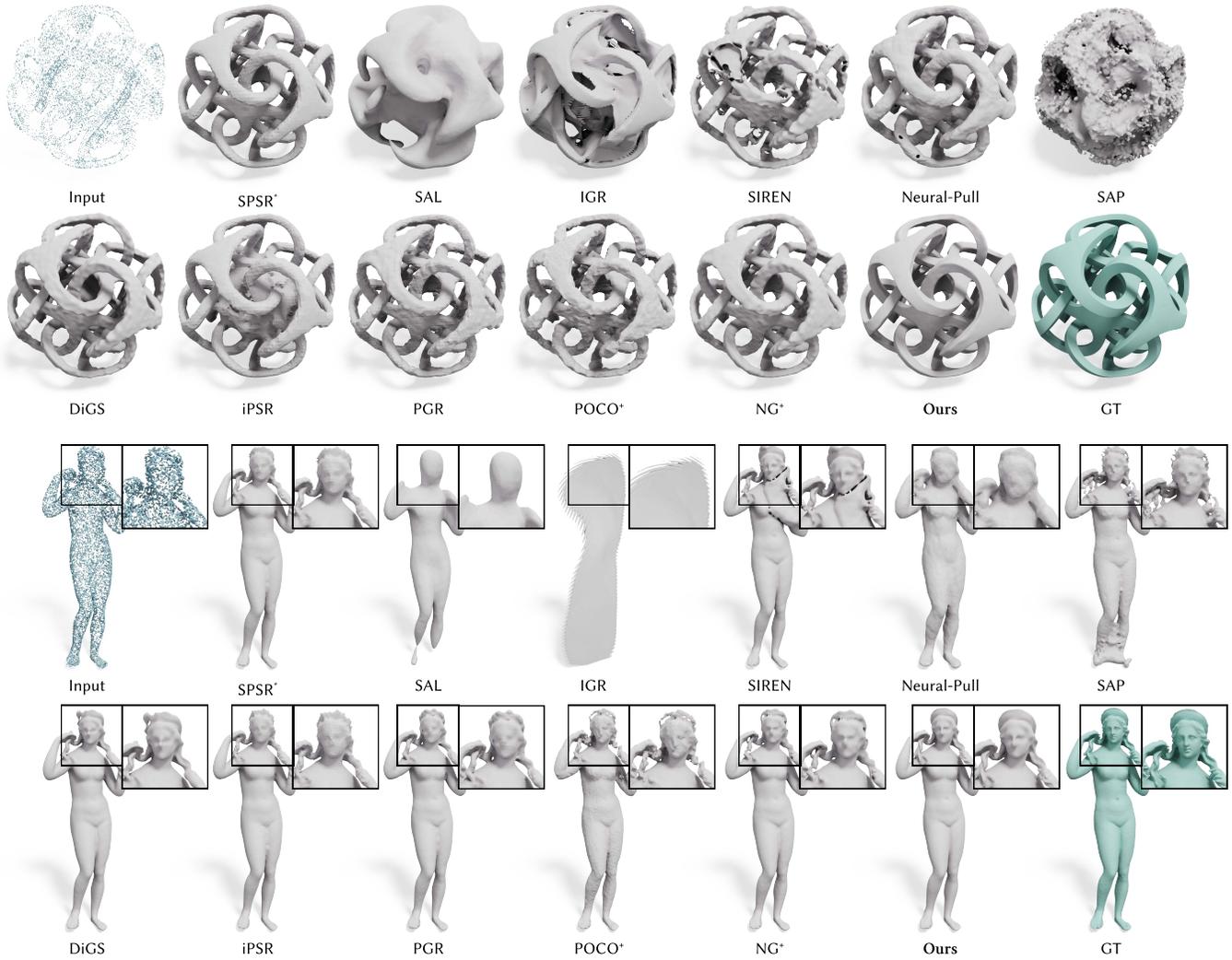}
    \caption{Visual comparison about different approaches on Thingi10K~\cite{ABC} with 10K points. Our method achieves a good trade-off between smoothness and 
    feature preservation. 
    %features~(high-genus and the hairs) preservation.
    }
    \label{fig:thingi_supp}
\end{figure*}

%\paragraph{Results}
We show the visual comparison of different approaches on ABC~\cite{ABC} with 10K points in Fig.~\ref{fig:abc_supp} and Fig.~\ref{fig:thingi_supp}.
The comparison shows that
Our method is better at recovering thin geometry features
and can achieve a good trade-off between smoothness and feature preservation.
%Either thin structures or great details geometry, our method can reliably recover it.

\subsection{Real Scans}

We report the results of baselines using their source code.
Among them, the supervised method Neural Galerkin~\cite{Galerkin} (without normals), and we retrained them with 10K points under ShapeNet~\cite{ShapeNet} to validate its generality. 
All methods leverage $256^3$ grids to extract the mesh.
We conduct 10K iterations for DiGS~\cite{DiGS}, PCP~\cite{PCP}, and ours.
Other parameters remain the same with the default settings.

\subsection{Large Scans}

We report the results of baselines using their source code.
All methods leverage $512^3$ grids to extract the mesh.
We conduct 50K iterations for DiGS and SIREN, where the SIREN network has four hidden layers, each containing 256 neurons, the same as ours,
For the supervision method Neural Galerkin~\cite{Galerkin} (without normals), we retrained them with 10K points under ShapeNet~\cite{ShapeNet} to validate its generality.
Other parameters remain the same with the default settings.

\begin{figure*}[!htp]
    \centering
    \includegraphics[width=\textwidth]{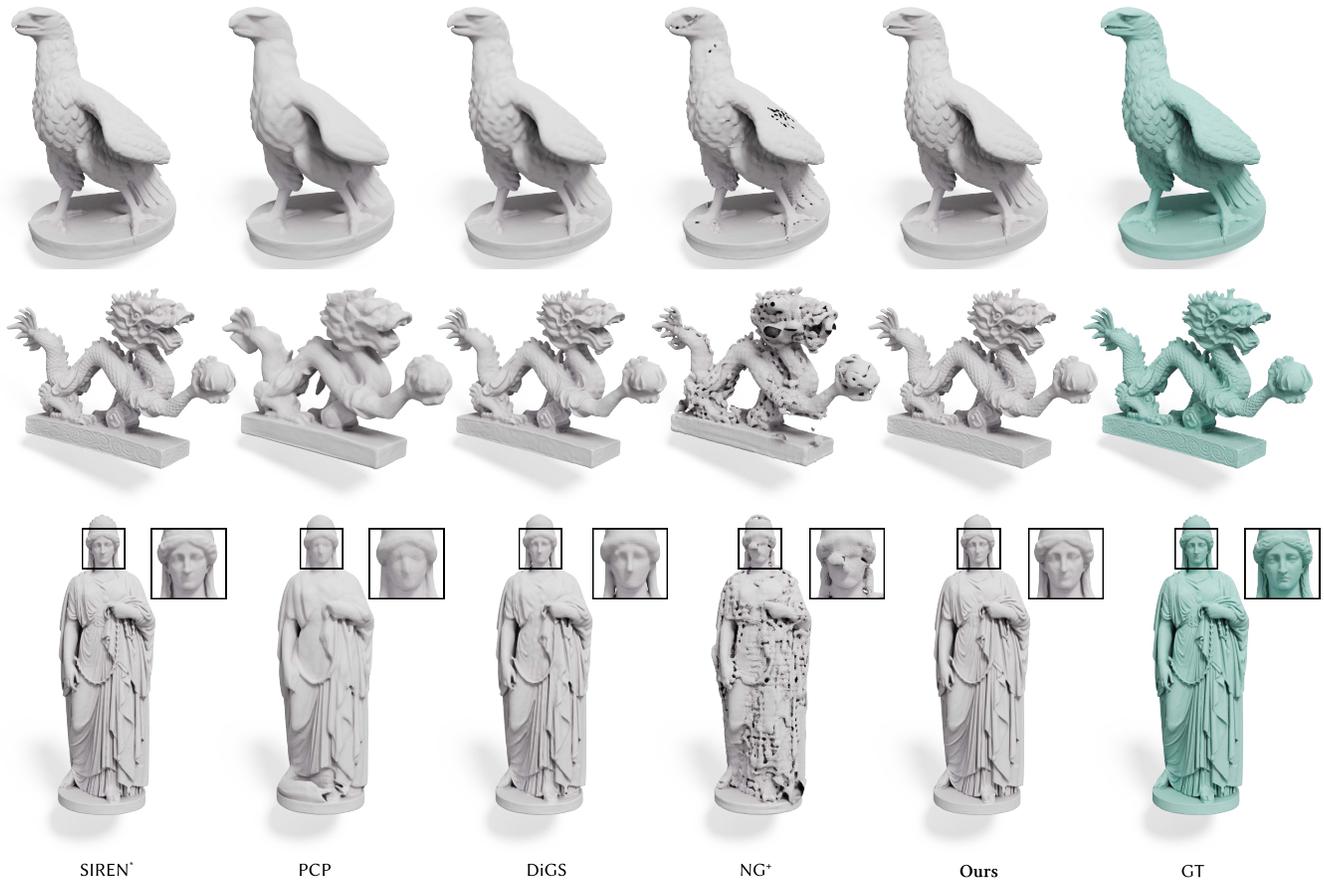}
    \caption{Visual comparison of different approaches on  ThreedScans~\cite{three_scans} with 10K points.
    Our method is comparable to the ``with normals'' version of SIREN.}
    \label{fig:large_supp}
\end{figure*}

The quantitative comparison statistics are reported 
in Tab.~\ref{tab:add_large}, while the visual comparison is available in Fig.~\ref{fig:large_supp}.

\subsection{Shape Space Learning}
Shape space learning requires training a single model to learn to represent multiple shapes from a class of related shapes, which is more challenging than the single overfitting shape.
For the encoder, we adopt the encoder from Convolutional Occupancy Network~\cite{Peng20ConvONet}.
Specifically, we project the sparse on-surface point features obtained using a modified PointNet~\cite{PointNet} onto a regular 3D grid, then use a convolutional module to propagate sparse on-surface point features to the off-surface area, and finally obtain the query feature using bilinear interpolation. 
For the decoder, we use the SIREN network has three hidden layers.
Further, we adopt the FiLM conditioning~\cite{piGAN2021} that 
applies an affine transformation to the network's intermediate features as SIREN is weak in handling high-dimensional inputs~\cite{piGAN2021, mehta2021modulated}. 
We train our models for 200 epochs using AMSGrad optimizer~\cite{adamgrad} with an initial learning rate of 0.0001 and decay to 0.000001 using cosine annealing~\cite{loshchilov2017sgdr}.
We divided the training set into mini-batches: a batch contains
32 different shapes (accumulate batches), where each shape is randomly sampled to produce 10K points.
The experiments are conducted with 8 RTX 3090 graphics cards.
In the inference stage, we fine-tune the whole network to perform high-ﬁdelity surface reconstruction for each shape 3000 iterations utilizing our loss without the critical term inspired by SA-ConvNet~\cite{SA-ConvONet}.

For baselines, we use the pre-trained model of IGR~\cite{IGR}, SAL~\cite{SAL}, SALD~\cite{SALD}, DualOctreeGNN~\cite{wang2022dual} and DiGS~\cite{DiGS}, and retrained the IGR and DualOctreeGNN for the version without normals supervision.
We optimize DiGS~\cite{DiGS} in the inference stage with 3000 iterations for auto-decoder looking forward to better performance.

\begin{figure*}[!htp]
    \centering
    \includegraphics[width=\textwidth]{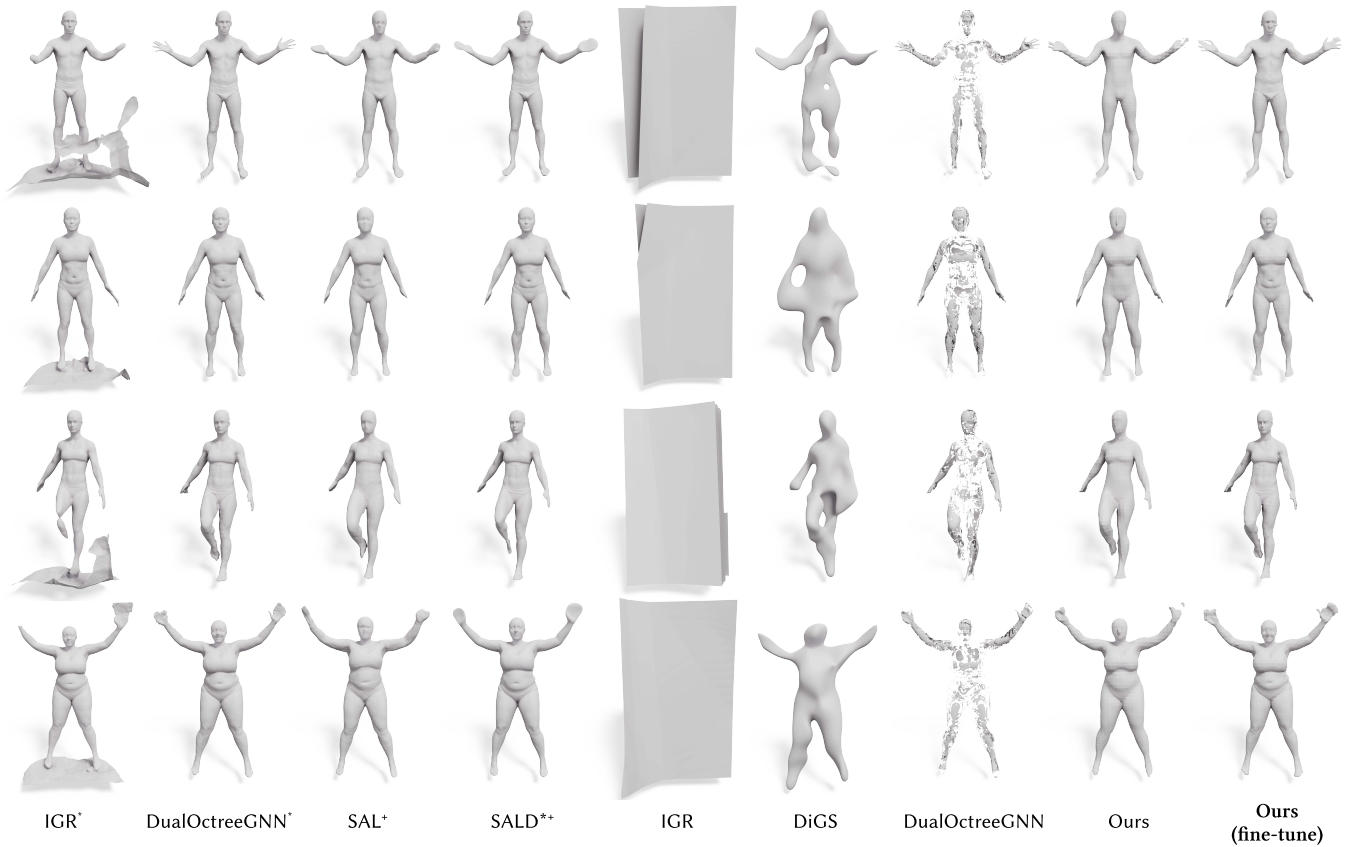}
    \caption{Visual comparison of shape space learning on  DFAUST~\cite{dfaust:CVPR:2017}. Our method can get reliable results without supervision and normals.}
    \label{fig:dfaust_supp}
\end{figure*}

The visual comparison of different approaches on the DFAUST~\cite{dfaust:CVPR:2017} dataset is available in Fig.~\ref{fig:dfaust_supp}.

% %\paragraph{Additional quantitative results}
% We show the quantitative results for each shape in Tab.~\ref{tab:add_large}.
% %\paragraph{Additional qualitative results}
% We show the qualitative results for each shape in Fig.~\ref{fig:large_supp}.
% Our results effectively learn great geometry details without normals.

\begin{table*}[!htp]
\centering
\caption{Quantitative comparison on the shapes from~\cite{three_scans}.}
\label{tab:add_large}
\resizebox{\linewidth}{!}{%
\begin{tabular}{l|ccc|ccc|ccc|ccc|ccc} 
\toprule
\multirow{2}{*}{} & \multicolumn{3}{c|}{Mean}                                  & \multicolumn{3}{c|}{Std.}                     & \multicolumn{3}{c|}{Eagle}                                 & \multicolumn{3}{c|}{Dragon}                                & \multicolumn{3}{c}{Hosmer}                                  \\
                  & Normal C.~$\uparrow$       & Chamfer~$\downarrow$ & F-Score~$\uparrow$ & Normal C.     & Chamfer       & F-Score~      & Normal C.~$\uparrow$      & Chamfer~$\downarrow$ & F-Score~$\uparrow$ & Normal C.~$\uparrow$       & Chamfer~$\downarrow$ & F-Score~$\uparrow$ & Normal C.~$\uparrow$      & Chamfer~$\downarrow$ & F-Score~$\uparrow$  \\ 
% \cline{1-2}\cmidrule{3-4}\cline{5-8}\cmidrule{9-10}\cline{11-11}\cmidrule{12-13}\cline{14-14}\cmidrule{15-16}
\midrule
SIREN*~\footnotesize{\cite{SIREN}}           & 98.38          & 0.96                 & 63.61              & 0.30          & 0.17          & 19.62         & \textbf{98.74} & 1.16                 & 42.55              & 98.21          & 0.84                 & 81.40              & 98.21          & 0.87                 & 66.89               \\ 
\midrule
PCP~\footnotesize{\cite{PCP}}               & 94.32          & 4.44                 & 11.83              & 2.58          & 1.41          & \textbf{9.41} & 97.27          & 3.01                 & 12.00              & 93.16          & 5.83                 & 2.33               & 92.51          & 4.47                 & 21.15               \\
DiGS~\footnotesize{\cite{DiGS}}              & 97.41          & 0.92                 & 63.78              & 0.62          & 0.22          & 21.91         & 98.08          & 1.10                 & 42.24              & 97.31          & 0.67                 & 86.06              & 96.84          & 0.98                 & 63.04               \\ 
\midrule
NG~\footnotesize{\cite{Galerkin}}                & 85.97          & 3.02                 & 31.01              & 9.10          & 2.09          & 20.67         & 95.89          & 1.43                 & 46.66              & 77.99          & 5.39                 & 7.57               & 84.03          & 2.24                 & 38.79               \\ 
\hline
\textbf{Ours}     & \textbf{98.44} & \textbf{0.74}        & \textbf{79.77}     & \textbf{0.15} & \textbf{0.09} & 9.90          & 98.53          & \textbf{0.84}        & \textbf{69.39}     & \textbf{98.53} & \textbf{0.66}        & \textbf{89.12}     & \textbf{98.26} & \textbf{0.73}        & \textbf{80.81}      \\
\bottomrule
\end{tabular}
}
\end{table*}

% \FloatBarrier
\bibliographystyle{ACM-Reference-Format}
\bibliography{bibliography}
\end{document}